\documentclass{article}

\usepackage{PRIMEarxiv}

\usepackage[T1]{fontenc}    
\usepackage{hyperref}       
\usepackage{url}            
\usepackage{booktabs}       
\usepackage{amsfonts}       
\usepackage{amsmath}		
\usepackage{amsfonts}       
\usepackage{bm}       		
\usepackage{bbm}       		
\usepackage{enumerate}      
\usepackage{url}			
\usepackage{caption}		
\usepackage{subcaption}		
\usepackage{graphicx}
\usepackage{fancyhdr}       
\usepackage{placeins}       
\usepackage{makecell}       
\usepackage{booktabs}

\usepackage[numbers]{natbib}

\newcommand{\lj}{{\ell_j}}
\newcommand{\li}{{\ell_i}}
\newcommand{\SVMperf}{$\text{SVM}^{perf}$}

\title{A Comparative Evaluation of Quantification Methods
}

\author{
  Tobias Schumacher \\
  University of Mannheim,\\
  RWTH Aachen University \\
  \texttt{tobias.schumacher@uni-mannheim.de} \\
   \And
  Markus Strohmaier \\
  University of Mannheim,\\
  GESIS - Leibniz Institute for the Social Sciences, and\\
  Complexity Science Hub\\
  \texttt{markus.strohmaier@uni-mannheim.de}
  \And
  Florian Lemmerich \\
  University of Passau\\
  \texttt{florian.lemmerich@uni-passau.de} \\
}

\begin{document}
\maketitle

\begin{abstract}
Quantification represents the problem of estimating the distribution of class labels on unseen data.
It also represents a growing research field in supervised machine learning, for which a large variety of different algorithms has been proposed in recent years.
However, a comprehensive empirical comparison of quantification methods that supports algorithm selection is not available yet. 
In this work, we close this research gap by conducting a thorough empirical performance comparison of 24 different quantification methods on overall more than 40 data sets, considering binary as well as multiclass quantification settings.
We observe that no single algorithm generally outperforms all competitors, but identify a group of methods including the threshold selection-based \emph{Median Sweep} and \emph{TSMax} methods, the \emph{DyS} framework including the \emph{HDy} method, \emph{Forman's mixture model}, and \emph{Friedman's method} that performs best in the binary setting.  
For the multiclass setting, we observe that a different, broad group of algorithms yields good performance, including the \emph{HDx} method, the \emph{Generalized Probabilistic Adjusted Count}, the \emph{readme} method, the \emph{energy distance minimization} method, the \emph{EM algorithm for quantification}, and \emph{Friedman's method}.
We also find that tuning the underlying classifiers has in most cases only a limited impact on the quantification performance.
More generally, we find that the performance on multiclass quantification is inferior to the results obtained in the binary setting.
Our results can guide practitioners who intend to apply quantification algorithms and help researchers to identify opportunities for future research.
\end{abstract}

\section{Introduction}

Quantification is the problem of estimating the distribution of class labels on unseen (test) data.
That is, after being trained on a dataset with known class labels, we want to estimate the number of instances of each class in a set of instances with given features but no class labels.
In contrast to traditional classification tasks, we are not interested in individual predictions, but only in aggregated values on a group level. 
For this problem setting, previous research has established that training a classification algorithm and counting instance-wise predictions does not generally yield accurate estimates \citep{forman_quantifying_2008, tasche_does_2016}. This has given rise to a relatively young, but vivid research field within the machine learning community.
As an increasing number of researchers are becoming aware of this issue, a growing number of novel methods have been proposed.

While a first review of existing quantification methods has been given by \citet{gonzalez_review_2017}, and recent publications also provide broader frameworks for quantification learning \citep{maletzke_dys_2019, maletzke_importance_2020}, a thorough, empirical, and independent comparison of quantification methods has not been presented yet.
In this work, we aim to fill this research gap by providing a comparison of 24 different quantification algorithms over 40 datasets.
Apart from assessing approaches for the binary class setting, we also include experiments for the multiclass quantification setting, which has received limited attention in quantification research so far.
For each dataset and algorithm, we evaluate several degrees of distribution shifts between training data and test data with varying training set sizes. 
Furthermore, we evaluate whether applying more accurate base classifiers will also yield better performance of the quantifiers using these.
Altogether, these experiments encompass more than 5 million algorithm runs.
To further validate our findings, we conduct a case study using the external competitive benchmark from the 2022 LeQua challenge \citep{esuli_lequa_2022,esuli_concise_2022}.

Our experiments with binary class labels show that there is not a single algorithm that outperforms all others---but we identify a group of algorithms consisting of the threshold selection-based \emph{Median Sweep} and \emph{TSMax} methods~\citep{forman_quantifying_2008}, \emph{Friedman's method} \citep{friedman_friedman_2014}, \emph{Forman's mixture model}, \citep{forman_counting_2005} and the \emph{DyS} framework~\citep{maletzke_dys_2019}, including the \emph{HDy} method \citep{gonzalez-castro_class_2013}, that on average perform significantly better.
We also show that algorithms that are based on optimizing a classifier for the quantification problem show on average worse performance, implying that their benefits in practice might be restricted to particular scenarios.

In the multiclass setting, we also find a group of algorithms which perform significantly better on average than the rest. This group includes the \emph{HDx method} \citep{gonzalez-castro_class_2013}, \emph{Generalized Probabilistic Adjusted Count} \citep{bella_quantification_2010, firat_unified_2016}, \emph{readme} \citep{hopkins_method_2010}, \emph{energy distance minimization} \citep{kawakubo_computationally_2016}, the \emph{EM algorithm for quantification} \citep{saerens_adjusting_2002}, and \emph{Friedman's method} \citep{friedman_friedman_2014}.
These algorithms share the characteristic that they naturally allow for multiclass quantification.
By contrast, extending predictions from binary quantifiers to the multiclass case in a one-vs.-rest fashion does not appear to yield competitive results, even when using strong base quantifiers such as the \emph{Median Sweep} or \emph{HDy}.
More generally, we observe significantly weaker performance for the multiclass case, corroborating that multiclass quantification constitutes a harder research problem and might need more research attention in the future.
Further, we observe that classifiers which were tuned for classification accuracy do not, in general, improve the predictions of the quantifiers applying them.

Overall, our results guide practitioners toward the most propitious quantification approaches for certain applications, and help researchers to identify promising future research avenues.

In the following, we first briefly introduce the quantification problem and describe how it conceptually differs from the classification problem.
Afterward, Section \ref{sec:algorithms} gives an overview of the algorithms included in our experimental comparison, providing a summary of the state-of-the-art in quantification.
Next, in Section \ref{sec:setup}, we provide a thorough description of the experimental setup of our comparison, before giving an in-depth presentation of the experimental results in Section \ref{sec:results}.
In Section \ref{sec:lequa}, we present the results of the case study on the dataset from the LeQua challenge.
Finally, in Section \ref{sec:discussion}, we discuss the results of our experiments, before closing with a conclusion in Section \ref{sec:conclusion}.

\section{The Quantification Problem}

Quantification is a supervised machine learning problem that aims to estimate the distribution of class labels in a test set instead of predicting the class of individual instances.
Throughout this paper, we use the following notations.
For training, we are given a dataset of instances $D_{\text{train}}$, for which we know the values of multiple (categorical or continuous) features $X$ and the corresponding class label $Y$.
Letting $L$ denote the number of possible values of the class label, we distinguish between the \emph{binary} case, i.e., there are exactly $L = 2$ possible values for the class label, and the \emph{multiclass} case, i.e., there are $L > 2$ options for the class label.
Using the training data, the goal is then to train a model that allows to predict the distribution of the class label in some test data $D_{\text{test}}$, for which only the values of the features $X$ are known.
In the following, we will often use the term \emph{prevalence} for the relative frequency of single labels in training or test data.
We formally denote the distributions of $X$ and $Y$ in the training set by $P_{\text{train}}(X)$ and $P_{\text{train}}(Y)$, and their distribution in the test set by $P_{\text{test}}(X)$ and $P_{\text{test}}(Y)$.
Since in the binary case, the full distribution is already specified by the share of one class, we will denote for shorter notation the instances from one arbitrary class as \emph{positives}, and label their prevalence in training and test data as $pos_{\text{train}}$ and $pos_{\text{test}}$, respectively.

In contrast to traditional classification, a \emph{shift} of the distribution of the class label $Y$, i.e., a difference between the class probabilities $P_{\text{train}}(Y)$ in the training set and the class probabilities $P_{\text{test}}(Y)$ in the test set, is expected.
However, it is assumed that the conditional distributions $P(X|Y)$ are stable between training and test sets---this kind of distribution shift is also known as \emph{prior probability shift} \citep{storkey_shift_2008}.
Furthermore, compared to classification, it is also more common to expect the occurrence of training or test instances with the exact same feature values but different labels.

A trivial approach to quantification, known as \emph{Classify and Count (CC)} method, applies an arbitrary classification method trained on the training data to the test data and predicts the distribution of the predicted labels.
However, this has been theoretically and empirically shown to achieve insufficient results in many scenarios~\citep{forman_quantifying_2008,tasche_does_2016}.

\section{Algorithms for Quantification}
\label{sec:algorithms}

We first outline the quantification algorithms under consideration. 
Following a previous categorization \citep{gonzalez_review_2017}, we distinguish between (i) adjusted count adaptations, (ii) distribution matching methods, and (iii) adaptations of traditional classification algorithms.
An overview of the considered algorithms is given in Table \ref{tab:algs}. 
In that table, the column \textit{Continuous} indicates whether an algorithm is also able to handle datasets with continuous features, and \textit{Multiclass} indicates whether an algorithm is not restricted to binary quantification. Next to that, this table also lists abbreviations that we are often going to use when discussing these algorithms.

\begin{table}[t]
\centering
\small
\resizebox{\linewidth}{!}{

\begin{tabular}{ l c  c  c c }\hline
\toprule

Quantification Algorithm & Abbreviation & Reference & Multiclass & Continuous\\ \midrule
Adjusted Count & AC  & \citet{forman_counting_2005}& OVR & Yes\\
Probabilistic Adjusted Count & PAC & \citet{bella_quantification_2010} & OVR & Yes\\
Threshold Sel. Policy X & TSX &\citet{forman_quantifying_2008}& OVR & Yes\\
Threshold Sel. Policy T50 & TS50 &\citet{forman_quantifying_2008}& OVR & Yes\\
Threshold Sel. Policy Max  & TSMax &\citet{forman_quantifying_2008}& OVR & Yes\\
Threshold Sel. Policy Median Sweep  & MS &\citet{forman_quantifying_2008}& OVR & Yes\\

\midrule
Generalized Adjusted Count & GAC &\citet{firat_unified_2016}& Yes & Yes\\
Generalized Prob. Adjusted Count & GPAC& \citet{firat_unified_2016}& Yes & Yes\\
DyS Framework (Topsøe Distance) & DyS& \citet{maletzke_dys_2019}& OVR & Yes \\
Forman's Mixture Model & FMM &\citet{forman_quantifying_2008}& OVR & Yes\\
readme & readme & ~\citet{hopkins_method_2010}& Yes & No\\
HDx & HDx &\citet{gonzalez-castro_class_2013}& Yes & No\\
HDy & HDy &\citet{gonzalez-castro_class_2013}& OVR & Yes\\
Friedman's method & FM &\citet{friedman_friedman_2014}& Yes & Yes\\
Energy Distance Minimization & ED &\citet{kawakubo_computationally_2016}& Yes & Yes \\
EM-Algorithm for Quantification & EM &\citet{saerens_adjusting_2002}& Yes & Yes\\
CDE Iteration & CDE &\citet{tasche_fisher_2017}& No & Yes\\
\midrule
Classify and Count & CC &\citet{forman_quantifying_2008}& Yes & Yes\\
Probabilistic Classify and Count & PCC &\citet{bella_quantification_2010}& Yes & Yes\\
\SVMperf using \emph{KLD} loss & SVM-K &\citet{esuli_ai_2010}& No & Yes\\
\SVMperf using $Q$-measure loss & SVM-Q &\citet{barranquero_quantification-oriented_2015}& No & Yes\\
Nearest-Neighbor-Quantification & PWK &\citet{barranquero_study_2013} & Yes & No\\
Quantification Forest  & QF &\citet{milli_quantification_2013}& Yes & No\\
AC-corrected Quantification Forest  & QF-AC &\citet{milli_quantification_2013}& No & No\\

\bottomrule
\end{tabular}
}
\caption{Overview of considered quantification algorithms. For each algorithm, we denote its abbreviation, a reference, and its ability to handle multiclass labels and continuous features. Regarding the multiclass setting, we distinguish between algorithms that can naturally handle this setting (Yes), algorithms that require a one-vs.-rest approach (OVR), and algorithms that were not considered in our multiclass experiments (No).}
\label{tab:algs}
\end{table}

\subsection{Adjusted Count Adaptations}\label{sec:ac}
The trivial \emph{Classify and Count (CC)} method just applies an arbitrary classifier $c$ on the test data and counts the number of respective predictions.
The core idea behind the \emph{Adjusted Count (AC)} approach is to adjust these results post hoc for potential biases.
This is accomplished by exploiting the assumption that the likelihood $P(X|Y)$ of the features $X$ given the class label $Y$ does not vary between training and test data.
Assuming binary labels, the true positive rate (\emph{tpr}) and false positive rate (\emph{fpr}) of a classifier, which correspond to the probabilities $P(c(X)=1|Y=1)$ and $P(c(X)=1|Y=0)$, respectively, can be expected to be identical between training and test data---see also Appendix \ref{ap:clf_performance}, Equation \ref{eq:tpr_fpr} for formal definitions of these rates. 
Letting $\widehat{pos}_{\text{test}}$ denote the predicted prevalence of positives by the \emph{CC} method, we can express this quantity in terms of the true prevalence of positives $pos_{\text{test}}$ and the (mis)classification rates \emph{tpr} and \emph{fpr} via 
\[
    \widehat{pos}_{\text{test}} = pos_{\text{test}} \cdot tpr + (1-pos_{\text{test}}) \cdot fpr,
\]
which we can solve for $pos_{\text{test}}$ to obtain the \emph{AC} estimation
\begin{equation}\label{eq:AC}
   pos_{\text{test}} = \frac{\widehat{pos}_{\text{test}} - fpr}{tpr-fpr}.
\end{equation}

In practice, it can occur that the estimate falls out of the feasible interval $[0,1]$, hence, in such cases, one would have to clip the outcome to the boundary values.

\vspace{\baselineskip}
Based on that idea, a few variations of the \emph{AC} method have been introduced in the literature, and the following methods will be included in our experiments:

\begin{enumerate}
    \item \textbf{Adjusted Count (AC). } As described above, we estimate the true positive and false positive rates from the training data and utilize them to adjust the output of the \emph{CC} method \citep{forman_counting_2005}.
    \item \textbf{Probabilistic Adjusted Count (PAC). } This method adapts the \emph{AC} approach by using average class-conditional confidences from a probabilistic classifier instead of true positive and false positive rates \citep{bella_quantification_2010}.
    \item \textbf{Threshold Selection Policies (TSX, TS50, TSMax, MS). } The core idea of these variations is to shift the decision boundary (e.g., classify an instance as positive if the original estimate $c(x)$ is larger than 0.7) of the underlying classifier in order to make the \emph{AC} estimation in Equation \ref{eq:AC} more stable numerically. 
    Different strategies involve using the threshold that maximizes the denominator $tpr - fpr$ (\emph{TSMax}), a threshold for which we have $fpr = 1 - tpr$ \emph{(TSX)}, a threshold at which $tpr \approx 0.5$ holds (\emph{TS50}), or, as in the \emph{Median Sweep} (\emph{MS}), using an ensemble of such threshold-based methods and taking the median prediction \citep{forman_quantifying_2008}.
\end{enumerate}

\subsection{Distribution Matching Methods}
The majority of existing quantification methods can be categorized as \emph{distribution matching algorithms}.  
These algorithms are all implicitly based on the assumption that the distribution of the features $X$ conditioned on the distribution of the class labels $Y$ does not change between training data and test data. Under that assumption, with $\lj, j\in\{1,\dots,L\}$, denoting the possible values of the labels $Y$, the law of total probability yields that
\begin{equation}\label{eq:dmbase}
P_{\text{test}}(X) = \sum_{j=1}^{L} P_{\text{train}}(X|Y = \ell_j) P_{\text{test}}(Y = \ell_j).
\end{equation}
As in this equation, both the left-hand side distribution $P_{\text{test}}(X)$ and the conditional distributions $P_{\text{train}}(X|Y = \ell_j)$ on the right-hand side can be seen as represented by given training and test data, only the sought-for probabilities $P_{\text{test}}(Y = \ell_j)$ are left as unknowns. 
To estimate these class probabilities, there are two main issues to be worked out from a methodological point of view.
First, estimating or modeling the distributions $P_{\text{train}}(X|Y = \ell_j)$ and $P_{\text{test}}(X)$ is not at all trivial.
There can be an arbitrary amount of features $X$, and the training data usually does not provide nearly enough samples to accurately represent the distribution of the feature space, even more when conditioning on the class labels $Y$.
Second, once an estimation of the distributions $P_{\text{test}}(X)$ and $P_{\text{train}}(X|Y = \ell_j)$ has been made, there are also various ways to predict the class probabilities $P_{\text{test}}(Y = \ell_j)$ from these estimations.

The methods discussed in this chapter tackle these issues in various ways.
One basic approach to tackle the first issue has, for instance, already been introduced when discussing the \emph{Adjusted Count}. 
In the \emph{Adjusted Count} approach, information on the distribution of the features $X$ was derived by the means of applying a classifier $c$ and considering the distribution of their outputs $P(c(X))$ instead of $P(X)$. 
That way, Equation \ref{eq:dmbase} would be transformed to the set of linear equations
\begin{equation}\label{eq:gac}
P_{\text{test}}(c(X)=\li) = \sum_{j=1}^{L} P_{\text{train}}(c(X)=\li|Y = \lj) P_{\text{test}}(Y = \lj),\quad i\in\{1,\dots,L\}.
\end{equation}
However, there are also methods that do not apply classifiers, and instead, for instance, estimate $P(X)$ based on the distributions of single features, or in terms of distances between individual instances in the data.
Regarding the second issue, most of the presented methods translate Equation \ref{eq:dmbase} into a set of linear equations, and then minimize some distance function between the left- and right-hand-side expressions, subject to the constraints that $\sum_{j=1}^L P_{\text{test}}(Y = \ell_j) = 1$ and $P_{\text{test}}(Y = \ell_j)\geq 0$ for all $j\in \{1, \dots, L\}$ have to hold. 
This pattern has already been noted by \citet{firat_unified_2016}.

Among all the methods of this category, we compare the following methods:
\begin{enumerate}
    \item \textbf{Generalized Adjusted Count Models (GAC, GPAC).} As described above, the most simple work-around to avoid estimating $P(X)$ is to apply a classifier to build a system of linear questions as in Equation \ref{eq:gac}, and solve it via constrained least-squares regression \citep{firat_unified_2016}. 
    That approach can be considered as a \textit{Generalized Adjusted Count (\emph{GAC})} method, which also naturally includes the multiclass case.
    Similarly, one can obtain the \textit{Generalized Probabilistic Adjusted Count} (\emph{GPAC}) method, by utilizing the posterior probabilities from probabilistic classifiers as in the \emph{PAC} method.

    \item \textbf{The DyS Framework (DyS, HDy).} More recently, \citet{maletzke_dys_2019} proposed the \emph{DyS} framework, in which the main idea is to utilize confidence scores resulting from the decision functions of a binary classifier.
    More precisely, the confidence scores obtained on the training data are divided into bins, and then the probability that the confidence score of an instance ends up in that bin is estimated from the training set.
    Thus, in our context, the number of linear equations we obtain from Equation \ref{eq:dmbase} equals the chosen number of bins, which, next to the distance function that this set of equations is optimized on, can be seen as a parameter of this framework.
    A main drawback of this framework is that it only works for the binary case, and that many of the distance functions that were proposed and evaluated for this framework are not convex, requiring methods such as ternary search to estimate the optimal solution.
    Since in their evaluation, using the Topsøe distance \citep{deza_encyclopedia_2009} has proven to yield consistently good results, we are applying this setup as \emph{DyS} method in our experiments.
    Furthermore, it is noteworthy that this framework was motivated as a generalization of the \emph{HDy} method \citep{gonzalez-castro_class_2013}, which uses the Hellinger distance to match  distributions.
    

    \item 
    \textbf{Forman's Mixture Model (FMM)}.
    Like the DyS framework, this method is based on matching distributions of classifier scores. 
    Yet, instead of matching probability density functions which are estimated from binned classifier scores, \citet{forman_counting_2005} proposed to match the cumulative distributions of classifier scores to avoid sparsity issues.
    To match these distributions, Forman proposed minimizing their \emph{PP-area}, which practically corresponds to minimizing the Manhattan distance \citep{firat_unified_2016}.

    \item \textbf{Friedman's Method (FM).} Similar to the \emph{GPAC} method, \citet{friedman_friedman_2014} proposed to utilize the confidence scores from probabilistic classifiers. However, instead of averaging the class-conditional confidence scores, he proposes to utilize the fraction of class-conditional confidence scores that are above and below the observed class prevalences in the training data. 
    
    \item 
    \textbf{Feature Distribution Matching (readme, HDx).}
    Instead of applying a classifier, one can also directly model the distribution of features by counting co-occurences of multiple features as in the \emph{readme} method~\citep{hopkins_method_2010}, or by counting occurrences of individual features as in the \emph{HDx} method~\citep{gonzalez-castro_class_2013}.
    This requires that all features are categorical, or preprocessed accordingly via binning.
    In the \emph{readme} method, one then matches the distributions via constrained least-squares regression.
    Due to sparsity issues, this is, however, only done by considering a random subset of all features. Yet, multiple of such subsets are drawn, and the resulting predictions are averaged to obtain the final estimate of the true class distribution. 
    In the \emph{HDx} method \citep{gonzalez-castro_class_2013}, by contrast, distributions of single features are aggregated and matched via the Hellinger distance.

   \item \textbf{Energy Distance Minimization (ED).} As the name of this method suggests, its core idea is to minimize the energy distance between the left-hand and right-hand side distribution in Equation \ref{eq:dmbase}. In that context, the distribution of the feature space is intrinsically modeled by the Euclidean distances between individual instances, and therefore no classifiers or additional parameters are required \citep{kawakubo_computationally_2016}.
   
   \item \textbf{The EM Algorithm for Quantification (EM).} This method applies the classic \emph{EM algorithm} \citep{dempster_maximum_1977} on the outputs of probabilistic classifiers to adjust them for potential distribution shift between the class distributions in training and test data. While quantification was not the main focus in the original proposal of the algorithm \citep{saerens_adjusting_2002}, the sought-for class prevalences are obtained as a side-product.
   
   \item \textbf{CDE Iteration (CDE).} The \emph{class distribution estimation (CDE) iterator} \citep{xue_quantification_2009} applies principles from cost-sensitive classification to account for changes in class distributions between training and test data. For that purpose, the misclassification costs are updated iteratively, and in the original proposition of the algorithm, the underlying classifier is retrained in every iteration step.
   In our experiments, we use the more efficient variant proposed by \citet{tasche_fisher_2017}, in which each iteration updates the decision threshold of an underlying probabilistic classifier. For this variant of the algorithm, Tasche has also proven that the iteration will eventually converge.
\end{enumerate}


\subsection{Classifiers for Quantification}
Classifiers for quantification apply established classification methods to the setting of quantification.
The main idea behind most of these methods is to optimize such classifiers based on a loss function that minimizes the quantification error, and then estimate the class distributions based on the predictions of the individual instances.
Thus, these approaches are all in some sense variants of the \emph{CC} method.
In our experiments, we included the following methods:

\begin{enumerate}
\item \textbf{Classify and Count (CC).} This trivial approach applies a classifier and counts the number of times each class is predicted \citep{forman_quantifying_2008}.

\item \textbf{Probabilistic Classify and Count (PCC).} This method takes probabilistic predictions, i.e., continuous values between zero and one, and averages the predictions of all instances to estimate the class prevalences \citep{forman_quantifying_2008, bella_quantification_2010}.

\item \textbf{\SVMperf optimization (SVM-Q, SVM-K).} This pair of methods applies the so-called \SVMperf classifier, which is an adaptation of traditional support vector machines which can be optimized for multivariate loss functions \citep{joachims_support_2005}.
Based on this algorithm, multiple classifiers with different quantification-oriented loss functions have been proposed. 
For instance, \citet{esuli_ai_2010} have proposed to use the Kullback-Leibler Divergence (SVM-K), while \citet{barranquero_quantification-oriented_2015} have developed the \emph{Q-measure} for that purpose (SVM-Q).

\item \textbf{Nearest-Neighbor based Quantification (PWK).} \citet{barranquero_study_2013} have adapted nearest-neighbor based classification to the setting of quantification. In their $k$-NN approach, they apply a weighting scheme which applies less weight on neighbors from the majority class.

\item \textbf{Quantification Forests (QF, QF-AC).} The decision tree and random forest classifiers have been adapted for quantification by \citet{milli_quantification_2013}. Other than in the traditional approach, the authors propose that the split in each decision tree is made based on a quantification-oriented loss functions. Since in their original proposition, applying the \emph{AC} method on the predictions from these random forests yielded particularly strong results, we include both the \emph{Quantification Forests} as well as the \emph{AC} adaption of them in our experiments.

\end{enumerate}

\subsection{Multiclass Quantification}
In the literature on quantification, multiclass quantification has received surprisingly little attention so far, despite \citet{forman_quantifying_2008} pointing out that this task is much harder than binary quantification.
In our comparative evaluation, we also take a closer look into this scenario.

In contrast to classification, approaches based on a one-vs.-one comparison of two class labels are, to the authors' knowledge, not used for quantification, potentially since it is unclear how to aggregate the results.
Therefore, methods for multiclass quantification can be broadly distinguished into two categories:
\begin{enumerate}
    \item \textbf{Natural Multiclass Quantifiers.}  Like in classification, some quantification algorithms can naturally handle the multiclass setting as well.
    This is the case for most distribution matching methods, as by Equation \ref{eq:dmbase}, there is no constraint on the number of classes that are summated. 
    Further, quantification-oriented classifiers such as Quantification Forests can also handle the multiclass setting, since the underlying classifier allows for it.
    \item \textbf{One-vs.-Rest Quantifiers}. Traditional quantification methods such as the \emph{adjusted count adaptations} have been designed specifically for the binary setting. 
    To extend such methods to the multiclass setting, one can estimate the prevalence of each individual class in a one-vs.-rest fashion, and then normalize the resulting estimations afterward so that they sum to 1 \citep{forman_quantifying_2008}.
    Next to all \emph{adjusted count adaptations}, we also applied this strategy for the distribution matching methods from the $DyS$ framework, and Forman's mixture model, as these do not naturally generalize to the multiclass setting. 
\end{enumerate}

We provide an overview regarding which multiclass strategy is used for each quantification algorithm in Table \ref{tab:algs}. 
For the \emph{SVM-K}, \emph{SVM-Q} and \emph{QF-AC} methods we did not conduct any multiclass experiments, as the underlying implementations did not provide a multiclass feature.
Furthermore, for the CDE iterator we did not run multiclass experiments either, since the individual one-vs.-rest predictions yielded extreme predictions of either 0 or 1 regularly.

\section{Experimental Setup}\label{sec:setup}

Overall, we compare 24 algorithms on 40 datasets. In the following, we provide details regarding the datasets, sampling protocols, algorithmic parameters, and evaluation measures. The implementation of the algorithms and experiments can be found on GitHub\footnote{\url{https://github.com/tobiasschumacher/quantification_paper}}.

\subsection{Datasets}
We applied all algorithms on a broad range of 40 datasets collected from the UCI machine learning repository\footnote{\url{https://archive.ics.uci.edu/ml/index.php}} and from Kaggle\footnote{\url{https://www.kaggle.com/datasets}}.
An overview of the datasets that we used in our experiments is provided by Table~\ref{tab:data}. 
It includes information on the amount of instances per dataset $N$, the number of features $D$, the number $L$ of classes of the corresponding class variable, and an indicator about whether a dataset contains non-categorical (continuous) features. 
The non-categorical aspect is relevant for quantification algorithms such as \emph{readme} that require a finite feature space.

\begin{table}
 \small
    \resizebox{\linewidth}{!}{
     \begin{tabular}{lcrcrrc}
\toprule
                                           Dataset &   Abbr. &    $D$ & Non-Categorical &      $N$ &  $L$ & Source \\
\midrule
                           Internet Advertisements &     ads & 1560 &             Yes &   2359 &  2 &    UCI \\
                                             Adult &   adult &   89 &             Yes &  45222 &  2 &    UCI \\
                       Student Alcohol Consumption &    alco &   57 &             Yes &   1044 &  2 & Kaggle \\
                                             Avila &   avila &   10 &             Yes &  20867 &  2 &    UCI \\
              Breast Cancer Wisconsin (Diagnostic) &  bc-cat &   31 &             Yes &    569 &  2 &    UCI \\
                Breast Cancer Wisconsin (Original) & bc-cont &   10 &             Yes &    683 &  2 &    UCI \\
                              Bike Sharing Dataset &    bike &   59 &             Yes &  17379 &  4 &    UCI \\
                                      BlogFeedback &    blog &  280 &             Yes &  52397 &  4 &    UCI \\
                 MiniBooNE Particle Identification &   boone &   50 &             Yes & 129569 &  2 &    UCI \\
                                   Credit Approval &   cappl &   44 &             Yes &    653 &  2 &    UCI \\
                                    Car Evaluation &    cars &   22 &              No &   1728 &  2 &    UCI \\
                    Default of Credit Card Clients &   ccard &   34 &             Yes &  30000 &  2 &    UCI \\
                     Concrete Compressive Strength &    conc &    8 &             Yes &   1030 &  3 &    UCI \\
                            Superconductivity Data &    cond &   89 &             Yes &  21263 &  4 &    UCI \\
                       Contraceptive Method Choice &  contra &   13 &             Yes &   1473 &  3 &    UCI \\
                          SkillCraft1 Master Table &   craft &   18 &             Yes &   3338 &  3 &    UCI \\
                                          Diamonds &    diam &   22 &             Yes &  53940 &  3 & Kaggle \\
                               Dota2 Games Results &    dota &  116 &              No & 102944 &  2 &    UCI \\
                                  Drug Consumption &   drugs &  136 &             Yes &   1885 &  3 &    UCI \\
                      Appliances Energy Prediction &    ener &   25 &             Yes &  19735 &  3 &    UCI \\
                   FIFA 19 Complete Player Dataset &    fifa &  117 &             Yes &  14751 &  4 & Kaggle \\
                                       Solar Flare &   flare &   28 &              No &   1066 &  2 &    UCI \\
          Electrical Grid Stability Simulated Data &    grid &   11 &             Yes &  10000 &  2 &    UCI \\
                             MAGIC Gamma Telescope &   magic &   10 &             Yes &  19020 &  2 &    UCI \\
                                          Mushroom &    mush &  111 &              No &   8124 &  2 &    UCI \\
                    Geographical Original of Music &   music &  116 &             Yes &   1059 &  2 &    UCI \\
                                  Musk (Version 2) &    musk &  166 &             Yes &   6598 &  2 &    UCI \\
News Popularity in Multiple Social Media Platforms &    news &   60 &             Yes &  39644 &  4 &    UCI \\
                                           Nursery &   nurse &   27 &              No &  12960 &  3 &    UCI \\
                               Occupancy Detection &   occup &    5 &             Yes &  20560 &  2 &    UCI \\
                                 Phishing Websites &   phish &   31 &              No &  11055 &  2 &    UCI \\
                                          Spambase &    spam &   58 &             Yes &   4601 &  2 &    UCI \\
                     Students Performance in Exams &   study &   19 &             Yes &   1000 &  2 & Kaggle \\
                              Telco Customer Churn &   telco &   45 &             Yes &   7032 &  2 & Kaggle \\
                       First-order Theorem Proving &    thrm &   51 &             Yes &   6117 &  3 &    UCI \\
                        Turkiye Student Evaluation &    turk &   31 &              No &   5820 &  3 &    UCI \\
                                  Video Game Sales &   vgame &  133 &             Yes &  6825 &  4 & Kaggle \\
                       Gender Recognition by Voice &   voice &   20 &             Yes &   3168 &  2 & Kaggle \\
                                      Wine Quality &    wine &   14 &             Yes &   6497 &  4 &    UCI \\
                                             Yeast &   yeast &    9 &             Yes &   1484 &  5 &    UCI \\
\bottomrule
\end{tabular}

    }
	\caption{Datasets used in our experiments. $D$ indicates the number of features, $L$ indicates the number of classes, $N$ corresponds to the number of instances in the data, and \textit{Non-Categorical} indicates whether a dataset contains features that required binning. This latter aspect is relevant for quantification algorithms such as \emph{readme} that require a finite feature space.}
	\label{tab:data}
\end{table}

Out of the 40 datasets, 17 had a non-binary set of class labels, or were even regression datasets. 
The regression datasets were converted to both multi-class and binary datasets by binning the values of the class variable.
This was usually done with the abstract goal of achieving groups of similar size with respect to the number of instances to allow for a more robust basis for potential shifts in the following steps.
The cut-off points for the bins were determined manually after looking at the distribution of the classes.
Furthermore, the real multiclass datasets were also converted to binary datasets.
In these cases, we kept the most populated class as is, and merged the other classes into a single class, like in a one-vs.-rest classification problem.
By doing so, we preserved meaningful class semantics that classifiers and quantifiers could recognize.
All datasets have been preprocessed the same way as for standard classification, including dummy coding their non-ordinal features, rescaling their continuous features, and removing missing values. 
Furthermore, to enable the application of algorithms that require a finite feature space, we created a variation of each dataset in which all non-categorical features were binned.
All algorithms that could handle a non-finite feature space were run on the non-binned datasets. 
While one may argue that due to these alterations in the datasets the results would be less comparable, the binning procedure ultimately simulates the loss of information that one would have to accept when applying such restricted algorithms in the first place.

Overall, these datasets represent a wide range of domains, and are differently shaped in terms of their number of instances as well as in the design of their feature spaces.

\subsection{Sampling Strategy}

As we aim to evaluate quantifiers under a large set of diverse conditions, we chose a sampling approach in which we varied (i) the training distribution, (ii) the test distribution and (iii) the (relative) sizes of training and test datasets.
Regarding training and test distributions, in the binary case, we considered different prevalences of training positives $pos_{\text{train}}$ and test positives $pos_{\text{test}}$ in the respective sets
\begin{align*}
&pos_{\text{train}}\in\{0.05, 0.1,0.3,0.5,0.7,0.9\}\quad  \text{and} \\
&pos_{\text{test}} \in \{0, 0.01,0.05, 0.1,0.2,0.3,0.4,0.5,0.6,0.7,0.8,0.9\},
\end{align*}
following the protocol introduced by \citet{forman_quantifying_2008}.
In both distributions, we sampled broadly across the interval $[0,1]$, including very unbalanced and thus presumably difficult settings with only very few (or, for the test set, even no) positive labels.

Concerning the multiclass case, we have considered datasets with a varying number of $L\in\{3,4,5\}$ different classes. 
For each of these values of $L$, we fixed a set of three training and five test class distributions, representing relatively uniform as well as polarized class distributions, which can be seen in Table \ref{tab:mcdists}.

\begin{table}[t]
	\centering
	\begin{tabular}[t]{|l|r|r|} \hline
		$L$ & Training Distributions $P_{\text{train}}(Y)$ & Test Distributions $P_{\text{test}}(Y)$ \\\hline
		3 & \begin{tabular}[t]{@{}r@{}} $(0.2, 0.5, 0.3)$,\\$(0.05, 0.8, 0.15)$,\\$(0.35,0.3,0.35)$\end{tabular} 
			& \begin{tabular}[t]{@{}r@{}} $(0.1,0.7,0.2)$,\\$(0.55,0.1,0.35)$,\\$(0.35,0.55,0.1)$,\\$(0.4,0.25,0.35)$,\\$(0.,0.05,0.95)$
			 \end{tabular}\\ \hline
		4 & \begin{tabular}[t]{@{}r@{}} $(0.5,0.3,0.1, 0.1)$,\\$(0.7,0.2,0.1,0.1)$,\\ $(0.25,0.25,0.25,0.25)$
		\end{tabular}
			& \begin{tabular}[t]{@{}r@{}} 		$(0.65,0.25,0.05,0.05)$,\\$(0.2,0.25,0.3,0.25)$,\\$(0.45,0.15,0.2,0.2)$,\\$(0.2,0,0,0.8)$, \\ $(0.3,0.25,0.35,0.1)$
 			\end{tabular}\\\hline
		5 & \begin{tabular}[t]{@{}r@{}} $(0.05,0.2,0.1,0.2, 0.45)$,\\$(0.05,0.1,0.7,0.1,0.05)$,\\$(0.2,0.2,0.2,0.2, 0.2)$
		\end{tabular} 
			& \begin{tabular}[t]{@{}r@{}} $(0.15,0.1,0.65,0.1,0)$,\\$(0.45,0.1,0.3,0.05,0.1)$,\\$(0.2,0.25,0.25,0.1,0.2)$,\\$(0.35,0.05,0.05,0.05,0.5)$,\\$(0.05,0.25,0.15,0.15,0.4)$
			\end{tabular}\\\hline
	\end{tabular}
\caption{List of training distributions $P_{\text{train}}(Y)$ and test distributions $P_{\text{test}}(Y)$ considered for experiments in the multiclass setting, ordered by number of classes $L$. In both the columns for training and test distribution, each row represents a distribution of instances that was sampled from the corresponding data. For instance, assuming that for a dataset with $L=3$ classes, the corresponding labels are given by $Y \in \{1,2,3\}$, the first row among the column of training distribution indicates that in our experiments, we have sampled training sets where Label $1$ had a prevalence of $0.2$, Label $2$ had a prevalence of $0.5$, and Label $3$ had a prevalence of $0.3$. For each combination of training and test distributions, we generated ten test scenarios by taking different samples.}
\label{tab:mcdists}
\end{table}

Both in the binary and the multiclass settings, we considered relative training versus test data splits in
\[
\{ (0.1,0.9), (0.3,0.7),(0.5,0.5),(0.7,0.3)\},
\]
thereby simulating scenarios in which we have little as well as relatively much information at hand to train our models.
We omitted splits with 90\% training data to save computational resources, as the computational complexity of most algorithms in our experiments is determined by the size of training data rather than test data.
Even without this particular split, in the binary setting we obtain 288 combinations of training distributions, test distributions, and training/test-splits, and in the multiclass setting we obtained 60 of such combinations for each dataset.

To collect experimental data from each dataset which satisfies these constraints, we use undersampling, sampling from a given dataset as many data instances as possible without replacement.
We illustrate this sampling strategy with an example: Assume a dataset with 1000 instances and a binary class attribute, consisting of 700 positive and 300 negative instances. As an example evaluation scenario, we aim to sample data with an 80\%/20\%  split in training and test sets, and with a 60\%/40\% distribution of positive and negative instances in both training and test sets.
Splitting the 300 negative instances randomly 80/20, we have $0.8 \cdot 300 = 240$ negative instances available for training and $0.2 \cdot 300 = 60$ instances available for the test data.
To obtain a 60/40 distribution of positives and negatives in the training data, we therefore have to choose $240 : \frac{40}{60} = 360$ positive instances to include in the training data, which we randomly sample from the full set of positive instances. 
The positives for the test data are randomly sampled analogously.
Note that the instance count for each label imposes a constraint on the number of sampled instances with other labels.
Overall, we use the maximum number of instances for each label that satisfies all constraints.

This sampling procedure can lead to a relatively small subset compared to the whole corpus in cases where the class distributions we aim to sample strongly deviates from the natural class distributions in the given dataset. 
This makes the quantification task comparatively more challenging in these settings.

To address possible variances in the drawn samples, we made ten independent draws for each combination of distributions that could occur within our protocol, and ran all corresponding algorithms on each of these draws. To ensure the reproducibility of all these draws, we used a set of ten fixed seeds for the random number generators.
In total, for the binary setting, we therefore performed 2880 draws per dataset, which, considering that we applied 24 algorithms on 40 datasets, yielded 2,764,800 experiments for that setting.
Including 204,000 additional experiments in the multiclass case and 2,666,520 more experiments on tuned alternative base classifiers (cf. Section \ref{sec:tuning}), 
we conducted for this paper a total number of more than 5 million experiments in our evaluation.

\subsection{Algorithms and Parameter Settings}
In our experiments, we compared all algorithms described in Section \ref{sec:algorithms} and listed in overview Table \ref{tab:algs}.
Except for the \SVMperf-based quantifiers and \emph{Quantification Forests}, all algorithms were implemented from scratch in Python 3, using \texttt{scikit-learn} as base implementation for the underlying classifiers, and the package \texttt{cvxpy} \citep{diamond2016cvxpy} to solve constrained optimization problems. 
For the \SVMperf algorithm, we used the corresponding open source software package by \citet{joachims_support_2005}, and adapted the code that \citet{esuli_recurrent_2018} have used as a baseline for their \emph{QuaNet} method to connect Joachim's C++ implementation to Python.
Regarding \emph{Quantification Forests}, we used the original implementation that was kindly provided by the authors \citep{milli_quantification_2013}.

We further compared and validated our code against the \texttt{QuaPy} package \citep{moreo_quapy_2021}, which implements a subset of the methods that are considered in this evaluation, and has been released after the initial publication of our preprint.
The results are presented in Appendix \ref{ap:quapy}.

As the focus of this work is on a general comparison of quantification algorithms, we initially fixed for all algorithms a set of default parameters based on which the main experiments were conducted.
When choosing the hyperparameters of each model, we followed recommendations of the original papers where possible.
For all quantification methods that required a base classifier, we used the same logistic regression classifier for each dataset split.
The logistic regression model was chosen since it is one of the most established and popular base classifiers
and also actively models its outputs as class probability scores that are required for quantification methods such as the \emph{PAC}, \emph{EM}, or \emph{FM} methods.
That way, results from different quantifiers could not be biased by differences in the underlying classification performances.
We acknowledge that fine-tuning the hyperparameters of each quantifier for each dataset could overall improve the performance, but argue that fixing parameters once allows for a fairer comparison of individual approaches and makes larger numbers of algorithm runs computationally feasible.
However, since one could suspect a strong dependence of the quantification performances from the performance of the underlying classifiers, we further conducted a series of experiments in which we used stronger classifiers with tuned parameters, see Sections 4.3.2 and Section 5.3.
Further, we also explore the impact of parameter tuning within our case study on the dataset from the LeQua challenge (cf. Section \ref{sec:lequa_tuned}).
Below, we will first outline the parameter settings for the main experiments before giving details on the experiments in which we used tuned classifiers.

\subsubsection{Parameter Settings in the Main Experiments}\label{sec:param_main}
For the main experiments, we made the following choices regarding the algorithms' hyperparameters:
\begin{itemize}

    \item As mentioned above, for all methods that utilize a classifier to perform quantification, we used the logistic regression classifier with the default \texttt{L-BFGS} solver along with its built-in probability estimator as provided by \texttt{scikit-learn} and set the number of maximum iterations to 1000. We always used stratified 10-fold cross-validation on the training set when estimating the misclassification rates or computing the set of scores and thresholds that were needed by the quantifiers. 
    
    \item In all adaptations of the \emph{Adjusted Count} that apply threshold selection policies, namely the \emph{TSX}, \emph{TS50}, \emph{TSMax} and \emph{MS} methods, we reduced the sets of scores and thresholds obtained from cross-validation by rounding to three decimals. Additionally, in the \emph{MS} algorithm, we followed Forman's recommendation to only use models that yield a value of at least 0.25 in the denominator of Equation \ref{eq:AC}.
	
    \item For the \emph{DyS} framework, including the \emph{HDy} method, we chose to divide its confidence scores into 10 bins, as this number of bins appeared to yield consistently strong results in the study by \citet{maletzke_dys_2019}. 
	
    \item For the \emph{EM} algorithm and the \emph{CDE} iterators, we chose $\varepsilon = 10^{-6}$ as the convergence parameter, and limited the number of iterations to a maximum of $m = 1000$ iterations, which was reached only very rarely.
	
	\item For the \emph{readme} algorithm, we set the size of each feature subset to $\lfloor \log_2(D)+1\rfloor$, with $D$ denoting the number of features in $X$. We considered an ensemble of 50 subsets, which were all drawn uniformly.
	
	\item In the \emph{QF} and \emph{QF-AC} algorithms, we used the \texttt{weka}-based implementation that has kindly been provided by Letizia Milli \citep{milli_quantification_2013}. We left all parameters at their default values, including the forest size of 100 trees.
	
	\item For both the \emph{SVM-Q} and the \emph{SVM-K} method, we chose $C=1$ as the regularization coefficient, which was however decreased to $C = 0.1$ when there were more than 10,000 training samples. This adaptation was chosen as in our experiments we observed that when large amounts of training data were present, a higher regularization parameter would slow down the convergence of the optimization significantly. 
	
	\item For the \emph{PWK} algorithm, we chose a neighborhood size of $k=10$, and a weighting parameter of $\alpha=1$, as different weight values did not yield significantly better results in the study by \citet{barranquero_quantification-oriented_2015}.
	
	\item In the rare case that in one-vs.-rest quantification, all individual class prevalences were predicted as 0, we returned the uniform distribution as prevalence estimation.
\end{itemize}

\subsubsection{Parameter Settings in the Experiments on Tuned Classifiers}\label{sec:setting_tuned}
Many quantification methods rely on the predictions of an underlying base classifier to form their class prevalence estimations. Since the quality of these underlying classification models could have a strong impact on the performance of the quantifier, we evaluated the impact of applying more advanced classification methods with tuned hyperparameters in our second set of experiments.
For that purpose, we conducted experiments with four classification models, namely random forests, AdaBoost, RBF-kernel support vector machines, and logistic regression models.
For each of these classifiers, we conducted a grid search to optimize the hyperparameters on every single dataset split in our experiments.
Due to scalability issues, we restricted ourselves to the 24 datasets which have not more than 10,000 instances in total.
After having determined their optimal parameter configuration for each dataset split, we used each of the four classification models with their optimal parameterization as base classifiers for the quantification methods.

For the \emph{CC}, \emph{AC}, \emph{GAC}, and \emph{HDy} methods, all four classification models could be applied, as these only require the pure (mis)classification rates from the training data for their estimations.
For all quantifiers which require scores from a classifier's decision function, namely the threshold selection policies \emph{TS50}, \emph{TSX}, \emph{TSMax}, and \emph{MS} as well as the \emph{DyS} and \emph{FMM} method, we only used the support vector classifier and the logistic regression model, since AdaBoost and random forests do not actively model such decision functions.
Furthermore, for all quantifiers which require probability scores, we only applied the tuned logistic regression model, because it is the only method for which the outputs are modeled to represent probabilities.

Regarding the grid search protocol, we applied standard 5-fold cross-validation on the training data---test data was not considered for tuning---when tuning classifiers both in the binary and the multiclass setting, and determined the optimal parameterization based on the accuracy of the resulting classifiers.
Given that in the multiclass case, many quantification methods apply the one-vs.-rest-approach to generalize to this setting, and thus use $L$ different binary quantifiers that each build on a binary classifier, we further applied a second protocol to accommodate this setting.
Specifically, for each parameter configuration in the given grid, we trained $L$ binary classifiers---one one-vs.-rest classifier for each class.
For each class-wise classifier, we computed the \emph{balanced accuracy}, i.e., the average of the true positive rate and the true negative rate in the given binary prediction settings---see also Appendix \ref{ap:clf_performance}, Equations \ref{eq:balanced_accuracy} and \ref{eq:bal_acc_binary} for formal definitions. 
For the one-vs.-rest quantification, we then used $L$ differently parameterized base classifiers, always applying the parameters which yielded the best balanced accuracy in the corresponding one-vs.-rest classification. 

Regarding the parameterization of all quantifiers and base classifiers in this experiment, we made the following choices:

\begin{itemize}
    \item All parameters of the quantification methods which do not regard the underlying classifiers were kept as described in Section \ref{sec:param_main}.
    \item For the logistic regression classifier, in the grid search we varied the regularization weight $C$ within the set $\{2^i: i\in \{-15,-13,-11,\dots,13,15\}\}$. Further, for all values of $C$, we varied the weighting strategy for the instances, by either setting the weights of all instances to 1, or weighting the instances inversely proportional to the prevalence of their corresponding class. Like in previous experiments, we applied the \texttt{L-BFGS} solver to efficiently learn the corresponding models, and set the number of maximum iterations to 1000.
    \item For the random forest, we varied the maximum numbers of features considered per tree among the values $\{2^i: i\in \{1,2,\dots,11\}\}$, and the minimum number of samples per tree, which we considered as the main parameter to control the tree size, within the set $\{2^i: i\in \{1,2,\dots,7\}\}$. Regarding the forest size, we kept a fixed high number of 1000 trees, since it is well-established that choosing a high number of trees yields more reliable results than any lower number of trees.
    \item In the support vector classifier, we varied the regularization weight $C\in\{2^i: i\in \{-5,-3,-1,\dots,13,15\}\}$ and the kernel parameter $\gamma\in \{2^i: i\in \{-17,-15,-13,\dots,3,5\}\}$.
    \item Finally, for the AdaBoost classifier, there is a well-established trade-off between the number of classifiers and the learning rate. Therefore, we only varied the learning rate $\alpha\in\{2^i: i\in \{-19,-17,-13,\dots,1,3\}\}$ and set the number of weak classifiers to a medium amount of 100. 
\end{itemize}

In addition to these experiments with tuned base classifiers, we further ran experiments on the same datasets with variants of the \emph{SVM-K} and \emph{SVM-Q} methods, which applied an RBF-kernel instead of the default linear kernel. 
Since these methods by design optimize for quantification-oriented loss functions, we did not perform any classification-oriented parameter tuning on these and thus, these methods in principle would not fit into this set of experiments.
Yet, given that these RBF-kernel based variants are very expensive computationally, we could not incorporate these in our main experiments where the size of the datasets was not restricted to 10,000 instances. 
For these variants, we chose $C=1$ as the regularization coefficient and $\gamma = 1$ as kernel parameter. 

\subsection{Evaluation}
Next, we describe the error measures that we used in our evaluation, as well as the procedure used to rank the quantifiers and determine statistically significant differences in the performance of the algorithms we have compared. 

\subsubsection{Error Measures for Quantification}
The choice of performance measures for quantification is in itself not a trivial issue, and for a thorough review and discussion of existing quantification measures we point to a recent survey by \citet{sebastiani_evaluation_2020}.
To evaluate the quantification performances in our experiments, we focus here on the \textit{absolute error (AE)} and the \textit{normalized Kullback-Leibler divergence (NKLD)}.
In the following, we let $p\in\Delta^{L-1}$ denote the true distribution of labels $Y$ in an unseen test set, and $\hat{p}\in\Delta^{L-1}$ denote the distribution of labels $Y$ that has been predicted from a given quantifier on the test set, with $\Delta^{L-1}$ denoting the probability simplex.
The absolute error between the true distribution $p$ and an estimated distribution $\hat{p}$ is then given by
\[
 e_{AE}(p,\hat{p}) := \sum_{i=1}^L |p_i - \hat{p}_i|\ ,
\]
whereas the normalized Kullback-Leibler divergence between $p$ and $\hat{p}$ is defined as
\[
e_\text{NKLD}(p,\hat{p}) := 2\cdot\frac{\exp\left\{e_\text{KLD}(p,\hat{p})\right\}}{1 + \exp\left\{e_\text{KLD}(p,\hat{p})\right\}}  - 1\ ,
\]
with
\[
   e_\text{KLD}(p,\hat{p}) := \sum_{i=1}^L p_i \log\left(\frac{p_i}{\hat{p}_i}\right)\  
\]
denoting the Kullback-Leibler divergence. Since the Kullback-Leibler divergence is not defined when $\hat{p}_i = 0$ and $p_i \not= 0$ for some $i\leq L$, we smoothed the distributions by a small value $\varepsilon=10^{-8}$ to avoid this problem.

We chose the AE score due to its interpretability as well as its robustness against outliers.
In contrast to related studies as conducted by \cite{gonzalez-castro_class_2013}, we do not use the \emph{Mean Absolute Error}, i.e., we do not divide by the number $L$ of predicted classes. 
This avoids having different upper bounds for the error depending on $L$, which may make the resulting values harder to interpret, specifically when the number of classes is high, such as in the LeQua case study where $L=28$.
The NKLD score---in contrast to the AE score---particularly punishes quantifiers which marginalize the minority class. 
Both measures are bounded to the same interval for both binary and multiclass quantification, with both values obtaining their minimum (and optimal value) at 0, and the maximum of the AE score being 2, while the maximum NKLD value is 1.

\begin{figure}[t]
	\centering
	
	\begin{subfigure}[b]{0.495\textwidth}
		\centering
		\includegraphics[width=\textwidth]{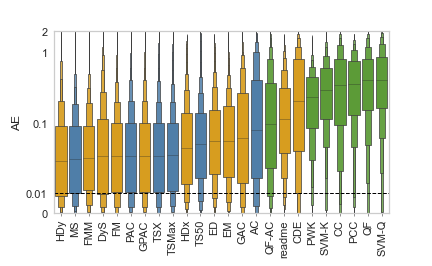}
		\caption{Distribution of absolute errors (AE)}
		\label{fig:gen_boxes_AE}
	\end{subfigure}
	\hfill
	\begin{subfigure}[b]{0.495\textwidth}
		\centering
		\includegraphics[width=\textwidth]{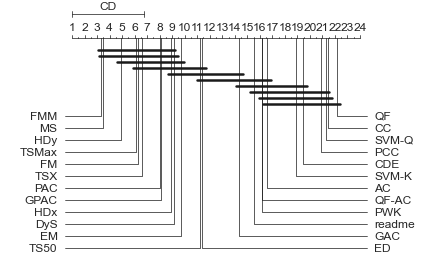}
		\caption{Average rankings with respect to the AE}
		\label{fig:gen_CDs_AE}
	\end{subfigure}
\hfill
	\begin{subfigure}[b]{0.495\textwidth}
		\centering
		\includegraphics[width=\textwidth]{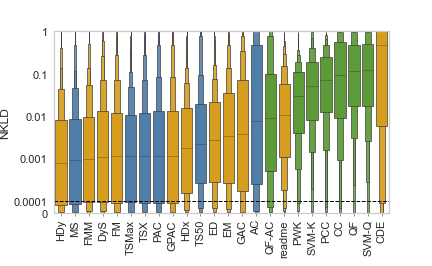}
		\caption{Distribution of NKLD scores}
		\label{fig:gen_boxes_NKLD}
	\end{subfigure}
	\hfill
	\begin{subfigure}[b]{0.495\textwidth}
		\centering
		\includegraphics[width=\textwidth]{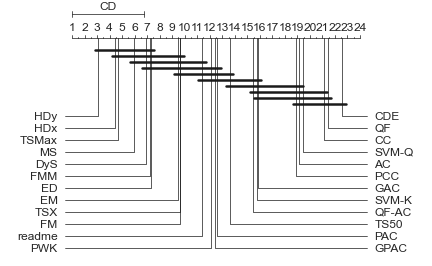}
				\caption{Average rankings with respect to the NKLD}
		\label{fig:gen_CDs_NKLD}
	\end{subfigure}
	\caption{Visual representation of the main results for binary quantification.
	The top row shows results for the absolute error (AE), the bottom row for normalized Kullback-Leibler divergence (NKLD) scores. On the left, letter-value plots for the distribution of error score across all scenarios per algorithm are shown. Colors indicate the category of the algorithm, with count adaptation-based algorithms shown in blue, distribution matching methods in orange, and adaptations of traditional classification algorithms in green. Plots are scaled logarithmically above the dotted vertical threshold, and linearly below. On the right, we plot the distributions of rankings with a Nemenyi post-hoc test at 5\% significance. For each algorithm, we depict the average performance rank over all algorithms. Horizontal bars indicate which average rankings do not differ to a degree that is statistically significant. The critical difference (CD) was  $5.6973$.
    Overall, the \emph{HDy}, \emph{MS}, \emph{FMM}, and \emph{DyS} methods appear to work best in general.}
	\label{fig:gen_binary}
\end{figure}

\subsubsection{Statistical Evaluation of Performance Rankings}
Regarding the actual comparison of the given quantifiers, we adapted a statistical procedure established by \citet{demsar_statistical_2006}, who, in the context of classification, suggested to conduct comparisons of multiple algorithms by statistical tests in a two-step approach that is based on the performance rankings of all algorithms considered with respect to a number of datasets they were applied on.

Within that two-step approach, at first a Friedman test \citep{friedman_comparison_1940} is conducted on the null hypothesis that all algorithms perform equally well over a given ensemble of datasets with respect to a chosen error measure. If that null hypothesis is rejected, one may follow up with the Nemenyi post-hoc test \citep{nemenyi_distribution_1963} to compare the performance rankings of each algorithm per dataset with each other, and determine which algorithms differ from each other in a statistically significant way. The margin of statistical significance is modeled by the critical distance value, which is determined by both the number of algorithms and datasets that are considered as well as the chosen significance level $\alpha$.

While in classification, the underlying rankings would usually be obtained based on a cross-validated accuracy score, in our context we averaged the quantification errors obtained from all the settings in our protocol over each dataset. Based on these average errors, for each dataset, we then determined a ranking of our algorithms for this dataset. 
To account for outliers, we also averaged the resulting scores via the mean and not the median value, which by design of our error measures became more noticeable when averaging with respect to the NKLD errors.

\section{Results}\label{sec:results}

This section presents the results of our extensive experimental evaluation for binary quantification (i.e., labels with exactly two values) and multiclass quantification (i.e., labels with more than two values).
For both types, we start by first showing the main results, which aggregate the performance of each algorithm over all datasets and settings. Then, we show more detailed results for subtypes of scenarios, i.e., for different shifts (differences between training and test distributions) and for varying amounts of training data. Finally, we compare the performance of all presented algorithms in the multiclass case, which is a setting that has not received much attention yet.

\subsection{Binary Quantification}\label{sec:res_binary}
We first describe our results for binary quantification, that is, quantification with binary class labels.

\subsubsection{Overall Results}
\begin{table}
\begin{subtable} {1.0 \linewidth}
    \resizebox{\linewidth}{!}{
        \begin{tabular}{l|cccccccccccccccccccccccc}
\toprule
{} &              AC &    PAC &    TSX &   TS50 &           TSMax &              MS &             GAC &   GPAC &    DyS &             FMM &  readme &             HDx &             HDy &              FM &     ED &              EM &    CDE &              CC &    PCC &    PWK &  SVM-K &           SVM-Q &     QF &           QF-AC \\
\midrule
adult   &           0.042 &  0.022 &  0.018 &  0.029 &           0.018 &           0.018 &           0.041 &  0.022 &  0.017 &  \textbf{0.013} &   0.032 &            0.02 &           0.014 &           0.018 &  0.024 &           0.017 &  0.225 &           0.467 &  0.443 &  0.272 &  0.447 &           0.528 &  0.570 &           0.118 \\
avila   &            0.56 &  0.086 &  0.079 &  0.081 &           0.071 &           0.069 &           0.459 &  0.086 &  0.186 &           0.066 &   0.086 &  \textbf{0.045} &           0.074 &           0.096 &  0.075 &           0.214 &  0.899 &           0.852 &  0.682 &  0.286 &  0.765 &           0.849 &  0.678 &           0.327 \\
bike    &           0.036 &  0.023 &  0.021 &  0.033 &           0.022 &           0.018 &           0.036 &  0.023 &  0.017 &           0.015 &   0.079 &           0.034 &  \textbf{0.014} &           0.021 &  0.073 &           0.044 &  0.096 &            0.29 &  0.309 &  0.281 &  0.209 &           0.363 &  0.498 &           0.065 \\
blog    &           0.072 &  0.036 &  0.034 &  0.034 &           0.031 &            0.03 &           0.072 &  0.036 &  0.030 &           0.029 &   0.042 &  \textbf{0.024} &           0.029 &           0.033 &  0.055 &           0.042 &  0.569 &           0.643 &  0.575 &  0.394 &  0.387 &           0.654 &  0.625 &           0.213 \\
bc-cat  &            0.23 &  0.112 &  0.077 &  0.137 &           0.079 &  \textbf{0.055} &           0.193 &  0.112 &  0.121 &           0.056 &   0.276 &           0.109 &           0.083 &           0.062 &  0.093 &           0.207 &  0.315 &            0.38 &  0.390 &  0.091 &  0.304 &           0.753 &  0.315 &           0.151 \\
bc-cont &           0.133 &  0.072 &  0.051 &  0.130 &           0.049 &           0.042 &           0.117 &  0.072 &  0.106 &           0.048 &   0.121 &           0.103 &           0.056 &  \textbf{0.039} &  0.052 &           0.125 &  0.123 &           0.172 &  0.245 &  0.058 &  0.167 &           0.838 &  0.249 &           0.133 \\
cars    &            0.13 &  0.080 &  0.063 &  0.110 &            0.06 &  \textbf{0.049} &           0.113 &  0.080 &  0.078 &           0.051 &   0.229 &           0.101 &           0.059 &           0.059 &  0.154 &           0.087 &  0.180 &           0.299 &  0.306 &  0.229 &  0.228 &           0.227 &  0.485 &           0.296 \\
conc    &           0.533 &  0.171 &  0.154 &  0.190 &           0.144 &  \textbf{0.121} &           0.369 &  0.171 &  0.175 &           0.125 &   0.258 &           0.172 &           0.178 &           0.155 &  0.184 &           0.336 &  0.745 &           0.699 &  0.608 &  0.300 &  0.304 &           0.601 &  0.627 &           0.275 \\
contra  &           0.613 &  0.332 &  0.351 &  0.371 &           0.326 &           0.307 &           0.472 &  0.331 &  0.434 &           0.297 &   0.366 &           0.284 &             0.4 &           0.351 &  0.408 &  \textbf{0.249} &  0.881 &           0.814 &  0.672 &  0.526 &  0.565 &           0.802 &  0.664 &            0.47 \\
cappl   &           0.323 &  0.155 &  0.127 &  0.200 &           0.128 &           0.104 &           0.289 &  0.156 &  0.205 &           0.109 &   0.374 &           0.233 &           0.172 &           0.115 &  0.222 &  \textbf{0.087} &  0.302 &           0.473 &  0.465 &  0.257 &  0.330 &           0.322 &  0.514 &           0.292 \\
ccard   &           0.312 &  0.061 &  0.066 &  0.054 &           0.054 &  \textbf{0.044} &            0.28 &  0.061 &  0.055 &            0.05 &   0.283 &           0.061 &           0.048 &           0.064 &  0.090 &           0.062 &  0.847 &           0.753 &  0.641 &  0.412 &  0.496 &            0.69 &  0.615 &            0.33 \\
diam    &           0.315 &  0.037 &  0.032 &  0.045 &           0.032 &           0.031 &            0.27 &  0.038 &  0.037 &           0.027 &   0.063 &  \textbf{0.021} &           0.027 &           0.031 &  0.085 &           0.217 &  0.746 &           0.709 &  0.609 &  0.323 &  0.627 &           0.853 &  0.497 &            0.27 \\
dota    &           1.074 &  0.048 &  0.054 &  0.054 &           0.056 &           0.056 &           0.397 &  0.048 &  0.063 &  \textbf{0.047} &   0.360 &           0.211 &           0.048 &           0.053 &  0.189 &            0.13 &  0.864 &           0.835 &  0.680 &  0.557 &  0.587 &           0.806 &  0.886 &            0.69 \\
drugs   &           0.168 &  0.118 &  0.102 &  0.115 &           0.106 &           0.088 &           0.174 &  0.119 &  0.144 &  \textbf{0.080} &   0.163 &           0.124 &           0.101 &           0.104 &  0.114 &           0.134 &  0.134 &           0.421 &  0.428 &  0.275 &  0.318 &           0.337 &  0.504 &           0.181 \\
ener    &           0.271 &  0.040 &  0.040 &  0.048 &           0.041 &           0.037 &           0.224 &  0.040 &  0.041 &  \textbf{0.032} &   0.207 &           0.073 &           0.034 &           0.041 &  0.067 &            0.12 &  0.699 &           0.672 &  0.596 &  0.270 &  0.399 &           0.742 &  0.741 &             0.5 \\
fifa    &            0.76 &  0.035 &  0.030 &  0.036 &           0.029 &           0.028 &           0.055 &  0.035 &  0.027 &           0.023 &   0.137 &           0.025 &  \textbf{0.022} &            0.03 &  0.040 &           0.031 &  0.204 &           0.461 &  0.447 &  0.329 &  1.240 &           1.188 &  0.418 &           0.056 \\
flare   &           0.584 &  0.344 &  0.353 &  0.345 &           0.306 &           0.269 &           0.482 &  0.342 &  0.454 &           0.291 &   0.316 &           0.267 &           0.416 &           0.346 &  0.314 &  \textbf{0.256} &  0.675 &           0.694 &  0.629 &  0.405 &  0.480 &           0.614 &  0.668 &           0.347 \\
grid    &            0.09 &  0.046 &  0.046 &  0.052 &           0.052 &           0.038 &           0.086 &  0.046 &  0.042 &           0.035 &   0.149 &            0.07 &  \textbf{0.033} &           0.044 &  0.058 &           0.048 &  0.258 &           0.492 &  0.468 &  0.225 &  0.749 &           0.668 &  0.782 &            0.51 \\
ads     &           0.175 &  0.103 &  0.075 &  0.113 &           0.067 &  \textbf{0.054} &           0.138 &  0.102 &  0.106 &            0.06 &   0.225 &           0.144 &           0.077 &           0.082 &  0.195 &           0.087 &  0.199 &           0.352 &  0.352 &  0.389 &  0.255 &           0.341 &  0.434 &           0.317 \\
magic   &           0.271 &  0.043 &  0.042 &  0.052 &           0.045 &           0.041 &           0.236 &  0.043 &  0.043 &           0.039 &   0.103 &           0.056 &  \textbf{0.038} &           0.044 &  0.044 &           0.057 &  0.469 &           0.606 &  0.542 &  0.314 &  0.607 &           0.587 &  0.661 &           0.477 \\
boone   &           0.013 &  0.009 &  0.007 &  0.015 &           0.007 &           0.009 &           0.013 &  0.009 &  0.007 &           0.007 &   0.087 &           0.008 &  \textbf{0.006} &           0.007 &  0.011 &           0.024 &  0.069 &           0.282 &  0.307 &  0.133 &  0.257 &           0.693 &  0.505 &            0.27 \\
mush    &           0.014 &  0.011 &  0.008 &  0.048 &           0.009 &  \textbf{0.007} &           0.014 &  0.011 &  0.014 &           0.016 &   0.033 &           0.018 &  \textbf{0.007} &           0.008 &  0.027 &           0.017 &  0.009 &           0.027 &  0.054 &  0.009 &  0.098 &           0.054 &  0.215 &           0.021 \\
music   &           0.547 &  0.324 &  0.327 &  0.346 &           0.299 &           0.272 &           0.462 &  0.324 &  0.429 &           0.283 &   0.682 &           0.479 &           0.371 &           0.328 &  0.404 &  \textbf{0.257} &  0.840 &           0.748 &  0.651 &  0.449 &  0.465 &           0.572 &  0.741 &           0.609 \\
musk    &            0.11 &  0.070 &  0.067 &  0.080 &           0.068 &           0.058 &           0.096 &  0.069 &  0.073 &  \textbf{0.053} &   0.434 &           0.117 &           0.058 &           0.068 &  0.126 &           0.065 &  0.188 &           0.367 &  0.379 &  0.276 &  0.248 &           0.321 &  0.489 &           0.177 \\
news    &           0.346 &  0.052 &  0.053 &  0.053 &           0.057 &           0.048 &           0.243 &  0.052 &  0.057 &  \textbf{0.046} &   0.433 &           0.087 &            0.05 &           0.053 &  0.089 &           0.058 &  0.866 &           0.772 &  0.651 &  0.470 &  0.475 &           0.842 &  0.842 &            0.57 \\
nurse   &  \textbf{0.000} &  0.002 &  0.007 &  0.351 &           0.007 &           0.061 &  \textbf{0.000} &  0.002 &  0.004 &           0.123 &   0.448 &           0.008 &           0.001 &  \textbf{0.000} &  0.024 &            0.02 &  0.002 &  \textbf{0.000} &  0.024 &  0.128 &  0.001 &  \textbf{0.000} &  0.065 &           0.005 \\
occup   &            0.04 &  0.017 &  0.006 &  0.057 &  \textbf{0.005} &           0.006 &           0.034 &  0.017 &  0.007 &           0.021 &   0.020 &            0.01 &  \textbf{0.005} &           0.006 &  0.012 &           0.103 &  0.098 &           0.125 &  0.192 &  0.015 &  0.241 &           0.531 &  0.113 &           0.012 \\
phish   &           0.821 &  0.023 &  0.020 &  0.037 &           0.021 &           0.018 &           0.029 &  0.023 &  0.020 &           0.016 &   0.058 &           0.021 &           0.015 &           0.019 &  0.033 &  \textbf{0.014} &  0.026 &           0.188 &  0.212 &  0.137 &  0.188 &           0.153 &  0.364 &           0.042 \\
craft   &           0.248 &  0.084 &  0.065 &  0.088 &           0.075 &           0.058 &           0.219 &  0.084 &  0.082 &  \textbf{0.053} &   0.211 &           0.069 &           0.058 &           0.067 &  0.070 &           0.144 &  0.528 &           0.602 &  0.543 &  0.319 &  0.344 &           0.684 &  0.568 &           0.222 \\
spam    &           0.274 &  0.069 &  0.047 &  0.071 &            0.05 &           0.043 &           0.236 &  0.069 &  0.072 &  \textbf{0.041} &   0.177 &           0.168 &           0.042 &           0.047 &  0.082 &           0.265 &  0.603 &           0.595 &  0.537 &  0.204 &  0.261 &           0.638 &  0.667 &           0.298 \\
alco    &            0.48 &  0.328 &  0.341 &  0.366 &             0.3 &  \textbf{0.277} &           0.451 &  0.337 &  0.431 &           0.282 &   0.584 &           0.415 &            0.36 &           0.342 &  0.468 &           0.296 &  0.695 &           0.693 &  0.625 &  0.491 &  0.495 &           0.608 &  0.653 &           0.495 \\
study   &           0.347 &  0.187 &  0.201 &  0.215 &           0.194 &           0.161 &           0.301 &  0.187 &  0.233 &           0.162 &   0.287 &           0.151 &           0.194 &           0.192 &  0.308 &           0.175 &  0.533 &           0.589 &  0.538 &  0.330 &  0.610 &           0.696 &  0.460 &  \textbf{0.145} \\
cond    &            0.04 &  0.018 &  0.017 &  0.034 &           0.017 &           0.015 &            0.04 &  0.018 &  0.015 &           0.014 &   0.090 &           0.017 &  \textbf{0.013} &           0.017 &  0.019 &           0.022 &  0.097 &           0.319 &  0.317 &  0.124 &  0.206 &           0.287 &  0.399 &           0.069 \\
telco   &           0.224 &  0.075 &  0.071 &  0.080 &           0.069 &            0.06 &           0.211 &  0.075 &  0.075 &  \textbf{0.056} &   0.097 &  \textbf{0.056} &           0.059 &            0.07 &  0.065 &           0.059 &  0.401 &           0.571 &  0.525 &  0.387 &  0.373 &           0.476 &  0.609 &           0.304 \\
thrm    &           0.612 &  0.318 &  0.320 &  0.355 &           0.298 &           0.272 &           0.462 &  0.318 &  0.423 &           0.291 &   0.534 &           0.348 &           0.358 &           0.309 &  0.355 &  \textbf{0.266} &  0.861 &           0.773 &  0.655 &  0.444 &  0.491 &           0.629 &  0.698 &           0.495 \\
turk    &           0.619 &  0.248 &  0.282 &  0.283 &            0.24 &           0.239 &           0.477 &  0.246 &  0.303 &           0.219 &   0.351 &           0.258 &           0.281 &            0.28 &  0.211 &  \textbf{0.164} &  0.881 &           0.847 &  0.684 &  0.529 &  0.558 &            0.64 &  0.844 &           0.702 \\
vgame   &           0.209 &  0.085 &  0.088 &  0.086 &           0.091 &           0.076 &           0.209 &  0.085 &  0.090 &           0.075 &   0.201 &            0.13 &           0.084 &           0.089 &  0.266 &  \textbf{0.066} &  0.586 &           0.631 &  0.570 &  0.400 &  0.407 &           0.594 &  0.654 &           0.302 \\
voice   &            0.15 &  0.048 &  0.035 &  0.060 &           0.032 &           0.034 &           0.134 &  0.047 &  0.037 &           0.038 &   0.210 &           0.045 &  \textbf{0.030} &           0.036 &  0.060 &           0.178 &  0.289 &           0.346 &  0.378 &  0.076 &  0.166 &           0.417 &  0.246 &           0.051 \\
wine    &           0.479 &  0.095 &  0.091 &  0.093 &           0.096 &           0.081 &           0.372 &  0.095 &  0.140 &  \textbf{0.079} &   0.198 &           0.123 &           0.096 &           0.102 &  0.162 &           0.233 &  0.815 &            0.75 &  0.637 &  0.350 &  0.662 &           0.905 &  0.649 &           0.319 \\
yeast   &           0.681 &  0.238 &  0.276 &  0.306 &           0.234 &  \textbf{0.212} &           0.471 &  0.241 &  0.338 &           0.221 &   0.343 &           0.386 &           0.273 &           0.261 &  0.246 &            0.38 &  0.873 &           0.839 &  0.680 &  0.428 &  0.569 &           0.881 &  0.672 &            0.45 \\\midrule
Mean    &           0.324 &  0.107 &  0.104 &  0.131 &           0.097 &  \textbf{0.088} &           0.224 &  0.107 &  0.131 &            0.09 &   0.234 &           0.127 &           0.107 &           0.102 &  0.139 &           0.134 &  0.467 &           0.529 &  0.481 &  0.297 &  0.414 &           0.585 &  0.547 &            0.29 \\
\bottomrule
\end{tabular}

    }
\subcaption{Absolute error (AE)}
\label{tab:main_results_ae}
\end{subtable}

\begin{subtable} {1.0 \linewidth}
    \resizebox{\linewidth}{!}{
        \begin{tabular}{l|cccccccccccccccccccccccc}
\toprule
{} &              AC &             PAC &             TSX &   TS50 &           TSMax &     MS &             GAC &            GPAC &             DyS &             FMM &  readme &             HDx &             HDy &              FM &              ED &              EM &    CDE &              CC &    PCC &             PWK &           SVM-K &           SVM-Q &     QF &  QF-AC \\
\midrule
adult   &           0.018 &           0.005 &           0.003 &  0.005 &           0.002 &  0.002 &           0.017 &           0.005 &  \textbf{0.001} &  \textbf{0.001} &   0.003 &           0.002 &  \textbf{0.001} &           0.002 &           0.003 &  \textbf{0.001} &  0.200 &           0.175 &  0.139 &           0.061 &           0.129 &           0.184 &  0.294 &  0.052 \\
avila   &           0.513 &           0.061 &           0.037 &  0.031 &           0.022 &  0.026 &           0.256 &            0.06 &           0.097 &           0.026 &   0.013 &  \textbf{0.005} &           0.015 &           0.056 &           0.016 &           0.156 &  0.850 &           0.642 &  0.292 &           0.063 &           0.335 &           0.593 &  0.395 &  0.238 \\
bike    &           0.013 &           0.007 &           0.005 &  0.007 &           0.003 &  0.003 &           0.013 &           0.007 &           0.002 &           0.003 &   0.012 &           0.004 &  \textbf{0.001} &           0.005 &           0.012 &           0.007 &  0.100 &           0.073 &  0.077 &           0.063 &           0.044 &           0.116 &  0.215 &  0.013 \\
blog    &           0.035 &            0.01 &           0.009 &  0.008 &           0.005 &  0.006 &           0.035 &           0.009 &           0.004 &           0.006 &   0.004 &  \textbf{0.002} &  \textbf{0.002} &           0.009 &           0.007 &           0.004 &  0.575 &           0.305 &  0.214 &           0.103 &           0.102 &           0.333 &  0.375 &  0.074 \\
bc-cat  &           0.161 &           0.065 &           0.024 &  0.088 &           0.017 &  0.015 &           0.089 &           0.065 &           0.038 &           0.016 &   0.075 &           0.022 &           0.018 &           0.023 &           0.017 &            0.16 &  0.409 &           0.182 &  0.123 &  \textbf{0.013} &            0.08 &           0.316 &  0.083 &  0.033 \\
bc-cont &           0.084 &            0.04 &           0.013 &  0.081 &            0.01 &  0.019 &           0.052 &            0.04 &           0.022 &           0.024 &   0.022 &           0.017 &           0.007 &           0.015 &           0.008 &           0.087 &  0.184 &           0.067 &  0.060 &  \textbf{0.006} &           0.035 &           0.447 &  0.061 &  0.028 \\
cars    &           0.074 &           0.051 &           0.028 &  0.057 &           0.016 &  0.019 &           0.051 &           0.049 &           0.021 &           0.019 &   0.053 &           0.018 &  \textbf{0.013} &            0.03 &           0.032 &           0.034 &  0.212 &           0.099 &  0.083 &           0.046 &           0.051 &           0.045 &  0.317 &  0.265 \\
conc    &           0.459 &            0.13 &           0.089 &  0.125 &           0.052 &  0.060 &           0.156 &            0.13 &           0.077 &           0.067 &   0.066 &  \textbf{0.039} &            0.07 &           0.091 &           0.045 &           0.325 &  0.799 &           0.495 &  0.245 &           0.072 &           0.074 &           0.306 &  0.324 &  0.132 \\
contra  &           0.537 &           0.247 &           0.258 &  0.271 &           0.175 &  0.172 &           0.242 &           0.247 &           0.245 &           0.199 &   0.107 &  \textbf{0.085} &           0.203 &            0.26 &            0.16 &           0.125 &  0.843 &           0.581 &  0.286 &           0.166 &           0.197 &           0.382 &  0.392 &  0.286 \\
cappl   &           0.238 &           0.093 &           0.061 &  0.128 &  \textbf{0.036} &  0.040 &           0.156 &           0.095 &           0.075 &           0.045 &   0.110 &           0.057 &           0.057 &           0.054 &           0.053 &           0.037 &  0.415 &           0.244 &  0.159 &           0.056 &           0.093 &           0.086 &  0.192 &  0.082 \\
ccard   &           0.206 &           0.029 &           0.026 &  0.021 &            0.01 &  0.012 &           0.164 &           0.029 &           0.008 &           0.015 &   0.074 &           0.008 &  \textbf{0.006} &           0.023 &           0.015 &           0.015 &  0.835 &           0.447 &  0.260 &           0.109 &            0.16 &           0.386 &  0.267 &  0.093 \\
diam    &           0.285 &           0.015 &           0.011 &  0.012 &           0.007 &  0.009 &           0.139 &           0.015 &           0.008 &           0.009 &   0.008 &  \textbf{0.002} &           0.003 &           0.011 &           0.013 &           0.185 &  0.811 &           0.474 &  0.244 &           0.082 &           0.309 &           0.665 &  0.322 &  0.261 \\
dota    &           0.877 &           0.015 &           0.018 &  0.019 &           0.013 &  0.013 &             0.2 &           0.015 &           0.013 &           0.015 &   0.104 &           0.049 &  \textbf{0.009} &           0.017 &           0.042 &           0.024 &  0.843 &           0.583 &  0.289 &           0.189 &           0.225 &           0.499 &  0.685 &  0.475 \\
drugs   &           0.093 &           0.057 &           0.041 &  0.059 &           0.025 &  0.031 &           0.093 &           0.057 &           0.037 &           0.028 &   0.032 &           0.022 &  \textbf{0.019} &           0.044 &            0.02 &           0.022 &  0.094 &           0.144 &  0.134 &           0.059 &           0.078 &           0.088 &  0.178 &  0.043 \\
ener    &            0.22 &           0.017 &           0.012 &  0.013 &           0.007 &  0.011 &           0.125 &           0.017 &           0.009 &           0.009 &   0.047 &           0.011 &  \textbf{0.004} &           0.013 &           0.013 &           0.066 &  0.803 &           0.409 &  0.232 &            0.06 &           0.112 &           0.376 &  0.520 &  0.429 \\
fifa    &           0.779 &           0.013 &           0.007 &  0.010 &           0.005 &  0.008 &           0.028 &           0.013 &           0.004 &           0.005 &   0.025 &           0.003 &  \textbf{0.002} &           0.008 &           0.005 &  \textbf{0.002} &  0.256 &           0.164 &  0.141 &           0.081 &           0.981 &           0.953 &  0.132 &  0.007 \\
flare   &           0.436 &           0.247 &           0.251 &  0.259 &           0.152 &  0.151 &           0.296 &           0.244 &           0.217 &           0.178 &   0.087 &           0.082 &           0.192 &           0.234 &           0.097 &  \textbf{0.081} &  0.711 &            0.42 &  0.256 &            0.11 &           0.159 &           0.243 &  0.324 &  0.138 \\
grid    &           0.041 &           0.015 &           0.009 &  0.010 &           0.007 &  0.007 &           0.034 &           0.015 &           0.005 &           0.005 &   0.028 &           0.009 &  \textbf{0.002} &           0.009 &           0.008 &           0.014 &  0.414 &           0.188 &  0.151 &           0.045 &           0.596 &           0.425 &  0.554 &  0.432 \\
ads     &           0.112 &           0.074 &           0.035 &  0.070 &  \textbf{0.016} &  0.021 &           0.078 &           0.074 &           0.033 &           0.024 &   0.052 &           0.027 &           0.018 &           0.039 &           0.044 &           0.027 &  0.187 &           0.134 &  0.108 &           0.134 &           0.071 &           0.107 &  0.152 &  0.101 \\
magic   &           0.252 &           0.009 &           0.007 &  0.009 &           0.006 &  0.007 &           0.101 &           0.009 &           0.006 &           0.006 &   0.016 &           0.007 &  \textbf{0.005} &           0.008 &           0.006 &           0.014 &  0.528 &           0.378 &  0.197 &           0.077 &           0.371 &           0.359 &  0.470 &  0.408 \\
boone   &           0.002 &           0.001 &  \textbf{0.000} &  0.001 &  \textbf{0.000} &  0.001 &           0.002 &           0.001 &  \textbf{0.000} &           0.001 &   0.013 &  \textbf{0.000} &  \textbf{0.000} &  \textbf{0.000} &           0.001 &           0.001 &  0.027 &            0.07 &  0.075 &           0.023 &           0.058 &           0.463 &  0.325 &  0.261 \\
mush    &           0.002 &           0.001 &           0.001 &  0.012 &           0.001 &  0.001 &           0.002 &           0.001 &           0.001 &           0.004 &   0.004 &           0.001 &  \textbf{0.000} &           0.001 &           0.004 &           0.001 &  0.004 &           0.003 &  0.006 &           0.001 &           0.016 &           0.007 &  0.041 &  0.002 \\
music   &           0.435 &           0.248 &           0.223 &  0.242 &           0.147 &  0.142 &           0.258 &           0.248 &           0.207 &           0.172 &   0.291 &           0.174 &           0.168 &           0.224 &           0.134 &  \textbf{0.082} &  0.829 &           0.474 &  0.270 &           0.131 &           0.136 &           0.204 &  0.474 &  0.390 \\
musk    &           0.057 &           0.029 &           0.029 &  0.036 &           0.016 &  0.019 &           0.045 &           0.028 &           0.017 &           0.016 &   0.143 &            0.02 &  \textbf{0.007} &           0.028 &           0.023 &           0.011 &  0.198 &           0.116 &  0.109 &           0.062 &           0.049 &           0.087 &  0.187 &  0.042 \\
news    &           0.254 &           0.016 &           0.015 &  0.016 &           0.011 &  0.009 &           0.107 &           0.015 &            0.01 &           0.011 &   0.139 &           0.013 &  \textbf{0.006} &           0.015 &           0.013 &  \textbf{0.006} &  0.843 &           0.484 &  0.268 &           0.134 &           0.144 &           0.466 &  0.625 &  0.480 \\
nurse   &  \textbf{0.000} &  \textbf{0.000} &           0.006 &  0.325 &           0.006 &  0.044 &  \textbf{0.000} &  \textbf{0.000} &  \textbf{0.000} &           0.107 &   0.129 &  \textbf{0.000} &  \textbf{0.000} &  \textbf{0.000} &           0.002 &           0.005 &  0.010 &  \textbf{0.000} &  0.002 &           0.017 &  \textbf{0.000} &  \textbf{0.000} &  0.017 &  0.001 \\
occup   &           0.022 &           0.005 &  \textbf{0.000} &  0.025 &  \textbf{0.000} &  0.001 &           0.016 &           0.005 &  \textbf{0.000} &           0.006 &   0.001 &  \textbf{0.000} &  \textbf{0.000} &           0.001 &           0.001 &           0.054 &  0.128 &           0.034 &  0.041 &           0.001 &           0.066 &             0.3 &  0.017 &  0.001 \\
phish   &           0.451 &           0.004 &           0.004 &  0.007 &           0.003 &  0.003 &           0.007 &           0.004 &           0.002 &           0.003 &   0.007 &           0.001 &  \textbf{0.000} &           0.004 &           0.004 &  \textbf{0.000} &  0.002 &           0.034 &  0.041 &           0.022 &           0.032 &           0.027 &  0.108 &  0.013 \\
craft   &           0.179 &           0.049 &            0.02 &  0.042 &           0.014 &  0.020 &           0.106 &           0.049 &           0.021 &           0.018 &   0.048 &            0.01 &  \textbf{0.008} &           0.027 &           0.013 &           0.089 &  0.733 &           0.318 &  0.199 &           0.076 &            0.09 &           0.306 &  0.311 &  0.151 \\
spam    &            0.22 &           0.036 &           0.011 &  0.031 &           0.009 &  0.012 &           0.121 &           0.036 &           0.025 &           0.011 &   0.036 &           0.032 &  \textbf{0.004} &           0.013 &           0.012 &           0.218 &  0.718 &           0.351 &  0.200 &            0.04 &           0.061 &           0.298 &  0.299 &  0.082 \\
alco    &           0.365 &           0.254 &           0.259 &  0.280 &           0.155 &  0.159 &           0.279 &            0.26 &           0.207 &           0.192 &   0.225 &           0.137 &           0.176 &           0.262 &           0.164 &  \textbf{0.102} &  0.783 &           0.392 &  0.254 &           0.147 &           0.167 &           0.238 &  0.319 &  0.200 \\
study   &           0.264 &           0.115 &           0.106 &  0.129 &           0.071 &  0.069 &           0.145 &           0.115 &           0.095 &           0.078 &   0.072 &  \textbf{0.030} &           0.075 &           0.103 &           0.088 &           0.084 &  0.689 &           0.337 &  0.202 &           0.082 &           0.213 &           0.283 &  0.166 &  0.047 \\
cond    &           0.017 &           0.003 &           0.002 &  0.008 &  \textbf{0.001} &  0.002 &           0.017 &           0.003 &           0.002 &           0.003 &   0.014 &           0.002 &  \textbf{0.001} &           0.003 &           0.003 &           0.003 &  0.148 &           0.101 &  0.082 &           0.021 &           0.043 &           0.069 &  0.170 &  0.030 \\
telco   &           0.152 &           0.038 &           0.032 &  0.035 &           0.015 &  0.024 &            0.12 &            0.04 &           0.016 &           0.021 &   0.015 &           0.011 &           0.011 &           0.032 &           0.012 &  \textbf{0.007} &  0.532 &           0.284 &  0.186 &           0.104 &           0.099 &           0.151 &  0.379 &  0.274 \\
thrm    &           0.505 &           0.252 &           0.235 &  0.267 &           0.169 &  0.174 &           0.224 &           0.251 &           0.222 &             0.2 &   0.200 &  \textbf{0.110} &           0.183 &           0.221 &           0.129 &           0.191 &  0.837 &           0.534 &  0.275 &           0.129 &           0.164 &           0.295 &  0.401 &  0.269 \\
turk    &           0.527 &           0.194 &           0.197 &  0.206 &           0.113 &  0.112 &           0.247 &           0.192 &           0.133 &           0.138 &   0.104 &           0.081 &           0.109 &           0.207 &           0.082 &  \textbf{0.048} &  0.843 &           0.613 &  0.292 &           0.163 &           0.215 &           0.294 &  0.633 &  0.538 \\
vgame   &           0.152 &           0.045 &           0.038 &  0.036 &           0.026 &  0.028 &           0.131 &           0.045 &            0.02 &            0.03 &   0.043 &           0.023 &           0.019 &            0.04 &            0.07 &  \textbf{0.013} &  0.763 &           0.323 &  0.215 &           0.106 &           0.114 &           0.267 &  0.350 &  0.099 \\
voice   &           0.107 &           0.025 &           0.013 &  0.021 &           0.006 &  0.010 &           0.067 &           0.024 &           0.006 &           0.014 &   0.048 &           0.004 &  \textbf{0.002} &           0.014 &            0.01 &           0.121 &  0.467 &           0.153 &  0.113 &           0.009 &           0.032 &           0.183 &  0.052 &  0.006 \\
wine    &           0.419 &            0.05 &           0.039 &  0.036 &           0.024 &  0.026 &           0.164 &           0.049 &           0.057 &           0.032 &   0.041 &           0.021 &  \textbf{0.020} &           0.048 &           0.059 &           0.211 &  0.831 &           0.524 &  0.262 &           0.089 &           0.248 &           0.513 &  0.350 &  0.174 \\
yeast   &           0.595 &            0.18 &           0.198 &  0.225 &           0.111 &  0.115 &             0.2 &           0.183 &           0.179 &           0.133 &   0.096 &            0.12 &           0.115 &            0.19 &  \textbf{0.079} &           0.373 &  0.842 &           0.652 &  0.291 &           0.121 &           0.228 &           0.636 &  0.393 &  0.263 \\\midrule
Mean    &           0.254 &           0.069 &           0.059 &  0.082 &           0.037 &  0.040 &           0.115 &           0.069 &           0.054 &           0.047 &   0.065 &  \textbf{0.032} &           0.039 &            0.06 &           0.038 &           0.075 &  0.507 &             0.3 &  0.177 &           0.077 &           0.159 &             0.3 &  0.297 &  0.173 \\
\bottomrule
\end{tabular}

    }
\subcaption{NKLD scores}
\label{tab:main_results_nkld}
\end{subtable}
\caption{Main results for \textbf{binary} quantification. We show the averages for absolute errors (AE) and normalized Kullback-Leibler divergence (NKLD) for all scenarios per algorithm and dataset. For absolute errors, \emph{MS} shows the best performance. For NKLD, \emph{HDx} and \emph{HDy} achieve the best results on the plurality of datasets.}
\label{tab:bin_general}
\end{table}

We show the general performance results of all quantification algorithms over all datasets in Figure \ref{fig:gen_binary} and Table \ref{tab:bin_general}.
The letter-value plots in Figures \ref{fig:gen_boxes_AE}) and \ref{fig:gen_boxes_NKLD}) represent the respective distributions of absolute error (AE) and normalized Kullback-Leibler divergence (NKLD) scores resulting from all experiments. Colors in the plot indicate algorithm categories, i.e., count adaptation-based algorithms are shown in blue, distribution matching methods in orange, and adaptations of traditional classification algorithms are shown in green. 
The plots in \ref{fig:gen_CDs_AE}) and \ref{fig:gen_CDs_NKLD}) depict the average performance ranks of all algorithms per dataset along with the critical differences between average ranks, which indicate whether the difference in average ranks is statistically significant according to the Nemenyi post-hoc test \citep{demsar_statistical_2006}. Here, horizontal bars show which average rankings do not differ to a degree that is statistically significant.
Tables \ref{tab:main_results_ae}) and \ref{tab:main_results_nkld}) complement these plots by providing the average absolute errors (AE) and normalized Kullback-Leibler divergences (NKLD) for all scenarios per algorithm and dataset. Based on these averages, the rankings for the plots \ref{fig:gen_CDs_AE}) and \ref{fig:gen_CDs_NKLD}) have been compiled. Further, for each algorithm, a total average error score over all datasets is provided.

Overall, under both the NKLD and the AE metric, we observe substantial differences between the algorithms. While there is no single individual best algorithm for all cases, the results suggest that there is a group of algorithms that perform particularly well compared to the rest. 
First and foremost, the \emph{HDy}, \emph{MS}, \emph{FMM}, \emph{DyS}, and \emph{FM} methods, in that order, appear to yield the best performances when considering the overall distributions of error scores with respect to both AE and NKLD.
When considering the aggregated rankings, these methods also tend to perform well, with the \emph{FMM} and \emph{MS} methods performing the strongest with respect to the AE, and \emph{HDy} performing strongest for the NKLD.
Except for the \emph{FM} method falling off in the NKLD-based rankings, there is, however, no statistically significant difference between these methods with respect to the Nemenyi post-hoc test.

Considering the overall distribution of error scores, the \emph{PAC} and \emph{GPAC} methods also appear to yield relatively robust performance over all datasets, but with respect to the NKLD, these methods are significantly worse in their average rankings than the top-ranking \emph{HDy} method.
Further, the \emph{TSMax} method also appears among the top performing methods in the aggregated rankings, and the \emph{HDx} method appears particularly strong in the NKLD-based rankings, despite not standing out in the overall error distributions.

These general impressions are confirmed by Table \ref{tab:bin_general}, where we see that the \emph{FMM} and \emph{HDy} algorithm take the top rank on most datasets with respect to the AE score, whereas for the NKLD score, the \emph{HDy} method is most dominant in these rankings.
Considering the overall means in these tables, it is further notable that the \emph{MS} method has the overall lowest average error with respect to AE, and \emph{HDx} the lowest mean error with respect to NKLD, indicating a relatively high robustness against outliers.

When considering the performance of basic algorithms such as \emph{(Probabilistic) Classify and Count} and \emph{Adjusted Count}, we observe that these baselines are clearly outperformed by the top algorithms. 
Moreover, all algorithms that we have categorized as \textit{classifiers for quantification}, and also the \emph{CDE iterator} consistently show the worst performances with respect to both measures.

\begin{figure}
	\centering
	\begin{subfigure}[b]{0.49\textwidth}
		\centering
		\includegraphics[width=\textwidth]{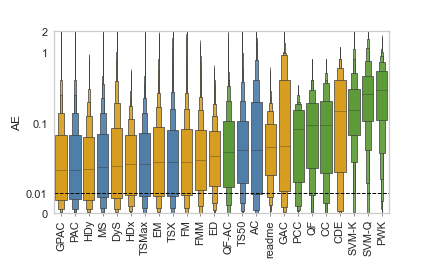}
		\caption{AE values under minor shift}
	\end{subfigure}
	\hfill
	\begin{subfigure}[b]{0.49\textwidth}
		\centering
		\includegraphics[width=\textwidth]{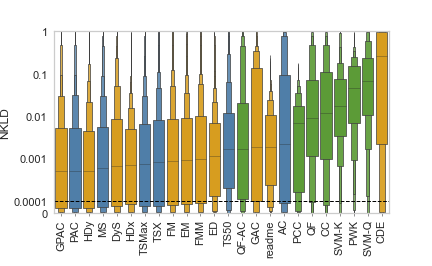}
		\caption{NKLD values under minor shift}
	\end{subfigure}
	\begin{subfigure}[b]{0.49\textwidth}
		\centering
		\includegraphics[width=\textwidth]{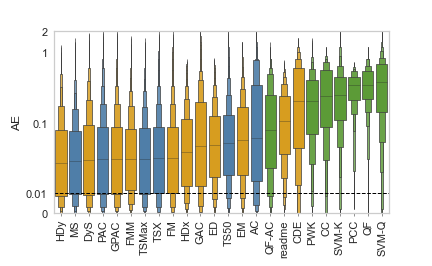}
		\caption{AE values under medium shift}
	\end{subfigure}
	\hfill
	\begin{subfigure}[b]{0.49\textwidth}
		\centering
		\includegraphics[width=\textwidth]{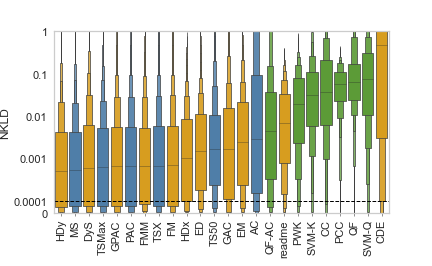}
		\caption{NKLD values under medium shift}
	\end{subfigure}
	\begin{subfigure}[b]{0.49\textwidth}
		\centering
		\includegraphics[width=\textwidth]{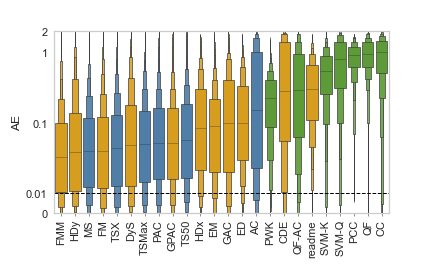}
		\caption{AE values under major shift}
	\end{subfigure}
	\hfill
	\begin{subfigure}[b]{0.49\textwidth}
		\centering
		\includegraphics[width=\textwidth]{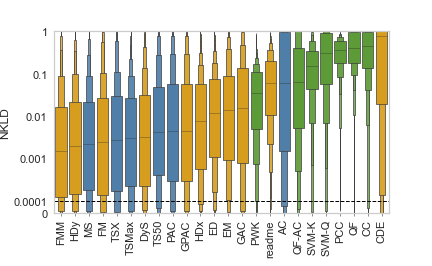}
		\caption{NKLD values under major shift}
	\end{subfigure}
	\caption{Impact of distribution shift. We show the distribution of error scores split by the amount of shift in the evaluation scenario. The left column shows results according to the absolute error, the right one according to NKLD scores.  Colors indicate the category of the algorithm. Plots are scaled logarithmically above the dotted vertical threshold, and linearly below. \emph{GPAC} appears to perform best under minor shifts, \emph{FMM} under major shifts.}
	\label{fig:shift}
\end{figure}

\subsubsection{Influence of Distribution Shift}

In the context of quantification, a shift of the distribution of the class labels $Y$ between training and test set is assumed. 
It could be expected that the size of distribution shift affects the difficulty of the quantification task, as we assume that stronger shifts make accurate quantification more challenging.
For that reason, we now take a closer look at the impact of this distribution shift to find out which methods are more or less sensitive to the severity of a distribution shift. 
In that context, we categorize all settings into three scenarios, namely a minor shift, a medium shift, and a major shift in these distributions. 
More precisely, we consider the shift to be
\begin{itemize}
    \item minor, if the distribution shift is lower than 0.4 in $L_1$ distance,
    \item medium, if the distribution shift is bigger or equal to  0.4 and lower than 0.8 in $L_1$ distance,
    \item major, if the distribution shift is bigger or equal to  0.8 in $L_1$ distance.
\end{itemize}

We depict the aggregated performance of the quantification algorithms under these three kinds of shifts in Figure \ref{fig:shift}.
Unsurprisingly, we can observe that the performance of all quantification algorithms generally deteriorates with an increasing shift in class distributions.
In that regard, the effect appears to be the strongest for classification-based approaches, in particular for the \emph{Quantification Forests} and the \emph{PCC} method.
The only exception to this principle appears to be the nearest-neighbor-based \emph{PWK} quantifier, which with respect to the NKLD error scores appears relatively robust toward distribution shift.
Further, the \emph{readme}, \emph{PAC} and \emph{GPAC} methods also appear to be affected strongly by increasing distribution shift, which is exemplified by the drop in their average rankings per dataset (cf. Appendix \ref{ap:main}, Figure \ref{fig:cd_shift}). 
Contrarily, the \emph{HDy} and \emph{FMM} methods appear the most robust to larger shifts.

For all other algorithms, except for the relatively robust \emph{PWK} method, the decrease in performance appears to be between the aforementioned robust algorithms and the classify and count-based quantifiers, with their overall rankings appearing mostly unaffected from distribution shift.
That implies that even though the overall performance deteriorates, the same methods perform well, independent of the amount of shift.

\begin{figure}
	\centering
	\begin{subfigure}[b]{0.49\textwidth}
		\centering
		\includegraphics[width=\textwidth]{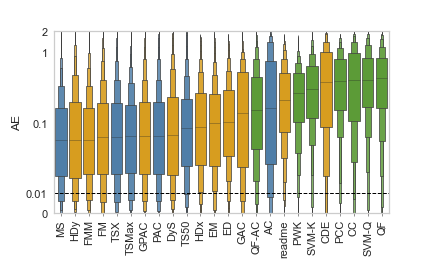}
		\caption{AE values for few training samples}
	\end{subfigure}
	\hfill
	\begin{subfigure}[b]{0.49\textwidth}
		\centering
		\includegraphics[width=\textwidth]{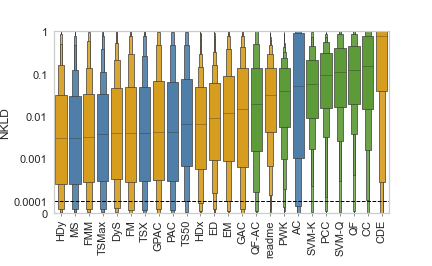}
		\caption{NKLD values for few training samples}
	\end{subfigure}
	\caption{Performance with small amounts of training data. Plots are scaled logarithmically above the dotted vertical threshold, and linearly below.  Colors indicate the category of the algorithm. We observe similar trends compared to the general setting, with \emph{MS}, \emph{HDy}, and \emph{FMM} being among the best-performing algorithms.}
	\label{fig:lowtrain}
\end{figure}

\subsubsection{Influence of Training Set Size}

Next, we consider the performance of quantification algorithms when relatively few training samples are given. For that purpose, we restrict the experimental data to only those cases in which the given data was split into 10\% training and 90\% test samples.
The overall distribution of error scores with respect to the AE and NKLD scores can be found in Figure \ref{fig:lowtrain}.
We observe that in general, the performance of all algorithms seems to be worse compared to the results when not being restricting to little training data, which is also to be expected intuitively.
However, again the methods which yielded the overall best performances, such as \emph{MS}, \emph{HDy}, and \emph{FMM}, also appear the most robust toward this scenario.
The average performance rankings of all algorithms per dataset (cf. Appendix \ref{ap:main}, Figure \ref{fig:cd_fewtrain}) are mostly in line with the general setting.


\begin{figure}
	\centering
	
	\begin{subfigure}[b]{0.5\textwidth}
		\centering
		\includegraphics[width=\textwidth]{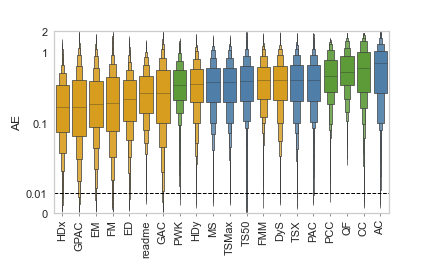}
		\caption{Distribution of absolute errors (AE)}
		\label{fig:mc_gen_boxes_AE}
	\end{subfigure}
	\hfill
	\begin{subfigure}[b]{0.49\textwidth}
		\centering
		\includegraphics[width=\textwidth]{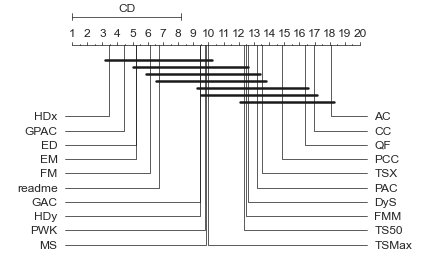}
		\caption{Average rankings with respect to the AE}
		\label{fig:mc_gen_CDs_AE}
	\end{subfigure}
\hfill
	\begin{subfigure}[b]{0.5\textwidth}
		\centering
		\includegraphics[width=\textwidth]{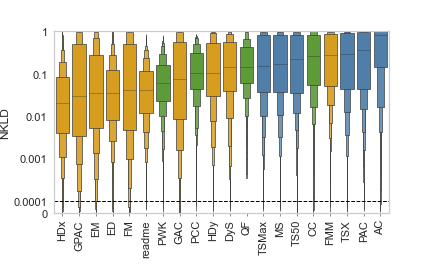}
		\caption{Distribution of NKLD scores}
		\label{fig:mc_gen_boxes_NKLD}
	\end{subfigure}
	\hfill
	\begin{subfigure}[b]{0.49\textwidth}
		\centering
		\includegraphics[width=\textwidth]{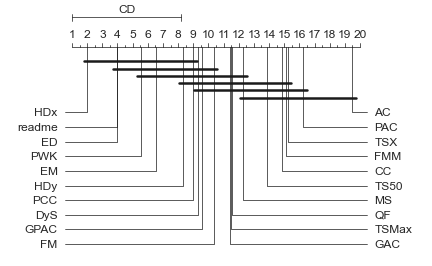}
				\caption{Average rankings with respect to NKLD}
		\label{fig:mc_gen_CDs_NKLD}
	\end{subfigure}
	\caption{Visual representation of the main results for \emph{multiclass} quantification.
	The top row shows results for the absolute error (AE), the bottom row for normalized Kullback-Leibler divergence (NKLD) scores. On the left, letter-value plots for the distribution of error score across all scenarios per algorithm are shown, colors indicate the category of the algorithm. Plots are scaled logarithmically above the dotted vertical threshold, and linearly below. On the right, we plot the distributions of rankings with a Nemenyi post-hoc test at 5\% significance. Horizontal bars indicate which average rankings do not differ to a degree that is statistically significant. The critical difference (CD) was  $7.0045$.
	Overall, performance scores are much worse than in the binary setting. 
    Best performances are generally achieved by distribution matching methods that naturally extend to the multiclass setting, with the \emph{HDx} method standing out.}
	\label{fig:gen_mc}
\end{figure}

\begin{table}[t]
\begin{subtable} {1.0 \linewidth}
	\resizebox{\linewidth}{!}{
        \begin{tabular}{l|cccccccccccccccccccc}
\toprule
{} &     AC &    PAC &    TSX &   TS50 &  TSMax &              MS &    GAC &            GPAC &    DyS &    FMM &  readme &             HDx &    HDy &              FM &              ED &              EM &     CC &    PCC &    PWK &     QF \\
\midrule
bike   &  0.675 &  0.469 &  0.426 &  0.397 &  0.455 &           0.465 &  0.113 &  \textbf{0.073} &  0.465 &  0.461 &   0.201 &           0.126 &  0.454 &           0.102 &           0.176 &           0.082 &  0.368 &  0.364 &  0.315 &  0.638 \\
blog   &  0.795 &  0.671 &  0.594 &  0.585 &  0.565 &           0.557 &  0.360 &           0.236 &  0.533 &  0.580 &   0.180 &  \textbf{0.148} &  0.541 &           0.285 &            0.29 &           0.196 &  0.588 &  0.500 &  0.422 &  0.547 \\
conc   &  0.864 &  0.574 &  0.615 &  0.591 &  0.502 &           0.508 &  0.486 &           0.473 &  0.562 &  0.564 &   0.432 &  \textbf{0.380} &  0.536 &            0.51 &           0.457 &           0.498 &  0.915 &  0.692 &  0.480 &  0.662 \\
contra &  0.829 &  0.483 &  0.496 &  0.508 &  0.466 &           0.462 &  0.600 &           0.515 &  0.538 &  0.467 &   0.424 &  \textbf{0.338} &  0.481 &           0.512 &           0.434 &           0.396 &  0.833 &  0.699 &  0.572 &  0.675 \\
diam   &  0.399 &  0.232 &  0.272 &  0.251 &  0.244 &           0.241 &  0.197 &           0.098 &  0.251 &  0.254 &   0.117 &  \textbf{0.044} &  0.207 &           0.118 &           0.209 &           0.214 &  0.784 &  0.645 &  0.404 &  0.501 \\
drugs  &  0.228 &  0.166 &  0.170 &  0.177 &  0.171 &  \textbf{0.147} &  0.256 &           0.199 &  0.213 &  0.160 &   0.338 &           0.203 &  0.180 &           0.181 &           0.238 &           0.218 &  0.465 &  0.482 &  0.407 &  0.600 \\
ener   &  0.634 &  0.354 &  0.354 &  0.322 &  0.351 &           0.346 &  0.273 &  \textbf{0.115} &  0.337 &  0.366 &   0.331 &           0.178 &  0.347 &           0.129 &           0.169 &           0.131 &  0.879 &  0.699 &  0.439 &  0.925 \\
fifa   &  0.838 &  0.656 &  0.616 &  0.615 &  0.567 &           0.564 &  0.313 &           0.181 &  0.581 &  0.599 &   0.221 &  \textbf{0.126} &  0.525 &           0.216 &           0.278 &           0.127 &  0.481 &  0.441 &  0.384 &  0.432 \\
news   &  0.825 &  0.581 &  0.548 &  0.541 &  0.522 &           0.523 &  0.498 &           0.335 &  0.535 &  0.545 &   0.446 &           0.237 &  0.522 &           0.376 &           0.245 &  \textbf{0.221} &  0.827 &  0.614 &  0.471 &  0.917 \\
nurse  &  0.077 &  0.104 &  0.064 &  0.159 &  0.068 &           0.082 &  0.023 &  \textbf{0.019} &  0.047 &  0.203 &   0.263 &           0.034 &  0.047 &            0.02 &           0.049 &           0.022 &  0.138 &  0.173 &  0.213 &  0.399 \\
craft  &  0.560 &  0.525 &  0.515 &  0.488 &  0.474 &           0.464 &  0.296 &  \textbf{0.190} &  0.494 &  0.531 &   0.412 &           0.228 &  0.475 &  \textbf{0.190} &           0.274 &           0.191 &  0.752 &  0.654 &  0.442 &  0.763 \\
cond   &  0.541 &  0.442 &  0.479 &  0.353 &  0.500 &           0.485 &  0.155 &           0.066 &  0.456 &  0.516 &   0.129 &           0.077 &  0.469 &           0.088 &           0.093 &  \textbf{0.059} &  0.343 &  0.362 &  0.213 &  0.431 \\
thrm   &  1.297 &  0.633 &  0.726 &  0.684 &  0.593 &           0.587 &  0.780 &           0.629 &  0.694 &  0.619 &   0.471 &  \textbf{0.441} &  0.634 &           0.663 &            0.47 &           0.494 &  1.042 &  0.769 &  0.511 &  0.827 \\
turk   &  0.651 &  0.326 &  0.375 &  0.392 &  0.349 &           0.348 &  0.525 &           0.342 &  0.455 &  0.324 &   0.489 &           0.421 &  0.372 &           0.392 &           0.356 &  \textbf{0.277} &  0.976 &  0.727 &  0.622 &  0.834 \\
vgame  &  0.741 &  0.640 &  0.630 &  0.626 &  0.574 &           0.575 &  0.520 &            0.46 &  0.557 &  0.600 &   0.364 &           0.334 &  0.521 &           0.474 &           0.424 &  \textbf{0.322} &  0.590 &  0.520 &  0.418 &  0.589 \\
wine   &  1.061 &  0.706 &  0.700 &  0.693 &  0.595 &           0.607 &  0.656 &           0.575 &  0.719 &  0.637 &   0.428 &  \textbf{0.416} &  0.546 &           0.605 &            0.44 &           0.757 &  0.965 &  0.636 &  0.496 &  0.613 \\
yeast  &  1.015 &  0.541 &  0.518 &  0.487 &  0.446 &           0.464 &  0.567 &           0.408 &  0.527 &  0.505 &   0.474 &           0.342 &  0.412 &           0.413 &  \textbf{0.289} &           0.613 &  0.878 &  0.612 &  0.295 &  0.526 \\\midrule
Mean   &  0.708 &  0.477 &  0.476 &  0.463 &  0.438 &           0.437 &  0.389 &           0.289 &  0.468 &  0.466 &   0.336 &  \textbf{0.240} &  0.428 &            0.31 &           0.288 &           0.284 &  0.696 &  0.564 &  0.418 &  0.640 \\
\bottomrule
\end{tabular}

    }
	\subcaption{Average AE values of each algorithm per dataset}
	\label{tab:mc_general_AE}
\end{subtable}

\begin{subtable} {1.0 \linewidth}
	\resizebox{\linewidth}{!}{
        \begin{tabular}{l|cccccccccccccccccccc}
\toprule
{} &     AC &    PAC &    TSX &   TS50 &  TSMax &     MS &    GAC &            GPAC &    DyS &    FMM &          readme &             HDx &    HDy &     FM &     ED &              EM &     CC &    PCC &             PWK &     QF \\
\midrule
bike   &  0.657 &  0.378 &  0.296 &  0.331 &  0.305 &  0.303 &  0.045 &  \textbf{0.016} &  0.266 &  0.351 &            0.05 &           0.026 &  0.282 &  0.032 &  0.045 &  \textbf{0.016} &  0.116 &  0.105 &           0.092 &  0.244 \\
blog   &  0.707 &  0.822 &  0.658 &  0.656 &  0.642 &  0.648 &  0.402 &           0.201 &  0.463 &  0.673 &            0.04 &  \textbf{0.031} &  0.565 &  0.243 &  0.113 &           0.044 &  0.315 &  0.155 &           0.135 &  0.206 \\
conc   &  0.841 &  0.443 &  0.439 &  0.410 &  0.362 &  0.393 &  0.310 &           0.467 &  0.304 &  0.407 &  \textbf{0.126} &           0.129 &  0.275 &  0.455 &  0.211 &            0.46 &  0.640 &  0.276 &           0.137 &  0.254 \\
contra &  0.662 &  0.425 &  0.412 &  0.433 &  0.333 &  0.350 &  0.448 &           0.469 &  0.312 &  0.395 &           0.131 &  \textbf{0.123} &  0.275 &  0.445 &  0.214 &           0.237 &  0.464 &  0.280 &           0.179 &  0.258 \\
diam   &  0.472 &  0.161 &  0.186 &  0.176 &  0.161 &  0.159 &  0.103 &           0.062 &  0.160 &  0.167 &           0.016 &  \textbf{0.003} &  0.143 &  0.092 &  0.091 &            0.17 &  0.531 &  0.254 &           0.096 &  0.225 \\
drugs  &  0.164 &  0.100 &  0.125 &  0.091 &  0.074 &  0.087 &  0.180 &            0.15 &  0.069 &  0.108 &           0.085 &  \textbf{0.039} &  0.046 &  0.126 &  0.053 &           0.049 &  0.151 &  0.147 &           0.112 &  0.204 \\
ener   &  0.598 &  0.383 &  0.366 &  0.327 &  0.330 &  0.337 &  0.137 &           0.085 &  0.222 &  0.390 &           0.086 &  \textbf{0.041} &  0.264 &  0.084 &  0.050 &           0.087 &  0.491 &  0.270 &            0.12 &  0.527 \\
fifa   &  0.761 &  0.790 &  0.660 &  0.594 &  0.621 &  0.623 &  0.316 &           0.115 &  0.476 &  0.652 &           0.049 &  \textbf{0.024} &  0.489 &  0.152 &  0.099 &           0.029 &  0.254 &  0.126 &           0.115 &  0.129 \\
news   &  0.751 &  0.456 &  0.398 &  0.396 &  0.358 &  0.363 &  0.539 &           0.318 &  0.316 &  0.389 &           0.143 &           0.068 &  0.337 &  0.400 &  0.076 &  \textbf{0.059} &  0.524 &  0.227 &            0.16 &  0.608 \\
nurse  &  0.060 &  0.063 &  0.008 &  0.018 &  0.007 &  0.038 &  0.011 &           0.005 &  0.003 &  0.189 &           0.055 &           0.002 &  0.002 &  0.007 &  0.005 &  \textbf{0.001} &  0.025 &  0.033 &           0.049 &  0.115 \\
craft  &  0.502 &  0.457 &  0.423 &  0.377 &  0.420 &  0.416 &  0.172 &            0.15 &  0.222 &  0.438 &           0.113 &  \textbf{0.052} &  0.218 &  0.117 &  0.080 &           0.159 &  0.398 &  0.242 &           0.113 &  0.403 \\
cond   &  0.652 &  0.525 &  0.515 &  0.382 &  0.496 &  0.493 &  0.089 &           0.011 &  0.301 &  0.524 &           0.022 &           0.009 &  0.330 &  0.027 &  0.018 &  \textbf{0.004} &  0.166 &  0.098 &           0.044 &  0.130 \\
thrm   &  0.969 &  0.608 &  0.729 &  0.706 &  0.530 &  0.533 &  0.605 &           0.648 &  0.517 &  0.641 &  \textbf{0.145} &           0.214 &  0.502 &  0.723 &  0.248 &           0.442 &  0.692 &  0.340 &           0.151 &  0.382 \\
turk   &  0.580 &  0.320 &  0.377 &  0.396 &  0.260 &  0.259 &  0.412 &           0.347 &  0.274 &  0.295 &           0.176 &           0.254 &  0.193 &  0.372 &  0.177 &  \textbf{0.105} &  0.585 &  0.296 &           0.216 &  0.435 \\
vgame  &  0.717 &  0.620 &  0.555 &  0.515 &  0.485 &  0.492 &  0.584 &           0.522 &  0.364 &  0.548 &           0.102 &  \textbf{0.098} &  0.385 &  0.509 &  0.134 &           0.133 &  0.238 &  0.170 &           0.134 &  0.205 \\
wine   &  0.810 &  0.714 &  0.690 &  0.665 &  0.521 &  0.552 &  0.434 &            0.62 &  0.537 &  0.620 &  \textbf{0.129} &           0.185 &  0.410 &  0.617 &  0.278 &           0.781 &  0.714 &  0.240 &           0.157 &  0.221 \\
yeast  &  0.817 &  0.598 &  0.580 &  0.502 &  0.485 &  0.519 &  0.358 &           0.431 &  0.479 &  0.593 &           0.143 &           0.105 &  0.342 &  0.401 &  0.115 &           0.702 &  0.585 &  0.224 &  \textbf{0.075} &  0.173 \\\midrule
Mean   &  0.631 &  0.463 &  0.436 &  0.410 &  0.376 &  0.386 &  0.303 &           0.272 &  0.311 &  0.434 &           0.095 &  \textbf{0.083} &  0.298 &  0.283 &  0.118 &           0.205 &  0.405 &  0.205 &           0.123 &  0.278 \\
\bottomrule
\end{tabular}

    }
\subcaption{Average NKLD scores for each algorithm per dataset}
	\label{tab:mc_general_NKLD}
\end{subtable}
\label{tab:mc_general}
	\caption{Main results for \textbf{multiclass} quantification. We show the averaged error scores for all scenarios per algorithm and dataset. Overall, distribution matching methods that naturally generalize to the multiclass setting appear to perform better than one-vs.-rest or classify and count-based approaches.}
\end{table}

\subsection{Multiclass Quantification}\label{sec:res_mc}

Next, we present results for multiclass quantification, i.e., quantification for labels with more than two values.

\subsubsection{Overall Results}

Tables \ref{tab:mc_general_AE}) and \ref{tab:mc_general_NKLD}) as well as Figure \ref{fig:gen_mc} present the main results for multiclass quantification.
Compared to the binary case, we obtain substantially different results.
First of all, the overall prediction performance is much worse, as both the AE scores and the NKLD scores appear multiple times higher on average. For instance, AE values below 0.1 and NKLD values below 0.01 are widespread in the binary case, whereas in the multiclass case, such scores are only rarely achieved.
Instead, the average AE values of each algorithm over all experiments are mostly around the interval [0.3,0.4], which is three to four times higher than the average AE values of the best algorithms in the binary case.
The second main difference regards the algorithms which appear to work best:
Algorithms such as the \emph{DyS}-Framework, the \emph{Median Sweep (MS)}, and the other threshold selection policies, which have worked very well for binary quantification, appear comparatively weak in their performance.
By contrast, the best performances seem to be achieved by distribution matching algorithms which also naturally extend to the multiclass setting, namely the \emph{GPAC}, \emph{ED}, \emph{FM}, \emph{EM}, \emph{readme}, and \emph{HDx} methods.
In that context, specifically the \emph{HDx} method stands out.
Further, the \emph{GPAC}, \emph{ED}, \emph{EM}, and \emph{FM} methods show strong performances with respect to the absolute error, whereas the \emph{ED}, \emph{readme}, and \emph{EM}, but also the classification-based \emph{PWK} method obtain the high average rankings with respect to the NKLD. 
These overall trends are also confirmed in Tables \ref{tab:mc_general_AE}) and \ref{tab:mc_general_NKLD}), where the \emph{HDx} method stands out with regard to both AE and NKLD.
In addition, from the overall distributions of errors in Figures \ref{fig:mc_gen_boxes_AE}) and \ref{fig:mc_gen_boxes_NKLD}) it becomes apparent that these algorithms also have strong differences in the variance of their performances.
In particular, the \emph{GPAC} method appears to have a much higher variance in its error scores compared to the rest, whereas the \emph{ED} and \emph{readme} methods display the lowest variance in their performances.
However, the given results also have one big similarity with the results from the binary setting, i.e., all algorithms which are based on the classify and count principle display subpar performances, even when optimizing quantification-based loss functions.

\subsubsection{Impact of Distribution Shift}

As in the binary case, we also investigate the effect that the shift of the distribution of the class labels $Y$ between training and test sets has on the resulting quantification performance. 
Since we have less experimental data than in the binary case, this time we only distinguish between a minor shift and a major shift. We consider the shift to be
\begin{itemize}
    \item minor, if the distribution shift is lower than 0.5 in $L_1$ distance,
    \item major, if the distribution shift is bigger or equal to  0.5 in $L_1$ distance.
\end{itemize}

\begin{figure}[t]
	\centering
	\begin{subfigure}[b]{0.49\textwidth}
		\centering
		\includegraphics[width=\textwidth]{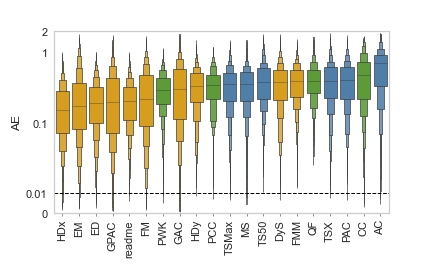}
		\caption{AE values under minor shift}
	\end{subfigure}
	\hfill
	\begin{subfigure}[b]{0.49\textwidth}
		\centering
		\includegraphics[width=\textwidth]{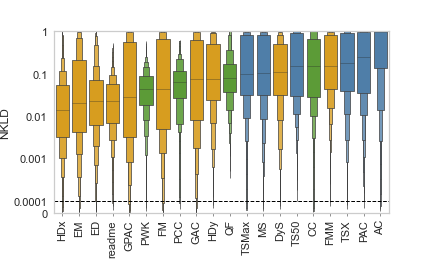}
		\caption{NKLD values under minor shift}
	\end{subfigure}
	\begin{subfigure}[b]{0.49\textwidth}
		\centering
		\includegraphics[width=\textwidth]{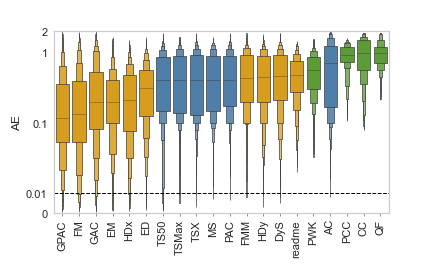}
		\caption{AE values under major shift}
	\end{subfigure}
	\hfill
	\begin{subfigure}[b]{0.49\textwidth}
		\centering
		\includegraphics[width=\textwidth]{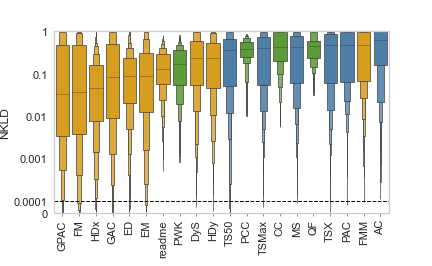}
		\caption{NKLD values under major shift}
	\end{subfigure}
	\caption{Impact of distribution shift in the multiclass setting.  We show the distribution of error scores split by the amount of shift in the evaluation scenario. The left column shows results according to the absolute error, the right one according to NKLD scores. Plots are scaled logarithmically above the dotted vertical threshold, and linearly below. Colors indicate the category of the algorithm. Performances generally deteriorate under major shift. Best performances under major shift are achieved by algorithms that also do work well in general. The \emph{GPAC} and \emph{FM} methods appear most robust toward major shifts.}
	\label{fig:mc_shift}
\end{figure}

The results of multiclass quantification under these scenarios are shown in Figure \ref{fig:mc_shift}. 
Similar to the binary case, we observe that the algorithms which appeared to work best in general, also appear the most robust with respect to high distribution shifts.
In particular, the GPAC method appears almost unaffected by a high shift in its average performance, consistently achieving higher performance ranks under higher shifts, although significant variance is present in its performance.
By contrast, all methods which apply the classify and count principle are again the most susceptible to higher error rates when applied in scenarios with higher shifts between training and test distribution.

\begin{figure}[b]
	\centering
	\begin{subfigure}[b]{0.49\textwidth}
		\centering
		\includegraphics[width=\textwidth]{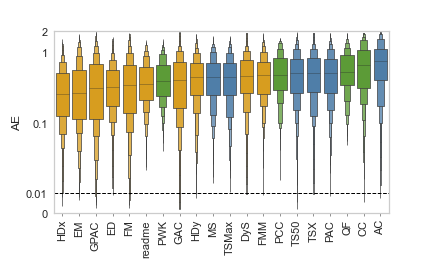}
		\caption{AE values for few training samples}
	\end{subfigure}
	\hfill
	\begin{subfigure}[b]{0.49\textwidth}
		\centering
		\includegraphics[width=\textwidth]{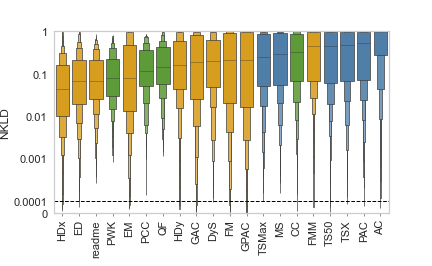}
		\caption{NKLD values for few training samples}
	\end{subfigure}
	\caption{Performance under small amount of training data in multiclass setting. Plots are scaled logarithmically above the dotted vertical threshold, and linearly below.  Colors indicate the category of the algorithm. The results are similar to the general setting, although the overall performance scores are slightly worse, and in particular the \emph{GPAC} method does deteriorate with respect to the NKLD metric.}
	\label{fig:mc_fewtrain}
\end{figure}
\FloatBarrier

\begin{figure}[th]
	\centering
	\begin{subfigure}[b]{\textwidth}
		\centering
		\includegraphics[width=\textwidth]{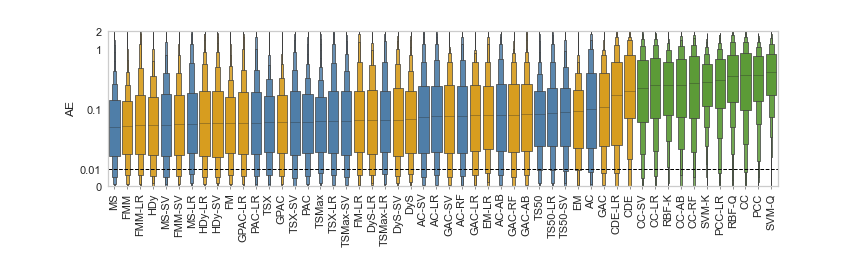}
		\caption{Distribution of absolute error (AE) scores}
	\end{subfigure}
	\hfill
	\begin{subfigure}[b]{\textwidth}
		\centering
		\includegraphics[width=\textwidth]{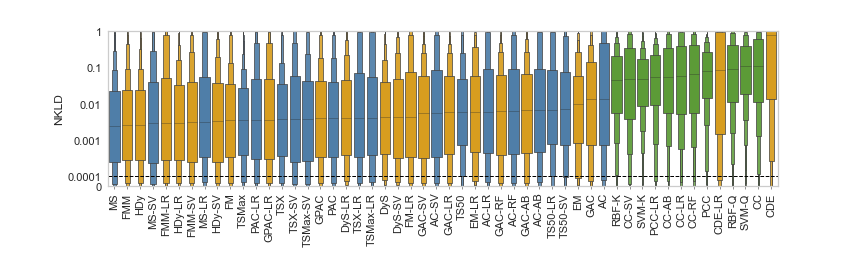}
		\caption{Distribution of NKLD scores}
	\end{subfigure}
	\caption{Results of our experiments in the binary setting, where the base classifiers were tuned with respect to their accuracy. Plots are scaled logarithmically above the dotted vertical threshold, and linearly below.  Colors indicate the category of the algorithm. Algorithms based on untuned logistic regression classifiers are denoted as before (no suffix), alternative tuned base classifiers are marked with respective suffixes: logistic regressors (LR), support vector machines (SV), random forests (RF) and AdaBoost (AB). In addition, we present results of the \emph{RBF-K} and \emph{RBF-Q} methods, which are variants of the \emph{SVM-K} and \emph{SVM-Q} that use an RBF-kernel instead of a linear one.  We observe that except for the \emph{CC}, \emph{PCC}, \emph{GAC} and \emph{CDE} methods, tuning the classifiers does not seem to have a significant positive effect on the outcome.}
	\label{fig:clf_bin}
\end{figure}

\subsubsection{Influence of Training Set Size}

Finally, we consider the performance of the given algorithms when the given data was split into 10\% training and 90\% test samples.
As before, this serves to investigate the impact of having a relatively small set of training data. The distributions of error scores with respect to the AE and NKLD metrics can be found in Figure \ref{fig:mc_fewtrain}.
Compared to the general distribution of  error scores from all experiments, the performance when only small training sets are given does deteriorate. 
In particular, we observe that the \emph{GPAC} is much less competitive than in the general scenario, in particular with respect to the NKLD score.
Conversely, the \emph{HDx}, \emph{EM} and \emph{ED} algorithms, and, with respect to the NKLD score, also the \emph{readme} method appear to be most robust toward this setting---this latter result may be due to \emph{readme} returning an average prediction of an ensemble, which makes it less likely to falsely predict class prevalences of 0 and obtain a high NKLD value in consequence.
This implies that those algorithms could be recommended if only small training sets are available.

\subsection{Impact of  Alternative Classifiers and Tuning}
\label{sec:tuning}

Finally, we present the results from our experiments with quantifiers that applied tuned base classifiers.
We begin with the results on binary data, before closing with the results on multiclass data.

\begin{figure}[t]
	\centering
	\begin{subfigure}[b]{\textwidth}
		\centering
		\includegraphics[width=\textwidth]{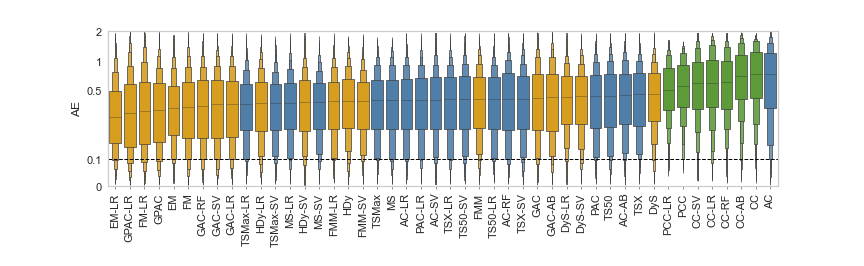}
		\caption{Distribution of absolute error (AE) scores}
	\end{subfigure}
	\hfill
	\begin{subfigure}[b]{\textwidth}
		\centering
		\includegraphics[width=\textwidth]{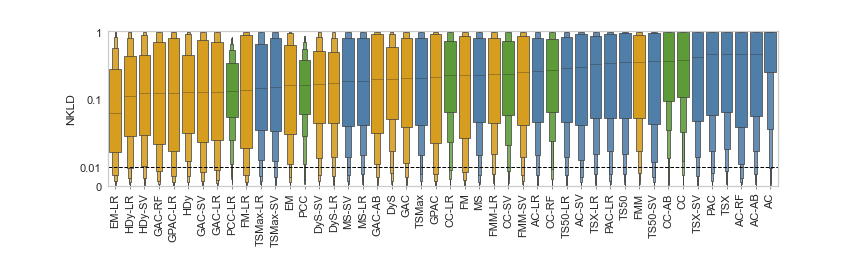}
		\caption{Distribution of NKLD scores}
	\end{subfigure}
	\caption{Results of our experiments with quantifiers that apply tuned classifiers in the multiclass setting.
    For natural multiclass quantifiers, base classifiers were tuned with respect to their accuracy. 
    For one-vs.-rest-based quantifiers, the binary base classifiers were tuned with respect to their \emph{balanced} accuracy.
    Plots are scaled logarithmically above the dotted vertical threshold, and linearly below.  Colors indicate the category of the algorithm. Algorithms based on untuned logistic regression classifiers are denoted as before (no suffix), alternative tuned base classifiers are marked with respective suffixes: logistic regressors (LR), support vector machines (SV), random forests (RF) and AdaBoost (AB).  In contrast to the binary case, we observe mostly positive effects from tuning the base classifiers, and the tuned variant of the \emph{EM} method appears to outperform the other quantifiers.}
	\label{fig:clf_mc}
\end{figure}

\subsubsection{Experiments on Binary Data}

In Figure \ref{fig:clf_bin}, we show the scores of all quantifiers using different tuned base classifiers aggregated over all considered datasets, cf. Section~\ref{sec:setting_tuned}.
As a baseline, we also include the results from the quantifiers that are based on the default logistic regressor. 
These results yield a few key findings. 
First, for most algorithms, tuning the base classifier does not seem to have a significant positive effect. 
Instead, for the best performing algorithms \emph{MS}, \emph{TSX}, \emph{FM}, and \emph{TSMax}, the performance even appears to deteriorate.
The few exceptions where tuning classifiers appear to strongly benefit the predictions include the \emph{CC}, \emph{PCC}, \emph{CDE}, \emph{GAC} methods.
While the first two directly apply the classify and count principle, where it is to be expected that more accurate classification will tend to yield more accurate quantification, the results for the  \emph{CDE} and \emph{GAC} methods pose as outliers.
It is also notable that the effects of parameter tuning often vary strongly over the given datasets, as can be seen in Tables \ref{tab:clf_binary_ac}, \ref{tab:clf_binary_dm} and \ref{tab:clf_binary_cc} in Appendix \ref{ap:main}.
Specifically for the \emph{PAC} and \emph{GPAC} methods, strong fluctuations over all datasets can be observed, while their overall distribution of error scores, as depicted in Figure \ref{fig:clf_bin}, appears quite robust.
Regarding the \emph{SVM-K} and \emph{SVM-Q} methods, we observe that applying the alternative RBF-kernel appears to have a slight positive effect, but these \emph{RBF-K} and \emph{RBF-Q} variants still show inferior performance compared to most of the other quantifiers, while at the same time coming at very high computational costs. 
Overall, the given results also appear consistent over both the \emph{AE} and the \emph{NKLD} metrics.

\subsubsection{Experiments on Multiclass Data}

The results of our experiments on quantification with tuned base classifiers in the multiclass setting can be found in Figure \ref{fig:clf_mc}.
In contrast to the binary setting, we observe that tuning the base classifiers appears to have a strong positive effect for almost every pair of quantifier and base classifier---the only base classifier for which the effect of tuning appears less consistent is the AdaBoost classifier.
However, when additionally considering the average error scores per dataset in Table \ref{tab:clf_multiclass}, 
it has to be noted that this effect is not consistent over all the datasets that we considered, but still yields a substantial improvement on aggregate.
Further, only the probability-based \emph{EM}, \emph{GPAC}, and \emph{FM} methods, in which the logistic base classifier have been tuned, appear to outperform all default variants of the given quantifiers with respect to both the AE and the NKLD scores.
The EM algorithm with tuned logistic base classifier also appears to stand out overall with respect to both error scores.

\section{A Case Study on the LeQua Challenge Data}\label{sec:lequa}

To validate our findings in an external benchmark framework, 
we further conducted a case study on the datasets from the \emph{LeQua 2022 challenge} \citep{esuli_lequa_2022, esuli_concise_2022}. 
In this challenge, \cite{esuli_lequa_2022} provided the participants with two textual datasets, one with binary labels and one with multiclass labels. Each were given both in a raw document format and in a preprocessed numerical vector format---the preprocessed features were derived from the average \emph{GloVe} \citep{pennington_glove_2014} embedding vectors of the words in each document, which were standardized to zero mean and unit variance. 
The data was collected from a large crawl of Amazon product reviews, where the binary labels were derived from the sentiment of the reviews, and the 28 labels in the multiclass task correspond to product categories.
The challenge then consisted of two main tasks, where the first task was to perform quantification on the preprocessed datasets, and the second task was to quantify on the raw documents in an end-to-end fashion that could occur in practical scenarios.
Both tasks were split into two subtasks, in which (i) the binary and (ii) the multiclass versions of the dataset were to be analyzed.

In our case study, we only considered the preprocessed data from the first task, since preprocessing techniques for textual datasets are out of scope for this work, and differences in preprocessing may further hinder comparability of results.
Both binary and multiclass dataset were split into training, validation and test data.
Only for the training data, which consisted of 5,000 documents in the binary and 20,000 documents in the multiclass setting, class labels were provided for each document.
The validation sets consisted of 1,000 samples of 250 (binary) and 1,000 (multiclass) documents each, where no class labels were given for any document, but the label distribution of each sample was known and could be used for model tuning.
Finally, the test sets in both the binary and the multiclass dataset contained 5,000 data samples, each consisting of 250 documents in the binary and 1,000 documents in the multiclass case.
We note that the setting in this challenge specifically differs from the experimental settings in this work by the availability of large amounts of validation data that is held out from the relatively small amount of training data.
Further, the number of labels in the multiclass part of the challenge ($L=28$) is significantly higher than the maximum number of classes  used in our experiments ($L=5$). 

On the LeQua dataset, we conducted three experiments. 
First, as in our main experiments, we applied all quantifiers using their default parameters.
In the second experiment, we again considered all quantification methods that use a base classifier, and tuned the parameters of these classifiers on the training data before applying the quantifiers with tuned base classifiers on the test data.
In the third and final experiment, we explored the effects of tuning the parameters, including base classifiers, for quantification, making use of the given validation samples.

In the following, we describe the results from these experiments, focusing in particular on results with respect to the AE measure. 
Additional results with respect to the NKLD score are presented in Appendix \ref{ap:lequa_plots}, where it can be seen that in the binary case, the results were mostly very similar.
For the multiclass setting on this dataset, where results differed more strongly from the AE-based results, we do not consider the NKLD measure to be very suitable.
This is due to the NKLD measure specifically punishing cases where prevalences of classes are falsely estimated to be zero.
Given that the multiclass dataset has $L = 28$ classes, very low prevalences of individual classes are, however, very frequent by nature and thus less of a concern.

\subsection{Comparison of Quantifiers With Default Parameters}\label{sec:lequa_main}

We begin with presenting the results from using quantifiers with their default parameters on the LeQua dataset---we used the same parametrization as in our main experiments, which has been outlined in Section \ref{sec:param_main}.
All quantifiers have been trained on the given training data, and directly applied on the test data without considering the validation samples.
The only optimization that was done was for the \emph{HDx}, \emph{readme}, and \emph{QF} methods, which require binned input data.
For these methods, we optimized the binning strategy by varying the number of bins that would be used for all features between 2 and 8, and by testing equidistant as well as quantile-based binning.
The results that we report are based on the binning strategy that yielded the best average AE score on the validation sets.

\begin{figure}[t]
	\centering
	\begin{subfigure}[b]{0.495\textwidth}
		\centering
		\includegraphics[width=\textwidth]{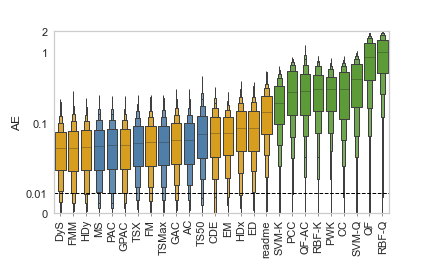}
		\caption{Distribution of absolute error (AE) scores on the \textbf{binary} LeQua test set}
	\end{subfigure}
    \hfill
	\begin{subfigure}[b]{0.495\textwidth}
		\centering
		\includegraphics[width=\textwidth]{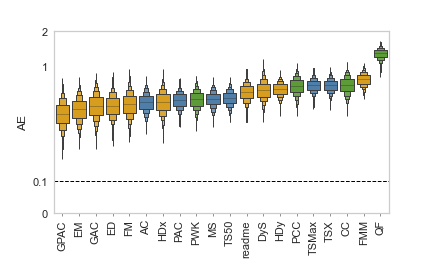}
		\caption{Distribution of absolute error (AE) scores on the \textbf{multiclass} LeQua test set}
	\end{subfigure}
	\caption{Results of our experiments with untuned quantifiers on the LeQua test sets. Plots are scaled logarithmically above the dotted vertical threshold, and linearly below.  Colors indicate the category of the algorithm. In addition to all quantifiers used in our main experiments, we present results of the \emph{RBF-K} and \emph{RBF-Q} methods, which are variants of the \emph{SVM-K} and \emph{SVM-Q} that use an RBF-kernel instead of a linear kernel.  Overall results are in line with our findings from the main experiments. On the binary data, the \emph{DyS} and \emph{FMM} methods appear to work best, on the multiclass data, the \emph{GPAC} and \emph{EM} methods appear to stand out.}
	\label{fig:lequa_main}
\end{figure}

The results of these experiments can be found in Figure \ref{fig:lequa_main}, where we depict the distribution of AE scores on the test datasets.
Overall, these results appear to be in line with the findings from our main experiments.
On the binary dataset, \emph{DyS} and \emph{MS} appear to work best, with methods such as \emph{PAC}, \emph{GPAC}, \emph{TSX}, \emph{FM}, \emph{TSMax} appearing relatively competitive, and classify-and-count based methods, even when optimized for quantification, appear to fall behind.
On the multiclass datasets, specifically the \emph{GPAC} and \emph{EM} methods appear to stand out, and overall, natural multiclass quantifiers seem to outperform the one-vs.-rest approaches.
As a notable difference to our main experiments, the \emph{HDx} and \emph{readme} methods overall appear to perform relatively weak. 
We suppose that this is due to these methods requiring binned inputs, for which we may not have found an optimal binning strategy. 
Even though, as noted before, we have conducted some optimization to the binning, more fine-grained optimization of bins, which could also include different strategies for different features, might be required.

\begin{figure}[t]
	\centering
	\begin{subfigure}[b]{\textwidth}
		\centering
		\includegraphics[width=\textwidth]{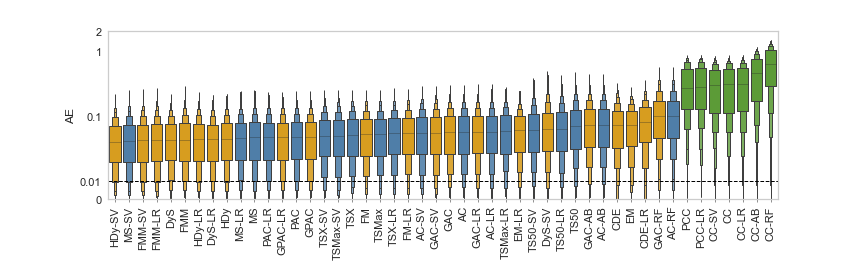}
		\caption{Distribution of absolute error (AE) scores on the binary LeQua dataset.}
	\end{subfigure}
	\hfill
	\begin{subfigure}[b]{\textwidth}
		\centering
		\includegraphics[width=\textwidth]{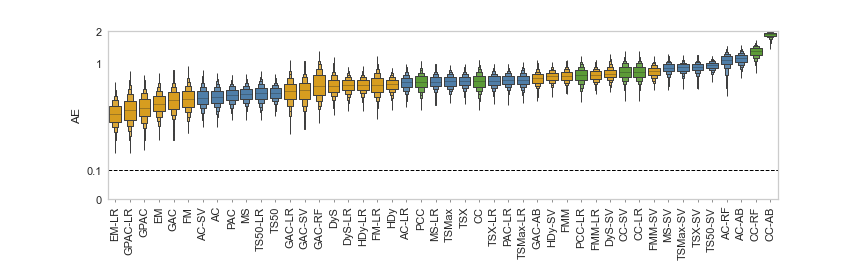}
		\caption{Distribution of absolute error (AE) scores on the multiclass LeQua dataset}
	\end{subfigure}
	\caption{Results from applying quantifiers with tuned base classifiers on the LeQua data.
    In the binary setting and for natural multiclass quantifiers, base classifiers were optimized with respect to their accuracy. 
    For quantifiers that apply the one-vs.-rest approach in the multiclass setting, the binary base classifiers were tuned with respect to \emph{balanced} accuracy. 
    Plots are scaled logarithmically above the dotted vertical threshold, and linearly below.  Colors indicate the category of the algorithm. Algorithms based on untuned logistic regression classifiers are denoted as before (no suffix), alternative tuned base classifiers are marked with respective suffixes: logistic regressors (LR), support vector machines (SV), random forests (RF) and AdaBoost (AB).  Overall, there appears to be no consistent positive effect from tuning base classifiers.}
	\label{fig:lequa_clf}
\end{figure}

\subsection{Comparison of Quantifiers with Tuned Base Classifiers}

Next, we present the results from applying quantification methods for which the base classifiers have been tuned.
Again, we applied the same parameter grid as in previous experiments (cf. Section \ref{sec:setting_tuned}), and tuned the parameters on the training set via cross-validation to optimize their accuracy---since the validation data did not provide labels for individual documents, this data could not be used for tuning.

The AE scores that we obtained from these experiments are depicted in Figure \ref{fig:lequa_clf}.
On both the binary and the multiclass data, we overall see a mixed picture regarding the benefits of tuning the base classifier.
Some methods, such as the \emph{EM} and \emph{GPAC} approaches seem to improve particularly in the multiclass case, while other methods, such as also the classify and count-based approaches, rather seem to deteriorate.
However, there are no general trends for any group of algorithms, which is overall in line with the results from our main experiments.

\begin{figure}[t]
	\centering
	\begin{subfigure}[b]{\textwidth}
		\centering
		\includegraphics[width=\textwidth]{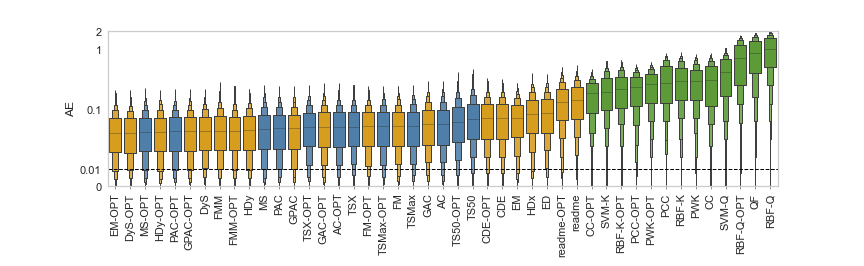}
		\caption{Distribution of absolute error (AE) scores on the binary LeQua dataset.}
	\end{subfigure}
	\hfill
	\begin{subfigure}[b]{\textwidth}
		\centering
		\includegraphics[width=\textwidth]{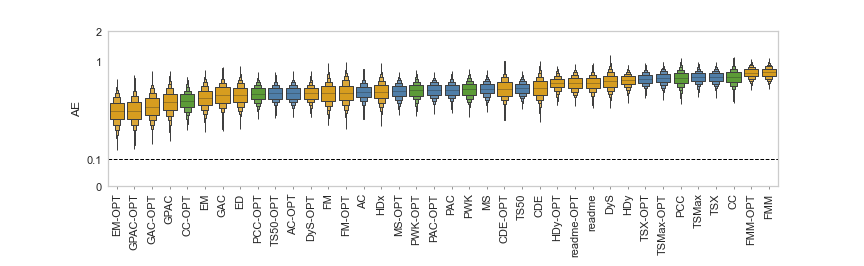}
		\caption{Distribution of absolute error (AE) scores on the multiclass LeQua dataset}
	\end{subfigure}
	\caption{ Results of our experiments on the LeQua test data using quantifiers that were tuned on the LeQua validation data.  Plots are scaled logarithmically above the dotted vertical threshold, and linearly below.  Colors indicate the category of the algorithm. Algorithms using their default parameters are denoted as before (no suffix), their tuned variants are marked with a short suffix (OPT). In both the binary and the multiclass task, the tuned \emph{EM} algorithm appears to perform best with respect to the AE score.
 }
	\label{fig:lequa_tuned}
\end{figure}

\subsection{Comparison of Tuned Quantifiers}\label{sec:lequa_tuned}

Finally, we discuss our results on the experiments in which we tuned the parameters of all quantification methods using the extensive validation data available within the LeQua dataset.
Parameters have been tuned with respect to the AE score on the validation data, and the optimization also considered parameters of the logistic regressor that was chosen as the base classifier for all quantifiers that require a base classifier to form their predictions.
A detailed overview of the used parameter grids can be found in Appendix \ref{ap:lequa_params}.

The distribution of the resulting AE scores is shown in Figure \ref{fig:lequa_tuned}, where we can see that tuning parameters appears to have a significant positive effect on the outcomes.

In the binary setting, the tuned \emph{EM} and \emph{DyS} methods appear to perform best, with the tuned \emph{MS}, \emph{HDy}, \emph{PAC} and \emph{GPAC} methods being behind only marginally.
Interestingly, the untuned \emph{DyS}, \emph{MS}, \emph{PAC} and \emph{GPAC} methods still appear to outperform the tuned variants of almost every other algorithm that we considered.
Further, it is notable that specifically the \emph{EM} algorithm appears to be strongly impacted from the parameter tuning.
Strong positive impact can also be observed for all classify and count-based approaches, but even after tuning, these methods perform worse than almost any other method with default parameters.

In the multiclass case, we also observe significant improvements on the resulting error scores.
Particularly, the \emph{EM} and \emph{GPAC} methods appear to perform stronger than the rest, with \emph{GAC} also showing strong results after tuning.
The untuned versions of these algorithms further appear to outperform almost all other methods, even after tuning, with respect to the AE scores.
The only exception is the \emph{CC} method, which performs surprisingly strong on this dataset.

\newpage
\section{Discussion}\label{sec:discussion}

Next, we discuss the main results and potential limitations of our study.

\subsection{Discussion of Results}
Our experiments yield substantially different results for the binary case compared to the multiclass case, both in terms of overall quality of performance, and in terms of which algorithms perform best.

In the binary case, we identified a few algorithms which appear to work particularly well with respect to both absolute error (AE) and NKLD, namely the \emph{HDy}, \emph{FMM}, \emph{MS}, \emph{TSMax}, \emph{Friedman's method}, and the \emph{DyS} framework. 
These methods stand out both in terms of their ranks and their overall error distribution (though \emph{HDy} appears to have a slight edge over the rest in that regard). 
Next to these algorithms, other methods show similarly strong performances, at least with respect to one of the two measures that were considered. In that regard, \emph{TSX} has shown very strong performances with respect to the AE score, while the \emph{ED} method appears to work particularly well with respect to the NKLD measure.
The strong performance of the \emph{MS} and \emph{TSMax} methods indicate that the simple idea behind the \emph{Adjusted Count} approach, even when using a rather unsophisticated baseline classifier, can still yield very decent results as long as numerical stability, i.e., a big denominator in Equation \ref{eq:AC}, is ensured. 
In that regard, the \emph{MS} method also benefits from the policy that all thresholds, for which the denominator is below 0.25, are excluded.
A similar argument can be made for the superiority of the \emph{DyS} framework, which includes the \emph{HDy} method, and Forman's mixture model (\emph{FMM}) compared to other distribution matching methods that utilize predictions from classifiers.
Specifically, the approach of binning confidence scores into more than just two classes, which ultimately adds more equations to the system in Equation \ref{eq:dmbase}, also appears to yield more robust results.
By contrast, classifiers that optimize quantification-oriented loss functions also tended to show worse performance than the majority of other quantifiers.
This is overall another strong indicator that pure classification without adjustments for potential distributions shifts does not perform well for quantification.
The reason for that is that under a shift in the class distribution, predictions are strongly biased toward the training distribution, as exemplified in our experiments. 
This practical outcome is also clearly in line with Forman's Theorem \citep{forman_quantifying_2008}, which states that when a distribution shift is given, a bias in the \emph{CC} estimates toward the training distribution is to be expected.
This finding bears some contrast to a recent discussion of this kind of approach by \citet{moreo_re-assessing_2021}, who have reassessed the performance of the \emph{Classify and Count} approach, and found that when doing careful optimization of hyperparameters, such quantification-oriented classification approaches would deliver near-state-of-the-art performance, although still being inferior.
Our experimental outcomes suggest that this type of approach should be used only carefully for quantification, as a vulnerability toward distribution shifts in theory as well as in experimental results can be observed clearly. 
Finally, the overall subpar performance of the CDE iterator is also in line with theoretical results that emphasized its lack of consistency \citep{tasche_fisher_2017}.

Considering the multiclass case, results are qualitatively different. Most notably, error scores were significantly higher than in binary quantification.
Another key difference is that methods such as \emph{HDy}, \emph{DyS}, \emph{MS} or \emph{TSMax}, which have excelled in binary quantification, only showed mediocre performance in the multiclass case.
By contrast, distribution matching methods that naturally extend to the multiclass setting appeared to work best, with the \emph{HDx} method appearing to stand out.
These results indicate that generalizing quantification methods to the multiclass case via a one-vs.-rest approach is not an optimal strategy for multiclass quantification.
This finding has been taken up and analyzed more deeply recently by \citet{donyavi_mc-sq_2023, donyavi_mc-sq_2024}, who pointed out that this is due a shift in the distributions $P(X|Y)$ which is introduced when binarizing multiclass labels for the one-vs.-rest settings.

From our experiments with tuned base classifiers, we can further infer that in general, more accurate base classifiers do not yield more accurate estimations of class prevalences when used by quantifiers.
Particularly in the binary case, we did hardly observe any positive effect from using tuned base classifiers.
For quantifiers that utilize misclassification rates, an explanation of this outcome might be that having somewhat higher misclassification rates may actually yield more numerical stability in the predictions.
The only exception to this pattern was given by the classify and count-based methods \emph{CC} and \emph{PCC}, for which it could also be expected that optimizing the base classifiers would beneficial.
Yet, these methods still did not appear on par with the best-performing methods after this kind of tuning.
This overall result appears to contradict findings from a simulation study by \citet{Tasche_2019_confidence}, who concluded that more acurate base classifiers led to shorter confidence intervals of class prevalence estimations.
However, Tasche only considered normally distributed synthetic data, which likely does not accurately represent the nature of real-world data. 
In the multiclass setting, tuned base classifiers appeared to have a more positive effect on aggregate over all datasets, specifically for the \emph{EM} and \emph{GPAC} methods, for which their tuned variants also appear strong in the LeQua case study.
Yet, when looking at the average error scores over the individual datasets, one can observe that this is not at all a consistent trend, and the strong aggregate performance appears to result from outstanding performances on a few of the overall relatively low number of only nine multiclass datasets on which we performed this hyperparameter tuning.
In conclusion, if in practice resources for parameter tuning are available, we recommend that these should not be used for training more accurate base classifers.
Instead, one should consider parameters of base classifiers as parameters of the quantifier applying it, and directly optimize for quantification performance.

Considering our case study on the LeQua data, the results obtained from applying quantifiers with default parameterization and quantifiers with tuned base classifiers were mostly in line with the main results.
Smaller variations like the slightly weaker performance of the \emph{TSMax} and \emph{FM} methods have also been observed over the individual datasets in the main experiments, and the relatively weak performance of the \emph{HDx} and \emph{readme} methods is likely due to the binning of the given data not being optimal.
Novel insights were, however, gained from the final part of the case study, in which the hyperparameters of all quantifiers, including those of base classifiers, were tuned for quantification performance.
In these experiments, we observed that those methods which already performed best with their default parameters also were among the best methods after tuning.
Specifically, the tuned \emph{DyS} and \emph{MS} methods were among the best methods in the binary setting, whereas the tuned \emph{EM} and \emph{GPAC} methods overall yielded the best performance in the multiclass setting.
Further, the untuned variants of these methods also performed better than the tuned versions of most other methods, with only a few exceptions.
Specifically, in the binary setting, the best performing algorithm was given by the \emph{EM} method, which appeared rather mediocre with default parametrization. 
Given that also in the multiclass setting, this method did strongly improve its performance with respect to the AE, this indicates that this algorithm strongly relies on proper calibration of its probabilistic base classifier, as has also been found by \cite{esuli_critical_2020}.
The results on the binary LeQua data also provide further evidence that classify-and-count-based approaches are not reliable quantifiers, given that even after tuning, these methods yielded worse performances than untuned variants of all other methods in the binary setting.
Somewhat surprisingly, however, the tuned \emph{CC} and \emph{PCC} method appeared to perform relatively well in the multiclass setting, although being clearly behind the strongest algorithms.
Given that these methods can be considered natural multiclass quantifiers, this could be attributed to the overall observation that one-vs.-rest approaches are not suitable for the multiclass setting.
By contrast, the only natural multiclass quantifiers that performed worse than these methods after tuning are given by the \emph{readme} and \emph{FM} approaches, which generally did not appear to work well on the LeQua dataset.

\subsection{Limitations}
This paper presents an extensive empirical comparison of state-of-the-art quantification methods. As such, our results are necessarily affected by some experimental design choices.

First, in our main experiments, we relied on default parameters of the individual algorithms and did not perform extensive hyperparameter optimization for the quantification algorithms on each dataset.
While on the one hand this is due to computational considerations---we have performed more than 295,000 experiments with 10 sampling iterations each, making extra hyperparameter optimization steps infeasible---this also reflects the performance that these methods would achieve when being used off-the-shelf.
Further, there is surprisingly little research regarding tuning protocols for quantification (cf. Section 3.5,  \cite{esuli_evaluation_2023}). 
Standard model selection approaches such as $k$-fold cross-validation may, for instance, not necessarily work well for quantification, as these are unlikely to yield strong shifts between training and test distributions.
Big validation sets, by contrast, are in general neither available nor trivial to construct, and thus, non-optimal optimization schemes may also bias the given results.
However, we tested hyperparameter tuning on the dataset from the LeQua challenge, where a huge set of validation samples has been provided.

Similarly, properly designing sampling protocols for evaluation is not trivial either, and design choices in our approach may have yielded unintended biases.
We aimed to cover a wide range of training set sizes, training/test distributions, and distribution shifts, but, for instance, our grids for training and test distributions in the multiclass experiments are much less coarse than in the binary case and, therefore, might not completely represent all possible scenarios.
In addition, while we tried to broadly sample from diverse distributions, there may be imbalances in the representations of individual classes, given that in our undersampling approach, instances from overall less populated classes are more likely to be used than instances from more populated classes.
However, such imbalances in given datasets are generally hard to work around, and different approaches such as oversampling, i.e., sampling with replacement large amounts of instances from a very limited pool, may also come with different caveats.
Although in the literature it is agreed that training and test distributions \citep{hassan2021pitfalls, esuli_evaluation_2023} as well as test set sizes \citep{maletzke_importance_2020} should be varied artificially for a proper evaluation of quantification methods, there has also been limited discussion on how to effectively sample such distributions from a given dataset in a representative fashion, specifically when it is limited in size or unbalanced in its class distribution.

Further, despite the broad range of datasets considered, an analysis as we have just conducted cannot realistically cover all possible application scenarios.
In that regard, we would like to note that this study does not include algorithms from the authors or collaborators, such that the authors do not have stakes in any particular outcome.

Finally, 
the research field of quantification is very dynamic, and 
more recently published methods such as novel ensemble approaches \citep{donyavi_mc-sq_2024} or the continuous sweep \citep{kloos2023continuous} have not been included in our evaluation.
Similarly, related problems such as ordinal quantification \citep{sakai-2021-evaluating,castano_ordinal_2024,bunse2024regularization} or multi-label quantification \citep{moreo_multilabel_2024}, which have gained some more research interest recently, are out of scope for this study, and systematic analyses of methods for these problems could pose an interesting avenue for future research.

\section{Conclusions}\label{sec:conclusion}

In this study, we have conducted a thorough experimental comparison of 24 quantification methods over 40 datasets, involving more than 5 million algorithm runs.
In our experiments, we have both considered the binary and the multiclass case in quantification, and have also specifically considered the impact of shifting class label distributions between training and test data, as well as the impact of having relatively small training sets.
In the binary case, we have identified a group of methods which generally appear to work best, namely the threshold selection-based \emph{Median Sweep} and \emph{TSMax} methods \citep{forman_quantifying_2008}, the distribution matching approaches from the \emph{DyS} framework \citep{maletzke_dys_2019} including \emph{HDy} \citep{gonzalez-castro_class_2013}, Forman's mixture model \citep{forman_counting_2005}, and \emph{Friedman's method} \citep{friedman_friedman_2014}.
Regarding the multiclass case, a group of distribution matching methods, which naturally extend to multiclass quantification, appeared to be generally superior to the other evaluated algorithms.
We provide further evidence that the multiclass setting in general is much harder to solve for established quantification methods, as the obtained error scores consistently were multiple times higher than in the binary case.
This indicates a certain potential for future research on this specific setting.
Further, our experiments demonstrate that more accurate base classifiers do not in general yield more accurate quantification.
In addition, our results demonstrate that algorithms that are based on the classify and count principle, even when the underlying classifier is optimized for quantification, on average exhibit worse performance compared to other specialized solutions. 
Overall, we hope our findings provide guidance to practitioners in choosing the right quantification algorithm for a given application and aid researchers in identifying promising directions for future research.


\FloatBarrier
\section*{Acknowledgements} 
We acknowledge support by the state of Baden-Württemberg through bwHPC and the German Research Foundation (DFG) through grant INST 35/1597-1 FUGG.
We thank Fabrizio Sebastiani and Letizia Milli for their help and for providing the code for the quantification forests.

\bibliographystyle{plainnat}  
\bibliography{bibliography}

\begin{thebibliography}{45}
\providecommand{\natexlab}[1]{#1}
\providecommand{\url}[1]{\texttt{#1}}
\expandafter\ifx\csname urlstyle\endcsname\relax
  \providecommand{\doi}[1]{doi: #1}\else
  \providecommand{\doi}{doi: \begingroup \urlstyle{rm}\Url}\fi

\bibitem[Barranquero et~al.(2013)Barranquero, González, Díez, and del
  Coz]{barranquero_study_2013}
Jose Barranquero, Pablo González, Jorge Díez, and Juan~José del Coz.
\newblock On the study of nearest neighbor algorithms for prevalence estimation
  in binary problems.
\newblock \emph{Pattern Recognition}, 46\penalty0 (2):\penalty0 472--482, 2013.

\bibitem[Barranquero et~al.(2015)Barranquero, Díez, and José~del
  Coz]{barranquero_quantification-oriented_2015}
Jose Barranquero, Jorge Díez, and Juan José~del Coz.
\newblock Quantification-oriented learning based on reliable classifiers.
\newblock \emph{Pattern Recognition}, 48\penalty0 (2):\penalty0 591--604, 2015.

\bibitem[Bella et~al.(2010)Bella, Ferri, Hernández-Orallo, and
  Ramírez-Quintana]{bella_quantification_2010}
Antonio Bella, Cesar Ferri, José Hernández-Orallo, and María~José
  Ramírez-Quintana.
\newblock Quantification via probability estimators.
\newblock In \emph{2010 IEEE International Conference on Data Mining}, pages
  737--742, Sydney, Australia, 2010.

\bibitem[Bunse et~al.(2024)Bunse, Moreo, Sebastiani, and
  Senz]{bunse2024regularization}
Mirko Bunse, Alejandro Moreo, Fabrizio Sebastiani, and Martin Senz.
\newblock Regularization-based methods for ordinal quantification.
\newblock \emph{Data Mining and Knowledge Discovery}, 38\penalty0 (6):\penalty0
  4076--4121, 2024.

\bibitem[Castaño et~al.(2024)Castaño, González, González, and del
  Coz]{castano_ordinal_2024}
Alberto Castaño, Pablo González, Jaime~Alonso González, and Juan~José del
  Coz.
\newblock Matching distributions algorithms based on the earth mover’s
  distance for ordinal quantification.
\newblock \emph{IEEE Transactions on Neural Networks and Learning Systems},
  35\penalty0 (1):\penalty0 1050--1061, 2024.

\bibitem[Dempster et~al.(1977)Dempster, Laird, and
  Rubin]{dempster_maximum_1977}
A.~P. Dempster, N.~M. Laird, and D.~B. Rubin.
\newblock Maximum likelihood from incomplete data via the {EM} algorithm.
\newblock \emph{Journal of the Royal Statistical Society: Series B
  (Methodological)}, 39\penalty0 (1):\penalty0 1--22, 1977.

\bibitem[Dem{\v{s}}ar(2006)]{demsar_statistical_2006}
Janez Dem{\v{s}}ar.
\newblock Statistical comparisons of classifiers over multiple data sets.
\newblock \emph{Journal of Machine Learning Research}, 7\penalty0 (1):\penalty0
  1--30, 2006.

\bibitem[Deza and Deza(2009)]{deza_encyclopedia_2009}
Michel~Marie Deza and Elena Deza.
\newblock \emph{Encyclopedia of {Distances}}.
\newblock Springer Berlin Heidelberg, Berlin \& Heidelberg, Germany, 2009.

\bibitem[Diamond and Boyd(2016)]{diamond2016cvxpy}
Steven Diamond and Stephen Boyd.
\newblock {CVXPY}: {A} {P}ython-embedded modeling language for convex
  optimization.
\newblock \emph{Journal of Machine Learning Research}, 17\penalty0
  (83):\penalty0 1--5, 2016.

\bibitem[Donyavi et~al.(2023)Donyavi, Serapião, and
  Batista]{donyavi_mc-sq_2023}
Zahra Donyavi, Adriane B.~S. Serapião, and Gustavo Batista.
\newblock {MC-SQ}: A highly accurate ensemble for multi-class quantification.
\newblock In \emph{Proceedings of the 2023 SIAM International Conference on
  Data Mining (SDM)}, pages 622--630, Minneapolis, Minnesota, 2023.

\bibitem[Donyavi et~al.(2024)Donyavi, Serapião, and
  Batista]{donyavi_mc-sq_2024}
Zahra Donyavi, Adriane B.~S. Serapião, and Gustavo Batista.
\newblock {MC-SQ} and {MC-MQ}: Ensembles for multi-class quantification.
\newblock \emph{IEEE Transactions on Knowledge and Data Engineering},
  36\penalty0 (8):\penalty0 4007--4019, 2024.

\bibitem[Esuli et~al.(2010)Esuli, Sebastiani, and Abasi]{esuli_ai_2010}
Andrea Esuli, Fabrizio Sebastiani, and Ahmed Abasi.
\newblock {AI} and opinion mining, part 2.
\newblock \emph{IEEE Intelligent Systems}, 25\penalty0 (4):\penalty0 72--79,
  2010.

\bibitem[Esuli et~al.(2018)Esuli, Fernández, and
  Sebastiani]{esuli_recurrent_2018}
Andrea Esuli, Alejandro~Moreo Fernández, and Fabrizio Sebastiani.
\newblock A recurrent neural network for sentiment quantification.
\newblock In \emph{Proceedings of the 27th ACM International Conference on
  Information and Knowledge Management}, pages 1775--1778, Torino, Italy, 2018.

\bibitem[Esuli et~al.(2021)Esuli, Molinari, and
  Sebastiani]{esuli_critical_2020}
Andrea Esuli, Alessio Molinari, and Fabrizio Sebastiani.
\newblock A critical reassessment of the {Saerens-Latinne-Decaestecker}
  algorithm for posterior probability adjustment.
\newblock \emph{ACM Transactions on Information Systems}, 39\penalty0
  (2):\penalty0 1--34, 2021.

\bibitem[Esuli et~al.(2022{\natexlab{a}})Esuli, Moreo, and
  Sebastiani]{esuli_lequa_2022}
Andrea Esuli, Alejandro Moreo, and Fabrizio Sebastiani.
\newblock {LeQua}@{CLEF 2022}: Learning to quantify.
\newblock In \emph{Advances in Information Retrieval: 44th European Conference
  on IR Research, Part II}, pages 374--381, Stavanger, Norway,
  2022{\natexlab{a}}.

\bibitem[Esuli et~al.(2022{\natexlab{b}})Esuli, Moreo, Sebastiani, and
  Sperduti]{esuli_concise_2022}
Andrea Esuli, Alejandro Moreo, Fabrizio Sebastiani, and Gianluca Sperduti.
\newblock A concise overview of {LeQua}@{CLEF 2022}: Learning to quantify.
\newblock In \emph{Experimental IR Meets Multilinguality, Multimodality, and
  Interaction: 13th International Conference of the CLEF Association}, pages
  362--381, Bologna, Italy, 2022{\natexlab{b}}. Springer.

\bibitem[Esuli et~al.(2023)Esuli, Fabris, Moreo, and
  Sebastiani]{esuli_evaluation_2023}
Andrea Esuli, Alessandro Fabris, Alejandro Moreo, and Fabrizio Sebastiani.
\newblock \emph{Learning to Quantify}.
\newblock Springer International Publishing, Cham, Switzerland, 2023.

\bibitem[Firat(2016)]{firat_unified_2016}
Aykut Firat.
\newblock Unified framework for quantification.
\newblock \emph{arXiv preprint arXiv:1606.00868}, 2016.

\bibitem[Forman(2005)]{forman_counting_2005}
George Forman.
\newblock Counting positives accurately despite inaccurate classification.
\newblock In \emph{Proceedings of the 16th {European} {Conference} on {Machine}
  {Learning}}, pages 564--575, Porto, Portugal, 2005.

\bibitem[Forman(2008)]{forman_quantifying_2008}
George Forman.
\newblock Quantifying counts and costs via classification.
\newblock \emph{Data Mining and Knowledge Discovery}, 17\penalty0 (2):\penalty0
  164--206, 2008.

\bibitem[Friedman(2014)]{friedman_friedman_2014}
Jerome~H. Friedman.
\newblock Class counts in future unlabeled samples, 2014.
\newblock Presentation at MIT CSAIL Big Data Event.

\bibitem[Friedman(1940)]{friedman_comparison_1940}
Milton Friedman.
\newblock A comparison of alternative tests of significance for the problem of
  m rankings.
\newblock \emph{The Annals of Mathematical Statistics}, 11\penalty0
  (1):\penalty0 86--92, 1940.

\bibitem[González et~al.(2017)González, Castaño, Chawla, and del
  Coz]{gonzalez_review_2017}
Pablo González, Alberto Castaño, Nitesh~V. Chawla, and Juan~José del Coz.
\newblock A review on quantification learning.
\newblock \emph{ACM Computing Surveys}, 50\penalty0 (5):\penalty0 1--40, 2017.

\bibitem[González-Castro et~al.(2013)González-Castro, Alaiz-Rodríguez, and
  Alegre]{gonzalez-castro_class_2013}
Víctor González-Castro, Rocío Alaiz-Rodríguez, and Enrique Alegre.
\newblock Class distribution estimation based on the {Hellinger} distance.
\newblock \emph{Information Sciences}, 218\penalty0 (1):\penalty0 146--164,
  2013.

\bibitem[Hassan et~al.(2021)Hassan, Maletzke, and Batista]{hassan2021pitfalls}
Waqar Hassan, Andr{\'e}~Gustavo Maletzke, and Gustavo Enrique de Almeida
  Prado~Alves Batista.
\newblock Pitfalls in quantification assessment.
\newblock In \emph{First International Workshop on Learning to Quantify:
  Methods and Applications (LQ 2021)}, pages 1--10, Virtual Event, Gold Coast,
  Australia, 2021.

\bibitem[Hopkins and King(2010)]{hopkins_method_2010}
Daniel~J. Hopkins and Gary King.
\newblock A method of automated nonparametric content analysis for social
  science.
\newblock \emph{American Journal of Political Science}, 54\penalty0
  (1):\penalty0 229--247, 2010.

\bibitem[Joachims(2005)]{joachims_support_2005}
Thorsten Joachims.
\newblock A support vector method for multivariate performance measures.
\newblock In \emph{Proceedings of the 22nd {International} {Conference} on
  {Machine} {Learning}}, pages 377--384, Bonn, Germany, 2005.

\bibitem[Kawakubo et~al.(2016)Kawakubo, du~Plessis, and
  Sugiyama]{kawakubo_computationally_2016}
Hideko Kawakubo, Marthinus~Christoffel du~Plessis, and Masashi Sugiyama.
\newblock Computationally efficient class-prior estimation under class balance
  change using energy distance.
\newblock \emph{IEICE Transactions on Information and Systems}, 99\penalty0
  (1):\penalty0 176--186, 2016.

\bibitem[Kloos et~al.(2023)Kloos, Karch, Meertens, and
  de~Rooij]{kloos2023continuous}
Kevin Kloos, Julian~D Karch, Quinten~A Meertens, and Mark de~Rooij.
\newblock Continuous sweep: An improved, binary quantifier.
\newblock \emph{arXiv preprint arXiv:2308.08387}, 2023.

\bibitem[Maletzke et~al.(2019)Maletzke, dos Reis, Cherman, and
  Batista]{maletzke_dys_2019}
André Maletzke, Denis dos Reis, Everton Cherman, and Gustavo Batista.
\newblock {DyS}: A framework for mixture models in quantification.
\newblock In \emph{Proceedings of the {AAAI} {Conference} on {Artificial}
  {Intelligence}}, pages 4552--4560, Honolulu, Hawaii, 2019.

\bibitem[Maletzke et~al.(2020)Maletzke, Hassan, Reis, and
  Batista]{maletzke_importance_2020}
André Maletzke, Waqar Hassan, Denis~dos Reis, and Gustavo Batista.
\newblock The importance of the test set size in quantification assessment.
\newblock In \emph{Proceedings of the {Twenty}-{Ninth} {International} {Joint}
  {Conference} on {Artificial} {Intelligence}}, pages 2640--2646, Yokohama,
  Japan, 2020.

\bibitem[Milli et~al.(2013)Milli, Monreale, Rossetti, Giannotti, Pedreschi, and
  Sebastiani]{milli_quantification_2013}
Letizia Milli, Anna Monreale, Giulio Rossetti, Fosca Giannotti, Dino Pedreschi,
  and Fabrizio Sebastiani.
\newblock Quantification trees.
\newblock In \emph{2013 {IEEE} 13th {International} {Conference} on {Data}
  {Mining}}, pages 528--536, Dallas, Texas, 2013.

\bibitem[Moreo and Sebastiani(2021)]{moreo_re-assessing_2021}
Alejandro Moreo and Fabrizio Sebastiani.
\newblock Re-assessing the ``classify and count'' quantification method.
\newblock In \emph{Advances in Information Retrieval: 43rd European Conference
  on IR Research, Part II}, pages 75--91, 2021.

\bibitem[Moreo et~al.(2021)Moreo, Esuli, and Sebastiani]{moreo_quapy_2021}
Alejandro Moreo, Andrea Esuli, and Fabrizio Sebastiani.
\newblock {QuaPy}: A python-based framework for quantification.
\newblock In \emph{Proceedings of the 30th ACM International Conference on
  Information \& Knowledge Management}, page 4534–4543, 2021.

\bibitem[Moreo et~al.(2023)Moreo, Francisco, and
  Sebastiani]{moreo_multilabel_2024}
Alejandro Moreo, Manuel Francisco, and Fabrizio Sebastiani.
\newblock Multi-label quantification.
\newblock \emph{ACM Transactions on Knowledge Discovery from Data}, 18\penalty0
  (1):\penalty0 1--36, 2023.

\bibitem[Nemenyi(1963)]{nemenyi_distribution_1963}
Peter~B. Nemenyi.
\newblock \emph{Distribution-free multiple comparisons}.
\newblock PhD thesis, Princeton University, Princeton, New Jersey, 1963.

\bibitem[Pennington et~al.(2014)Pennington, Socher, and
  Manning]{pennington_glove_2014}
Jeffrey Pennington, Richard Socher, and Christopher Manning.
\newblock {G}lo{V}e: Global vectors for word representation.
\newblock In \emph{Proceedings of the 2014 Conference on Empirical Methods in
  Natural Language Processing}, pages 1532--1543, Doha, Qatar, 2014.

\bibitem[Saerens et~al.(2002)Saerens, Latinne, and
  Decaestecker]{saerens_adjusting_2002}
Marco Saerens, Patrice Latinne, and Christine Decaestecker.
\newblock Adjusting the outputs of a classifier to new a priori probabilities:
  A simple procedure.
\newblock \emph{Neural Computation}, 14\penalty0 (1):\penalty0 21--41, 2002.

\bibitem[Sakai(2021)]{sakai-2021-evaluating}
Tetsuya Sakai.
\newblock Evaluating evaluation measures for ordinal classification and ordinal
  quantification.
\newblock In \emph{Proceedings of the 59th Annual Meeting of the Association
  for Computational Linguistics and the 11th International Joint Conference on
  Natural Language Processing (Volume 1: Long Papers)}, pages 2759--2769,
  Virtual Event, 2021.

\bibitem[Sebastiani(2020)]{sebastiani_evaluation_2020}
Fabrizio Sebastiani.
\newblock Evaluation measures for quantification: An axiomatic approach.
\newblock \emph{Information Retrieval Journal}, 23\penalty0 (3):\penalty0
  255--288, 2020.

\bibitem[Storkey(2008)]{storkey_shift_2008}
Amos Storkey.
\newblock When training and test sets are different: Characterizing learning
  transfer.
\newblock In Joaquin Quiñonero-Candela, Masashi Sugiyama, Anton Schwaighofer,
  and Neil~D. Lawrence, editors, \emph{{Dataset Shift in Machine Learning}}.
  The MIT Press, Cambridge, Massachusetts, 2008.

\bibitem[Tasche(2016)]{tasche_does_2016}
Dirk Tasche.
\newblock Does quantification without adjustments work?
\newblock \emph{arXiv preprint arXiv:1602.08780}, 2016.

\bibitem[Tasche(2017)]{tasche_fisher_2017}
Dirk Tasche.
\newblock Fisher consistency for prior probability shift.
\newblock \emph{Journal of Machine Learning Research}, 18\penalty0
  (95):\penalty0 1--32, 2017.

\bibitem[Tasche(2019)]{Tasche_2019_confidence}
Dirk Tasche.
\newblock Confidence intervals for class prevalences under prior probability
  shift.
\newblock \emph{Machine Learning and Knowledge Extraction}, 1\penalty0
  (3):\penalty0 805--831, 2019.

\bibitem[Xue and Weiss(2009)]{xue_quantification_2009}
Jack~Chongjie Xue and Gary~M. Weiss.
\newblock Quantification and semi-supervised classification methods for
  handling changes in class distribution.
\newblock In \emph{Proceedings of the 15th {ACM} {SIGKDD} {International}
  {Conference} on {Knowledge} {Discovery} and {Data} {Mining}}, pages 897--906,
  Paris, France, 2009.

\end{thebibliography}

\FloatBarrier
\appendix

\section{Performance Measures for Base Classifiers}\label{ap:clf_performance}

Several quantifiers that are analyzed in this study apply base classifiers and consider performance measures for these classifiers to form their predictions.
Similarly, we also consider such performance measures in our experiments on tuned base classifiers.
In the following, we briefly provide definitions for the performance measures that are used in this work.

We assume that we are given a dataset $D = \{ (\bm{x}_i, y_i) \}_{i=1}^N$ of $N$ instances, where $\bm{x}_i\in\mathbb{R}^k$ denotes the feature vector of each instance, and $y_i\in\{\ell_1, \dots, \ell_L\}$ the corresponding ground truth label.
In addition, we assume that we are given a classifier $c: \mathbb{R}^k \longrightarrow \{\ell_1, \dots, \ell_L\}$, which we apply on the given data to obtain the instance-wise predictions $\hat{y}_i = c(\bm{x}_i)$.
Then, the \emph{accuracy} of the classifier $c$ on this dataset is given by
\begin{equation*}
	e_\text{acc}(y, \hat{y}) = \frac{1}{N} \cdot \sum_{i=1}^N \mathbbm{1}(y_i = \hat{y}_i),
\end{equation*}
where $\mathbbm{1}(\bm\cdot)$ denotes the indicator function. When the distribution of class labels is unbalanced, the predictions of instances from minority classes carry little weight with respect to the resulting accuracy score.
In such cases, one may consider the \emph{balanced accuracy}, which is defined as
\begin{equation}\label{eq:balanced_accuracy}
	e_\text{bal-acc}(y, \hat{y}) = \frac{1}{L} \cdot \sum_{j=1}^L \frac{\sum_{i=1}^N\mathbbm{1}(y_i = \hat{y}_i= \ell_j)}{\sum_{i=1}^N\mathbbm{1}(y_i = \ell_j)}.
\end{equation}

In the binary setting, we distinguish more specifically between \emph{positive} and \emph{negative} instances, for which the ground-truth labels are given by $y_i = 1$ and $y_i = 0$, respectively.
Adjusted count-based quantifiers then specifically consider the ratio of predicted positives $\widehat{pos} = \frac{1}{N} \sum_{i=1}^N\mathbbm{1}(\hat{y}_i= 1)$, and adjust these for true positive rate ($tpr$) and false positive rate $fpr$ of their base classifiers, which are defined as
\begin{equation}\label{eq:tpr_fpr}
	tpr := tpr(y, \hat{y}) = \frac{\sum_{i=1}^N\mathbbm{1}(y_i = \hat{y}_i= 1)}{\sum_{i=1}^N\mathbbm{1}(y_i = 1)}  \,\,  \text{and}  \,\,  fpr := fpr(y, \hat{y}) = \frac{\sum_{i=1}^N\mathbbm{1}(y_i = 0 \wedge \hat{y}_i= 1)}{\sum_{i=1}^N\mathbbm{1}(y_i = 0)}.
\end{equation}

Similarly, one may consider the true negative rate ($tnr$) and false negative rate ($fnr$), which are defined as 
\begin{equation*}
	tnr := tnr(y, \hat{y}) = \frac{\sum_{i=1}^N\mathbbm{1}(y_i = \hat{y}_i= 0)}{\sum_{i=1}^N\mathbbm{1}(y_i = 0)} \,\, \text{and} \,\, fnr := fnr(y, \hat{y}) = \frac{\sum_{i=1}^N\mathbbm{1}(y_i = 1 \wedge \hat{y}_i= 0)}{\sum_{i=1}^N\mathbbm{1}(y_i = 1)}.
\end{equation*}

In the binary setting, the balanced accuracy corresponds to the average of true positive rate and true negative rate of the given classifier, i.e., in this setting it holds that
\begin{equation}\label{eq:bal_acc_binary}
	e_\text{bal-acc}(y, \hat{y}) = \tfrac{1}{2}(tpr + tnr).
\end{equation}

\begin{figure}[t]
	\centering
	\begin{subfigure}[b]{0.49\textwidth}
		\centering
		\includegraphics[width=\textwidth]{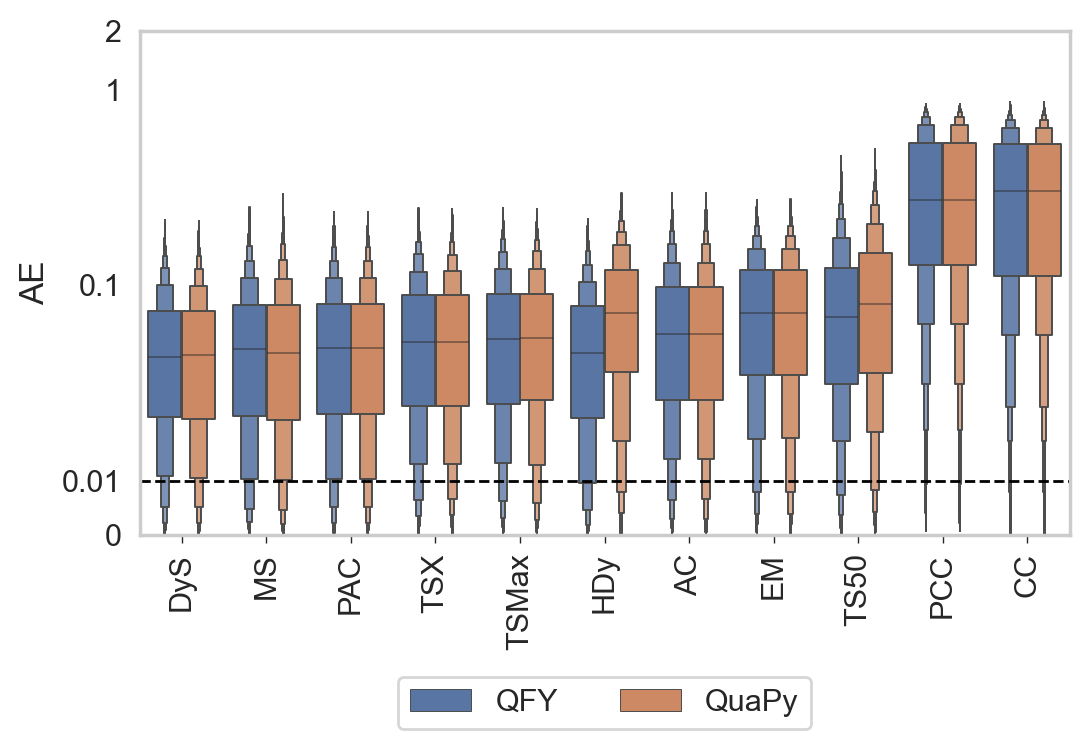}
		\caption{Comparison of implementations on binary LeQua data}
	\end{subfigure}
	\hfill
	\begin{subfigure}[b]{0.49\textwidth}
		\centering
		\includegraphics[width=\textwidth]{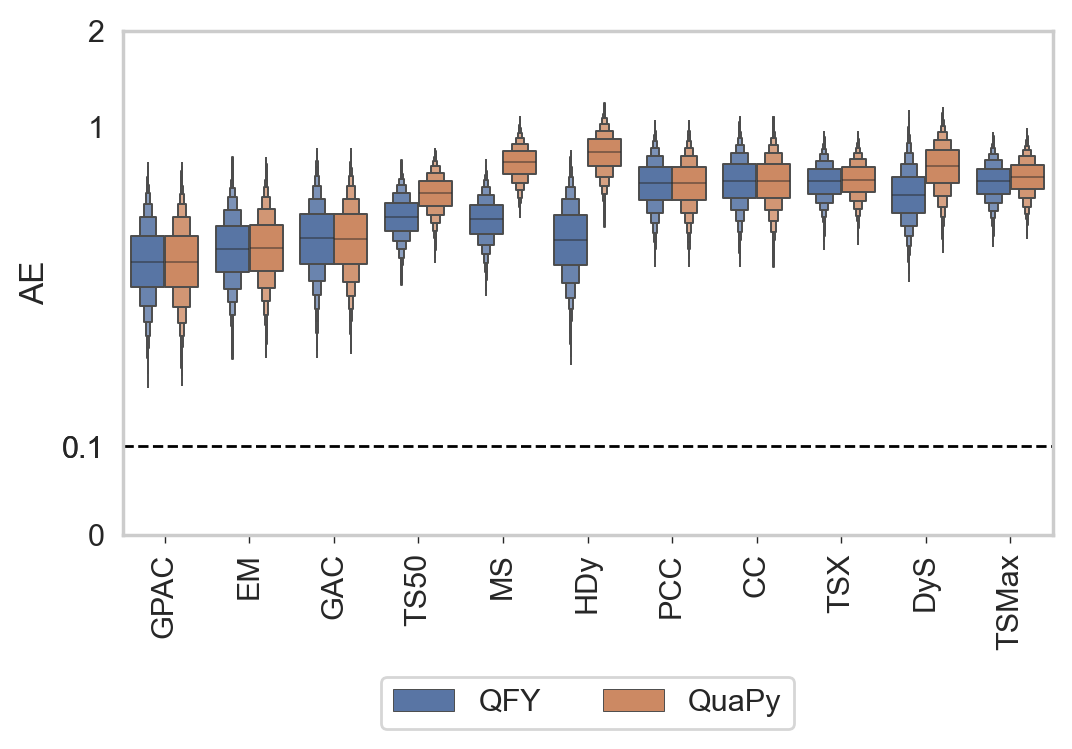}
		\caption{Comparison of implementations on multiclass LeQua data}
	\end{subfigure}
	\caption{Comparison of our implementation (\texttt{QFY}) with the \texttt{QuaPy} package. We plot the distribution of absolute error (AE) scores of all algorithms that are implemented in both codebases after applying them with the same parameterization on the LeQua data. 
    Overall, results from these packages appear either almost identical, or the results from the \texttt{QuaPy} implementation have higher AE scores than those resulting from our implementation.
 }
	\label{fig:quapy}
\end{figure}

\section{Comparison to the QuaPy Package}\label{ap:quapy}

After the publication of our initial preprint, the \texttt{QuaPy} \citep{moreo_quapy_2021} package has been published, which also implements a number of methods that are analyzed in this paper.
To further validate the correctness of our implementation, we conduct a comparison of the \texttt{QuaPy} and our \texttt{QFY} implementation. 
Toward that end, we use the dataset from the LeQua challenge (cf. Section \ref{sec:lequa}). 
We considered \texttt{QuaPy} version 0.1.9, which is currently the latest release of this package.

The methods which are included in both implementations are the \emph{CC}, \emph{PCC}, \emph{AC}, \emph{PAC}, \emph{TSX}, \emph{TSMax}, \emph{TS50}, \emph{MS}, \emph{EM}, and \emph{HDy} methods.
We leave out the SVMperf-based methods, since in our implementation, these have been adapted from an earlier implementation by the same research group that has published the \texttt{QuaPy} package \citep{esuli_lequa_2022}.
We note that in the multiclass case, the \texttt{QuaPy} implementation of \emph{AC} and \emph{PAC} rather corresponds to what we denoted as \emph{GAC} and \emph{GPAC}, since no one-vs.-rest approach is applied there, but rather a direct least-squares-based solution of the system outlined in Equation \ref{eq:dmbase}.
Further, a notable difference lies in the \texttt{QuaPy} implementation of the \emph{HDy} method, which utilizes an ensemble approach, matching distributions based on varying numbers of bins in $\{10,20,\dots,110\}$, and then returing the average prediction, as originally proposed by \citet{gonzalez-castro_class_2013}.
By contrast, in our implementation we only match distributions once, using 10 bins as default value.

In the comparison, we used the same experimental setting as in Section \ref{sec:lequa_main}.
We tried to keep the parameterization of the algorithms as consistent as possible over both implementations, including the use of the same logistic regression base classifier.

In Figure \ref{fig:quapy}, we depict the distribution of AE scores over the test samples from the LeQua data, both for the binary and multiclass versions of this challenge.
Overall, we observe that the results from our \texttt{QFY} implementation are either (close to) identical, or better than the results from the \texttt{QuaPy} package with respect to the AE.

In the binary case, the only notable difference in performance can be seen for the \emph{HDy} method, where we also identified a difference in implementation that we discussed above.
The subpar performance of the \texttt{QuaPy} implementation can likely be explained by the finding that when using more than 10 bins, the performance of this method tends to deteriorate \citep{maletzke_dys_2019}.

In the multiclass case, there are some differences in the performances of one-vs.-rest quantififiers, specifically for the \emph{TS50}, \emph{MS}, \emph{DyS}, and \emph{HDy} methods.
We suppose that these are resulting from minor differences in the implementations for the binary case, that could get amplified when normalizing binary one-vs.-rest predictions over 28 classes. 

Overall, we find that our \texttt{QFY} implementation provides similar results to the \texttt{QuaPy} implementation and---where results differ---our implementation generally tends to yield lower error scores.

\FloatBarrier

\section{Additional Plots and Tables for the Main Experiments}\label{ap:main}

In the following, we present additional analytical results regarding the ranking of algorithms. We compute the average ranks of all algorithms aggregated per data set, filtered by several conditions.
Then, we apply a Nemenyi post-hoc test at 5\% significance. 
In the individual plots, we then show for each algorithm the average performance rank over all algorithms. Horizontal bars indicate which algorithm's average rankings do not differ to a degree that is statistically significant, cf.~\cite{demsar_statistical_2006}.

\begin{figure}[t]
	\centering
	\begin{subfigure}[b]{0.49\textwidth}
		\centering
		\includegraphics[width=\textwidth]{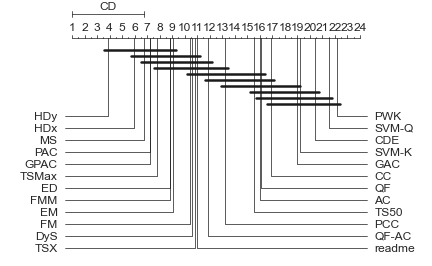}
		\caption{AE values under low shift}
	\end{subfigure}
	\hfill
	\begin{subfigure}[b]{0.49\textwidth}
		\centering
		\includegraphics[width=\textwidth]{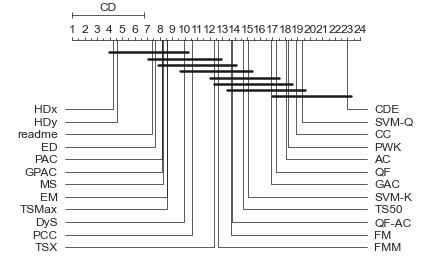}
		\caption{NKLD values under low shift}
	\end{subfigure}
	\begin{subfigure}[b]{0.49\textwidth}
		\centering
		\includegraphics[width=\textwidth]{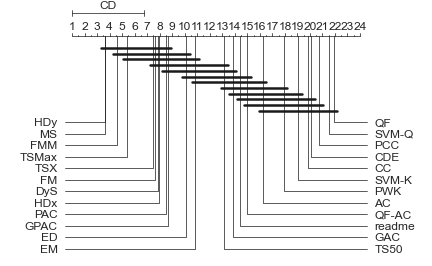}
		\caption{AE values under medium shift}
	\end{subfigure}
	\hfill
	\begin{subfigure}[b]{0.49\textwidth}
		\centering
		\includegraphics[width=\textwidth]{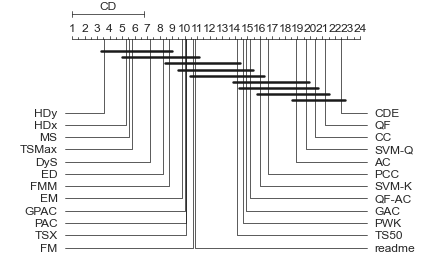}
		\caption{NKLD values under medium shift}
	\end{subfigure}
	\begin{subfigure}[b]{0.49\textwidth}
		\centering
		\includegraphics[width=\textwidth]{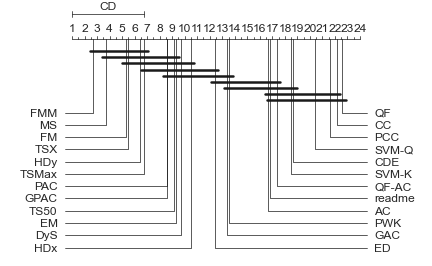}
		\caption{AE values under high shift}
	\end{subfigure}
	\hfill
	\begin{subfigure}[b]{0.49\textwidth}
		\centering
		\includegraphics[width=\textwidth]{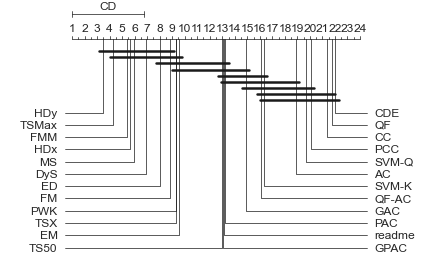}
		\caption{NKLD values under high shift}
	\end{subfigure}
	\caption{Impact of distribution shifts on algorithm rankings for \textbf{binary} labels. We  plot the distributions of rankings separated by  minor, medium, and major shifts with a Nemenyi post-hoc test at 5\% significance. }
	\label{fig:cd_shift}
\end{figure}

\begin{figure}
	\centering
	\begin{subfigure}[b]{0.49\textwidth}
		\centering
		\includegraphics[width=\textwidth]{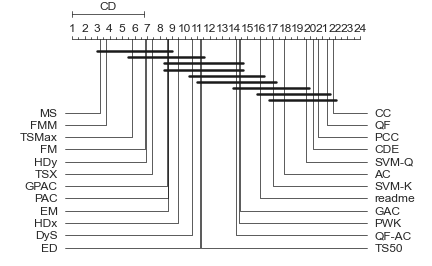}
		\caption{AE values for few training data}
	\end{subfigure}
	\hfill
	\begin{subfigure}[b]{0.49\textwidth}
		\centering
		\includegraphics[width=\textwidth]{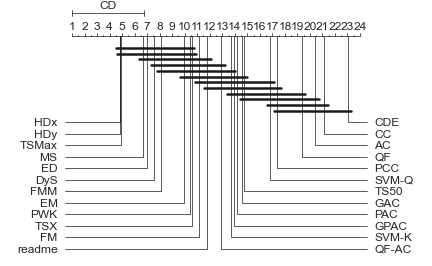}
		\caption{NKLD values for few training data}
	\end{subfigure}
	\caption{Impact of low training data amount on algorithm rankings for \textbf{binary} class labels.  We  plot the distributions of rankings obtained by 10/90 training/test splits with a Nemenyi post-hoc test at 5\% significance. }
	\label{fig:cd_fewtrain}
\end{figure}

Complementing the results from Section \ref{sec:res_binary}, Figure \ref{fig:cd_shift} displays the distributions of rankings under varying shifts between training and test data. Figure \ref{fig:cd_fewtrain} displays the rankings of quantification methods when only few training samples are given. In both figures, we observe that the rankings are very similar to the general cases. However, we observe a stronger distinction in the average ranks for high shifts and few training data.
 
\begin{figure}[ht]
	\centering
	\begin{subfigure}[b]{0.49\textwidth}
		\centering
		\includegraphics[width=\textwidth]{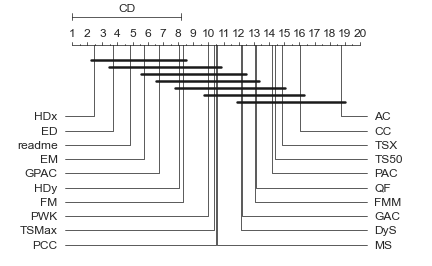}
		\caption{AE values under low shift}
	\end{subfigure}
	\hfill
	\begin{subfigure}[b]{0.49\textwidth}
		\centering
		\includegraphics[width=\textwidth]{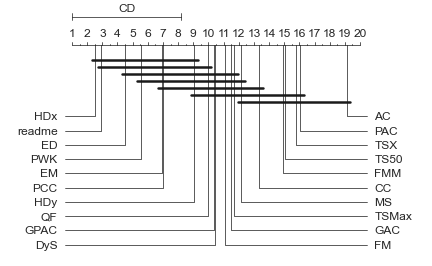}
		\caption{NKLD values under low shift}
	\end{subfigure}
	\begin{subfigure}[b]{0.49\textwidth}
		\centering
		\includegraphics[width=\textwidth]{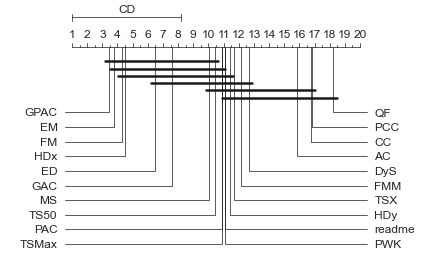}
		\caption{AE values under high shift}
	\end{subfigure}
	\hfill
	\begin{subfigure}[b]{0.49\textwidth}
		\centering
		\includegraphics[width=\textwidth]{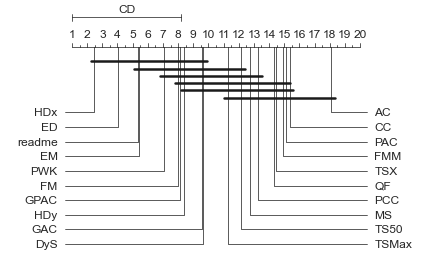}
		\caption{NKLD values under high shift}
	\end{subfigure}
	\caption{Impact of distribution shifts on algorithm rankings for \textbf{multiclass} labels. We plot the distributions of rankings separated by  minor, medium, and major shifts with a Nemenyi post-hoc test at 5\% significance.}
	\label{fig:cd_mc_shift}
\end{figure}

\begin{figure}[t]
	\centering
	\begin{subfigure}[b]{0.49\textwidth}
		\centering
		\includegraphics[width=\textwidth]{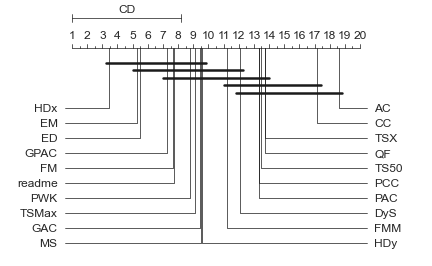}
		\caption{AE values for few training data}
	\end{subfigure}
	\hfill
	\begin{subfigure}[b]{0.49\textwidth}
		\centering
		\includegraphics[width=\textwidth]{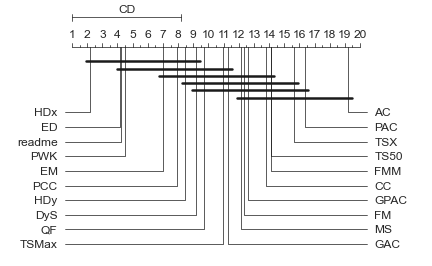}
		\caption{NKLD values for few training data}
	\end{subfigure}
	\caption{Impact of low training data amount on algorithm rankings for \textbf{multiclass} labels.  We  plot the distributions of rankings obtained by 10/90 training/test splits with a Nemenyi post-hoc test at 5\% significance. }
	\label{fig:cd_mc_fewtrain}
\end{figure}

Figure \ref{fig:cd_mc_shift} and Figure \ref{fig:cd_mc_fewtrain} complement the results from Section \ref{sec:res_mc} by presenting additional rankings in the multiclass settings. Figure \ref{fig:cd_mc_shift} displays the distributions of rankings of quantification algorithms under minor and major shifts between training and test data. Figure \ref{fig:cd_mc_fewtrain} displays the rankings of multiclass quantifiers when only settings with few training samples are considered.

Finally, we present additional results from our experiments on quantification with tuned classifiers. Table \ref{tab:clf_binary_ac}, Table \ref{tab:clf_binary_dm}, and Table \ref{tab:clf_binary_cc} display the average error scores of all considered algorithms per dataset in the binary setting, 
whereas Table \ref{tab:clf_multiclass} shows the corresponding results in the multiclass setting.

\FloatBarrier


\begin{table}
\begin{subtable} {1.0 \linewidth}
	\resizebox{\linewidth}{!}{
        \begin{tabular}{l|ccccccccccccccccccc}
\toprule
{} &     AC &  \makecell{AC\\-LR} &  \makecell{AC\\-RF} &  \makecell{AC\\-AB} &  \makecell{AC\\-SV} &    PAC & \makecell{PAC\\-LR} &    TSX & \makecell{TSX\\-LR} & \makecell{TSX\\-SV} &   TS50 &  \makecell{TS50\\-LR} &  \makecell{TS50\\-SV} &  TSMax & \makecell{TSMax\\-LR} &  \makecell{TSMax\\-SV} &              MS &  \makecell{MS\\-LR} & \makecell{MS\\-SV} \\
\midrule
bc-cat  &  0.230 &               0.087 &               0.125 &               0.128 &               0.114 &  0.112 &               0.076 &  0.077 &               0.088 &                0.08 &  0.137 &                 0.153 &                 0.129 &  0.079 &                 0.081 &                  0.079 &  \textbf{0.055} &               0.074 &              0.071 \\
bc-cont &  0.133 &               0.071 &               0.085 &               0.102 &               0.089 &  0.072 &               0.066 &  0.051 &               0.065 &               0.061 &  0.130 &                 0.128 &                 0.141 &  0.049 &                  0.06 &                  0.061 &  \textbf{0.042} &               0.061 &              0.057 \\
cars    &  0.130 &               0.091 &               0.106 &               0.087 &               0.086 &  0.080 &               0.074 &  0.063 &               0.078 &               0.061 &  0.110 &                 0.115 &                 0.107 &  0.060 &                 0.075 &                  0.061 &  \textbf{0.049} &               0.068 &              0.055 \\
conc    &  0.533 &               0.201 &               0.224 &               0.210 &               0.177 &  0.171 &               0.195 &  0.154 &               0.203 &               0.143 &  0.190 &                 0.223 &                 0.164 &  0.144 &                 0.214 &                  0.153 &  \textbf{0.121} &               0.185 &              0.131 \\
contra  &  0.613 &               0.532 &               0.430 &               0.439 &               0.549 &  0.332 &                0.48 &  0.351 &               0.539 &                 0.5 &  0.371 &                 0.535 &                 0.505 &  0.326 &                 0.563 &                  0.479 &  \textbf{0.307} &               0.545 &              0.464 \\
cappl   &  0.323 &               0.296 &               0.272 &               0.240 &               0.297 &  0.155 &               0.238 &  0.127 &                0.28 &               0.183 &  0.200 &                 0.322 &                 0.235 &  0.128 &                 0.291 &                  0.182 &  \textbf{0.104} &               0.273 &              0.162 \\
drugs   &  0.168 &               0.213 &               0.316 &               0.250 &               0.236 &  0.118 &               0.138 &  0.102 &                0.19 &               0.119 &  0.115 &                 0.196 &                 0.137 &  0.106 &                 0.221 &                  0.128 &  \textbf{0.088} &               0.193 &              0.108 \\
flare   &  0.584 &               0.601 &               0.617 &               0.630 &               0.621 &  0.344 &               0.483 &  0.353 &               0.577 &               0.527 &  0.345 &                 0.559 &                 0.531 &  0.306 &                 0.555 &                  0.509 &  \textbf{0.269} &               0.535 &              0.496 \\
grid    &  0.090 &               0.075 &               0.083 &               0.062 &               0.029 &  0.046 &                0.05 &  0.046 &               0.048 &                0.02 &  0.052 &                 0.056 &                 0.040 &  0.052 &                 0.055 &                  0.021 &           0.038 &               0.038 &     \textbf{0.017} \\
ads     &  0.175 &               0.114 &               0.156 &               0.119 &               0.139 &  0.103 &               0.095 &  0.075 &               0.094 &               0.084 &  0.113 &                 0.126 &                 0.121 &  0.067 &                  0.09 &                  0.084 &  \textbf{0.054} &               0.082 &              0.074 \\
mush    &  0.014 &               0.008 &               0.010 &               0.009 &               0.018 &  0.011 &      \textbf{0.007} &  0.008 &      \textbf{0.007} &               0.012 &  0.048 &                 0.044 &                 0.053 &  0.009 &        \textbf{0.007} &                  0.012 &  \textbf{0.007} &               0.018 &              0.016 \\
music   &  0.547 &               0.577 &               0.592 &               0.606 &               0.460 &  0.324 &               0.532 &  0.327 &               0.535 &               0.405 &  0.346 &                 0.549 &                 0.411 &  0.299 &                 0.557 &                  0.409 &  \textbf{0.272} &               0.542 &              0.386 \\
musk    &  0.110 &               0.088 &               0.129 &               0.093 &               0.066 &  0.070 &               0.074 &  0.067 &               0.076 &               0.047 &  0.080 &                 0.093 &                 0.087 &  0.068 &                 0.072 &                  0.048 &           0.058 &               0.064 &     \textbf{0.044} \\
craft   &  0.248 &               0.146 &               0.146 &               0.183 &               0.169 &  0.084 &               0.112 &  0.065 &                0.11 &               0.108 &  0.088 &                 0.120 &                 0.110 &  0.075 &                 0.127 &                  0.105 &  \textbf{0.058} &               0.110 &               0.09 \\
spam    &  0.274 &               0.060 &               0.071 &               0.056 &               0.061 &  0.069 &                0.05 &  0.047 &               0.043 &               0.041 &  0.071 &                 0.066 &                 0.067 &  0.050 &                 0.049 &                  0.044 &           0.043 &               0.048 &     \textbf{0.040} \\
alco    &  0.480 &               0.548 &               0.506 &               0.581 &               0.490 &  0.328 &               0.504 &  0.341 &                0.58 &               0.427 &  0.366 &                 0.550 &                 0.432 &  0.300 &                 0.588 &                  0.416 &  \textbf{0.277} &               0.568 &              0.397 \\
study   &  0.347 &               0.197 &               0.190 &               0.174 &               0.199 &  0.187 &               0.189 &  0.201 &               0.192 &               0.176 &  0.215 &                 0.197 &                 0.186 &  0.194 &                 0.214 &                  0.183 &           0.161 &               0.182 &     \textbf{0.147} \\
telco   &  0.224 &               0.226 &               0.206 &               0.217 &               0.276 &  0.075 &               0.115 &  0.071 &               0.141 &               0.232 &  0.080 &                 0.159 &                 0.244 &  0.069 &                 0.149 &                  0.216 &  \textbf{0.060} &               0.137 &               0.21 \\
thrm    &  0.612 &               0.486 &               0.461 &               0.496 &               0.454 &  0.318 &               0.451 &  0.320 &               0.445 &               0.418 &  0.355 &                 0.468 &                 0.418 &  0.298 &                  0.46 &                  0.405 &  \textbf{0.272} &               0.438 &              0.376 \\
turk    &  0.619 &               0.653 &               0.652 &               0.693 &               0.575 &  0.248 &                0.36 &  0.282 &               0.523 &               0.581 &  0.283 &                 0.542 &                 0.579 &  0.240 &                 0.531 &                  0.517 &  \textbf{0.239} &               0.531 &              0.517 \\
vgame   &  0.209 &               0.192 &               0.256 &               0.233 &               0.201 &  0.085 &               0.156 &  0.088 &               0.186 &               0.135 &  0.086 &                 0.186 &                 0.134 &  0.091 &                 0.198 &                  0.133 &  \textbf{0.076} &               0.190 &              0.122 \\
voice   &  0.150 &               0.029 &               0.033 &               0.031 &               0.031 &  0.048 &               0.026 &  0.035 &               0.023 &      \textbf{0.022} &  0.060 &                 0.063 &                 0.069 &  0.032 &                 0.024 &                  0.025 &           0.034 &               0.033 &              0.025 \\
wine    &  0.479 &               0.286 &               0.185 &               0.338 &               0.232 &  0.095 &               0.198 &  0.091 &               0.228 &               0.172 &  0.093 &                 0.229 &                 0.170 &  0.096 &                 0.259 &                  0.168 &  \textbf{0.081} &               0.239 &              0.155 \\
yeast   &  0.681 &               0.513 &               0.365 &               0.421 &               0.425 &  0.238 &               0.449 &  0.276 &               0.475 &               0.378 &  0.306 &                 0.501 &                 0.397 &  0.234 &                 0.526 &                  0.372 &  \textbf{0.212} &               0.517 &              0.353 \\\midrule
Mean    &  0.332 &               0.262 &               0.259 &               0.267 &               0.250 &  0.155 &               0.213 &  0.151 &               0.239 &               0.205 &  0.177 &                 0.258 &                 0.228 &  0.140 &                 0.249 &                  0.200 &  \textbf{0.124} &               0.236 &              0.188 \\
\bottomrule
\end{tabular}

    }
	\subcaption{Average AE scores of adjusted count-based quantifiers}
	\label{tab:clf_binary_AC_AE}
\end{subtable}

\begin{subtable} {1.0 \linewidth}
	\resizebox{\linewidth}{!}{
        \begin{tabular}{l|ccccccccccccccccccc}
\toprule
{} &     AC & \makecell{AC\\-LR} &  \makecell{AC\\-RF} &  \makecell{AC\\-AB} &  \makecell{AC\\-SV} &             PAC & \makecell{PAC\\-LR} &             TSX & \makecell{TSX\\-LR} &  \makecell{TSX\\-SV} &   TS50 &  \makecell{TS50\\-LR} &  \makecell{TS50\\-SV} &           TSMax & \makecell{TSMax\\-LR} & \makecell{TSMax\\-SV} &              MS &  \makecell{MS\\-LR} &  \makecell{MS\\-SV} \\
\midrule
bc-cat  &  0.161 &              0.031 &               0.066 &               0.071 &               0.052 &           0.065 &               0.038 &           0.024 &               0.039 &                0.028 &  0.088 &                 0.092 &                 0.071 &           0.017 &                 0.025 &                  0.02 &  \textbf{0.015} &               0.030 &               0.027 \\
bc-cont &  0.084 &              0.024 &               0.036 &               0.044 &               0.034 &            0.04 &                0.04 &           0.013 &               0.022 &                0.017 &  0.081 &                 0.077 &                 0.087 &  \textbf{0.010} &                 0.018 &                 0.014 &           0.019 &               0.032 &               0.030 \\
cars    &  0.074 &              0.043 &               0.059 &               0.040 &               0.038 &           0.051 &               0.038 &           0.028 &               0.039 &                0.027 &  0.057 &                 0.059 &                 0.051 &  \textbf{0.016} &                  0.03 &                 0.019 &           0.019 &               0.034 &               0.023 \\
conc    &  0.459 &              0.137 &               0.151 &               0.131 &               0.100 &            0.13 &               0.116 &           0.089 &               0.126 &                0.077 &  0.125 &                 0.151 &                 0.090 &  \textbf{0.052} &                 0.116 &                 0.061 &            0.06 &               0.114 &               0.065 \\
contra  &  0.537 &              0.449 &               0.321 &               0.356 &               0.422 &           0.247 &               0.303 &           0.258 &               0.436 &                0.375 &  0.271 &                 0.436 &                 0.388 &           0.175 &                 0.412 &                 0.326 &  \textbf{0.172} &               0.411 &               0.325 \\
cappl   &  0.238 &              0.201 &               0.187 &               0.152 &               0.205 &           0.093 &               0.158 &           0.061 &               0.192 &                0.107 &  0.128 &                 0.237 &                 0.155 &  \textbf{0.036} &                 0.184 &                  0.09 &            0.04 &               0.192 &               0.091 \\
drugs   &  0.093 &              0.145 &               0.254 &               0.171 &               0.165 &           0.057 &               0.085 &           0.041 &               0.118 &                0.058 &  0.059 &                 0.128 &                 0.069 &  \textbf{0.025} &                 0.121 &                 0.051 &           0.031 &               0.119 &               0.052 \\
flare   &  0.436 &              0.482 &               0.483 &               0.488 &               0.476 &           0.247 &               0.338 &           0.251 &               0.467 &                0.387 &  0.259 &                 0.462 &                 0.400 &           0.152 &                 0.442 &                 0.339 &  \textbf{0.151} &               0.440 &               0.341 \\
grid    &  0.041 &              0.027 &               0.051 &               0.021 &               0.007 &           0.015 &               0.017 &           0.009 &               0.011 &                0.004 &  0.010 &                 0.013 &                 0.006 &           0.007 &                 0.009 &        \textbf{0.003} &           0.007 &               0.008 &               0.004 \\
ads     &  0.112 &              0.062 &               0.095 &               0.059 &               0.081 &           0.074 &               0.063 &           0.035 &               0.053 &                0.041 &  0.070 &                 0.079 &                 0.069 &  \textbf{0.016} &                 0.042 &                 0.032 &           0.021 &               0.049 &               0.035 \\
mush    &  0.002 &     \textbf{0.001} &               0.002 &               0.002 &               0.008 &  \textbf{0.001} &      \textbf{0.001} &  \textbf{0.001} &      \textbf{0.001} &                0.007 &  0.012 &                 0.012 &                 0.019 &  \textbf{0.001} &        \textbf{0.001} &                 0.005 &  \textbf{0.001} &               0.004 &               0.005 \\
music   &  0.435 &              0.472 &               0.475 &               0.460 &               0.349 &           0.248 &               0.343 &           0.223 &               0.443 &                0.293 &  0.242 &                 0.450 &                 0.306 &           0.147 &                 0.433 &                  0.26 &  \textbf{0.142} &               0.433 &               0.262 \\
musk    &  0.057 &              0.039 &               0.076 &               0.042 &               0.028 &           0.029 &                0.03 &           0.029 &               0.036 &                0.016 &  0.036 &                 0.051 &                 0.046 &           0.016 &                 0.023 &        \textbf{0.010} &           0.019 &               0.025 &               0.013 \\
craft   &  0.179 &              0.087 &               0.077 &               0.121 &               0.091 &           0.049 &               0.057 &            0.02 &               0.059 &                0.051 &  0.042 &                 0.066 &                 0.051 &  \textbf{0.014} &                 0.053 &                 0.033 &            0.02 &               0.055 &               0.039 \\
spam    &  0.220 &              0.022 &               0.033 &               0.020 &               0.018 &           0.036 &               0.018 &           0.011 &               0.016 &                0.013 &  0.031 &                 0.029 &                 0.026 &  \textbf{0.009} &                 0.015 &        \textbf{0.009} &           0.012 &               0.016 &               0.010 \\
alco    &  0.365 &              0.452 &               0.407 &               0.457 &               0.374 &           0.254 &               0.337 &           0.259 &               0.456 &                0.323 &  0.280 &                 0.452 &                 0.329 &  \textbf{0.155} &                 0.428 &                 0.276 &           0.159 &               0.432 &               0.276 \\
study   &  0.264 &              0.121 &               0.103 &               0.088 &               0.114 &           0.115 &               0.104 &           0.106 &                0.11 &                0.092 &  0.129 &                 0.122 &                 0.106 &           0.071 &                 0.103 &                 0.079 &  \textbf{0.069} &               0.102 &               0.075 \\
telco   &  0.152 &              0.172 &               0.148 &               0.160 &               0.216 &           0.038 &               0.075 &           0.032 &                 0.1 &                0.167 &  0.035 &                 0.110 &                 0.177 &  \textbf{0.015} &                  0.09 &                 0.146 &           0.024 &               0.092 &               0.149 \\
thrm    &  0.505 &              0.366 &               0.310 &               0.342 &               0.301 &           0.252 &               0.296 &           0.235 &               0.363 &                0.300 &  0.267 &                 0.377 &                 0.300 &  \textbf{0.169} &                 0.335 &                 0.235 &           0.174 &               0.341 &               0.232 \\
turk    &  0.527 &              0.546 &               0.557 &               0.586 &               0.438 &           0.194 &               0.278 &           0.197 &               0.418 &                0.456 &  0.206 &                 0.439 &                 0.459 &           0.113 &                 0.404 &                 0.375 &  \textbf{0.112} &               0.404 &               0.375 \\
vgame   &  0.152 &              0.132 &               0.172 &               0.154 &               0.129 &           0.045 &               0.096 &           0.038 &               0.131 &                0.076 &  0.036 &                 0.126 &                 0.077 &  \textbf{0.026} &                 0.124 &                 0.064 &           0.028 &               0.128 &               0.066 \\
voice   &  0.107 &              0.008 &               0.010 &               0.008 &               0.007 &           0.025 &               0.007 &           0.013 &               0.004 &                0.007 &  0.021 &                 0.021 &                 0.026 &           0.006 &        \textbf{0.003} &                 0.004 &            0.01 &               0.009 &               0.009 \\
wine    &  0.419 &              0.201 &               0.096 &               0.245 &               0.136 &            0.05 &               0.118 &           0.039 &               0.172 &                0.099 &  0.036 &                 0.173 &                 0.097 &  \textbf{0.024} &                 0.175 &                 0.081 &           0.026 &               0.174 &               0.081 \\
yeast   &  0.595 &              0.425 &               0.258 &               0.304 &               0.319 &            0.18 &               0.297 &           0.198 &               0.418 &                0.266 &  0.225 &                 0.444 &                 0.289 &  \textbf{0.111} &                 0.393 &                 0.229 &           0.115 &               0.406 &               0.230 \\\midrule
Mean    &  0.259 &              0.194 &               0.184 &               0.188 &               0.171 &           0.106 &               0.136 &           0.092 &               0.176 &                0.137 &  0.114 &                 0.192 &                 0.154 &  \textbf{0.058} &                 0.166 &                 0.115 &            0.06 &               0.169 &               0.117 \\
\bottomrule
\end{tabular}

    }
\subcaption{Average NKLD scores of adjusted count-based quantifiers}
	\label{tab:clf_binary_AC_NKLD}
\end{subtable}

\caption{Results of adjusted count-based quantifiers with tuned base classifiers in the in the \textbf{binary} setting, where the base classifiers were tuned with respect to their accuracy. We show the averaged error scores for all scenarios per algorithm and dataset. Algorithms based on untuned logistic regression classifiers are denoted as before (no suffix), alternative tuned base classifiers are marked with respective suffixes: logistic regressors (LR), support vector machines (SV), random forests (RF) and AdaBoost (AB).
Overall, it does not appear that tuning base classifiers is consistently beneficial for quantification with these methods.}
\label{tab:clf_binary_ac}
\end{table}

\begin{table}
\begin{subtable} {1.0 \linewidth}
	\resizebox{\linewidth}{!}{
        \begin{tabular}{l|cccccccccccccccccccccc}
\toprule
{} &    GAC &  \makecell{GAC\\-LR} &  \makecell{GAC\\-RF} &  \makecell{GAC\\-AB} &  \makecell{GAC\\-SV} &   GPAC & \makecell{GPAC\\-LR} &    DyS &  \makecell{DyS\\-LR} & \makecell{DyS\\-SV} &             FMM &  \makecell{FMM\\-LR} & \makecell{FMM\\-SV} &             HDy & \makecell{HDy\\-LR} & \makecell{HDy\\-SV} &              FM & \makecell{FM\\-LR} &              EM &  \makecell{EM\\-LR} &    CDE &  \makecell{CDE\\-LR} \\
\midrule
bc-cat  &  0.193 &                0.084 &                0.124 &                0.127 &                0.107 &  0.112 &                0.076 &  0.121 &                0.118 &               0.103 &  \textbf{0.056} &                0.064 &               0.065 &           0.083 &               0.084 &               0.096 &           0.062 &              0.106 &           0.207 &               0.195 &  0.315 &                0.145 \\
bc-cont &  0.117 &                0.072 &                0.090 &                0.108 &                0.080 &  0.072 &                0.066 &  0.106 &                0.070 &               0.099 &           0.048 &                0.060 &               0.064 &           0.056 &               0.058 &               0.087 &  \textbf{0.039} &              0.132 &           0.125 &               0.262 &  0.123 &                0.176 \\
cars    &  0.113 &                0.083 &                0.093 &                0.077 &                0.079 &  0.080 &                0.074 &  0.078 &                0.069 &               0.071 &  \textbf{0.051} &                0.063 &      \textbf{0.051} &           0.059 &               0.059 &               0.061 &           0.059 &              0.078 &           0.087 &               0.086 &  0.180 &                0.098 \\
conc    &  0.369 &                0.194 &                0.216 &                0.206 &                0.177 &  0.171 &                0.193 &  0.175 &                0.172 &               0.172 &  \textbf{0.125} &                0.174 &      \textbf{0.125} &           0.178 &               0.156 &               0.147 &           0.155 &              0.187 &           0.336 &               0.216 &  0.745 &                0.284 \\
contra  &  0.472 &                0.438 &                0.369 &                0.370 &                0.445 &  0.331 &                0.479 &  0.434 &                0.448 &               0.538 &           0.297 &                0.505 &               0.455 &             0.4 &               0.416 &               0.489 &           0.351 &              0.505 &  \textbf{0.249} &               0.422 &  0.881 &                0.830 \\
cappl   &  0.289 &                0.247 &                0.237 &                0.229 &                0.258 &  0.156 &                0.238 &  0.205 &                0.250 &                0.26 &           0.109 &                0.240 &               0.155 &           0.172 &                0.23 &               0.228 &           0.115 &              0.333 &  \textbf{0.087} &               0.413 &  0.302 &                0.416 \\
drugs   &  0.174 &                0.185 &                0.261 &                0.227 &                0.208 &  0.119 &                0.138 &  0.144 &                0.174 &               0.142 &  \textbf{0.080} &                0.194 &               0.107 &           0.101 &               0.174 &               0.131 &           0.104 &              0.219 &           0.134 &               0.187 &  0.134 &                0.312 \\
flare   &  0.482 &                0.437 &                0.464 &                0.526 &                0.462 &  0.342 &                0.483 &  0.454 &                0.476 &                0.64 &           0.291 &                0.510 &                0.46 &           0.416 &               0.428 &                0.54 &           0.346 &              0.643 &  \textbf{0.256} &               0.547 &  0.675 &                0.721 \\
grid    &  0.086 &                0.075 &                0.080 &                0.059 &                0.028 &  0.046 &                 0.05 &  0.042 &                0.034 &      \textbf{0.015} &           0.035 &                0.038 &               0.016 &           0.033 &               0.035 &      \textbf{0.015} &           0.044 &              0.051 &           0.048 &               0.068 &  0.258 &                0.213 \\
ads     &  0.138 &                0.090 &                0.122 &                0.116 &                0.112 &  0.102 &                0.095 &  0.106 &                0.092 &               0.095 &  \textbf{0.060} &                0.082 &                0.08 &           0.077 &                0.08 &               0.091 &           0.082 &              0.162 &           0.087 &               0.383 &  0.199 &                0.282 \\
mush    &  0.014 &                0.008 &                0.010 &                0.009 &                0.018 &  0.011 &       \textbf{0.007} &  0.014 &                0.009 &               0.025 &           0.016 &                0.015 &               0.021 &  \textbf{0.007} &      \textbf{0.007} &               0.015 &           0.008 &     \textbf{0.007} &           0.017 &               0.015 &  0.009 &                0.008 \\
music   &  0.462 &                0.429 &                0.440 &                0.497 &                0.387 &  0.324 &                0.532 &  0.429 &                0.471 &               0.416 &           0.283 &                0.542 &               0.366 &           0.371 &               0.436 &                 0.4 &           0.328 &              0.555 &  \textbf{0.257} &               0.479 &  0.840 &                0.809 \\
musk    &  0.096 &                0.078 &                0.105 &                0.087 &                0.062 &  0.069 &                0.074 &  0.073 &                0.067 &               0.051 &           0.053 &                0.062 &      \textbf{0.044} &           0.058 &               0.061 &                0.05 &           0.068 &              0.074 &           0.065 &               0.102 &  0.188 &                0.130 \\
craft   &  0.219 &                0.144 &                0.142 &                0.176 &                0.164 &  0.084 &                0.112 &  0.082 &                0.196 &                0.11 &  \textbf{0.053} &                0.086 &               0.098 &           0.058 &               0.189 &               0.099 &           0.067 &              0.113 &           0.144 &               0.096 &  0.528 &                0.276 \\
spam    &  0.236 &                0.059 &                0.069 &                0.057 &                0.060 &  0.069 &                 0.05 &  0.072 &                0.065 &               0.052 &           0.041 &                0.045 &      \textbf{0.039} &           0.042 &               0.054 &               0.046 &           0.047 &              0.046 &           0.265 &               0.067 &  0.603 &                0.074 \\
alco    &  0.451 &                0.425 &                0.407 &                0.477 &                0.415 &  0.337 &                0.504 &  0.431 &                0.437 &                0.42 &  \textbf{0.282} &                0.547 &               0.369 &            0.36 &               0.433 &               0.395 &           0.342 &              0.566 &           0.296 &               0.457 &  0.695 &                0.720 \\
study   &  0.301 &                0.188 &                0.182 &                0.176 &                0.189 &  0.187 &                0.188 &  0.233 &                0.156 &               0.161 &           0.162 &                0.162 &      \textbf{0.140} &           0.194 &               0.153 &               0.154 &           0.192 &              0.191 &           0.175 &               0.167 &  0.533 &                0.205 \\
telco   &  0.211 &                0.188 &                0.174 &                0.186 &                0.227 &  0.075 &                0.115 &  0.075 &                0.142 &               0.225 &  \textbf{0.056} &                0.135 &               0.205 &           0.059 &               0.112 &               0.199 &            0.07 &              0.155 &           0.059 &               0.122 &  0.401 &                0.428 \\
thrm    &  0.462 &                0.369 &                0.372 &                0.449 &                0.389 &  0.318 &                0.451 &  0.423 &                0.409 &               0.456 &           0.291 &                0.440 &               0.372 &           0.358 &               0.361 &               0.423 &           0.309 &              0.479 &  \textbf{0.266} &               0.419 &  0.861 &                0.688 \\
turk    &  0.477 &                0.451 &                0.455 &                0.477 &                0.438 &  0.246 &                0.359 &  0.303 &                0.492 &               0.548 &           0.219 &                0.488 &               0.484 &           0.281 &               0.483 &               0.502 &            0.28 &              0.558 &  \textbf{0.164} &               0.460 &  0.881 &                0.878 \\
vgame   &  0.209 &                0.163 &                0.215 &                0.192 &                0.175 &  0.085 &                0.156 &  0.090 &                0.147 &               0.137 &           0.075 &                0.177 &               0.123 &           0.084 &               0.145 &               0.129 &           0.089 &              0.182 &  \textbf{0.066} &               0.188 &  0.586 &                0.348 \\
voice   &  0.134 &                0.030 &                0.033 &                0.031 &                0.031 &  0.047 &                0.026 &  0.037 &                0.024 &               0.022 &           0.038 &                0.028 &               0.028 &            0.03 &      \textbf{0.021} &               0.022 &           0.036 &              0.025 &           0.178 &               0.067 &  0.289 &                0.050 \\
wine    &  0.372 &                0.230 &                0.176 &                0.260 &                0.207 &  0.095 &                0.198 &  0.140 &                0.186 &               0.183 &  \textbf{0.079} &                0.236 &               0.158 &           0.096 &               0.183 &                0.17 &           0.102 &              0.241 &           0.233 &               0.201 &  0.815 &                0.650 \\
yeast   &  0.471 &                0.414 &                0.328 &                0.380 &                0.374 &  0.241 &                0.445 &  0.338 &                0.435 &               0.422 &  \textbf{0.221} &                0.512 &               0.339 &           0.273 &               0.387 &               0.377 &           0.261 &              0.469 &            0.38 &               0.404 &  0.873 &                0.743 \\\midrule
Mean    &  0.273 &                0.212 &                0.215 &                0.229 &                0.212 &  0.155 &                0.213 &  0.192 &                0.214 &               0.224 &  \textbf{0.126} &                0.225 &               0.182 &            0.16 &               0.198 &               0.203 &           0.148 &              0.253 &           0.174 &               0.251 &  0.496 &                0.395 \\
\bottomrule
\end{tabular}

    }
	\subcaption{Average AE scores of adjusted count-based quantifiers}
	\label{tab:clf_binary_DM_AE}
\end{subtable}

\begin{subtable} {1.0 \linewidth}
	\resizebox{\linewidth}{!}{
        \begin{tabular}{l|cccccccccccccccccccccc}
\toprule
{} &    GAC &  \makecell{GAC\\-LR} &  \makecell{GAC\\-RF} &  \makecell{GAC\\-AB} &  \makecell{GAC\\-SV} &   GPAC &  \makecell{GPAC\\-LR} &    DyS & \makecell{DyS\\-LR} & \makecell{DyS\\-SV} &             FMM &  \makecell{FMM\\-LR} &  \makecell{FMM\\-SV} &             HDy & \makecell{HDy\\-LR} & \makecell{HDy\\-SV} &     FM &  \makecell{FM\\-LR} &              EM &  \makecell{EM\\-LR} &    CDE &  \makecell{CDE\\-LR} \\
\midrule
bc-cat  &  0.089 &                0.026 &                0.062 &                0.072 &                0.030 &  0.065 &                 0.037 &  0.038 &               0.028 &               0.026 &  \textbf{0.016} &                0.026 &                0.026 &           0.018 &               0.018 &               0.028 &  0.023 &               0.046 &            0.16 &               0.120 &  0.409 &                0.135 \\
bc-cont &  0.052 &                0.023 &                0.037 &                0.048 &                0.023 &  0.040 &                 0.040 &  0.022 &               0.013 &               0.019 &           0.024 &                0.037 &                0.037 &  \textbf{0.007} &               0.009 &               0.018 &  0.015 &               0.045 &           0.087 &               0.188 &  0.184 &                0.231 \\
cars    &  0.051 &                0.030 &                0.037 &                0.027 &                0.027 &  0.049 &                 0.037 &  0.021 &               0.014 &               0.017 &           0.019 &                0.030 &                0.018 &           0.013 &      \textbf{0.011} &               0.013 &  0.030 &               0.039 &           0.034 &               0.030 &  0.212 &                0.058 \\
conc    &  0.156 &                0.082 &                0.114 &                0.105 &                0.067 &  0.130 &                 0.109 &  0.077 &               0.057 &               0.058 &           0.067 &                0.103 &                0.059 &            0.07 &      \textbf{0.044} &               0.048 &  0.091 &               0.111 &           0.325 &               0.097 &  0.799 &                0.295 \\
contra  &  0.242 &                0.204 &                0.177 &                0.152 &                0.191 &  0.247 &                 0.297 &  0.245 &               0.211 &               0.291 &           0.199 &                0.404 &                0.321 &           0.203 &               0.166 &               0.264 &  0.260 &               0.410 &  \textbf{0.125} &               0.181 &  0.843 &                0.813 \\
cappl   &  0.156 &                0.093 &                0.102 &                0.109 &                0.103 &  0.095 &                 0.151 &  0.075 &               0.098 &               0.104 &           0.045 &                0.169 &                0.084 &           0.057 &               0.078 &               0.088 &  0.054 &               0.202 &  \textbf{0.037} &               0.280 &  0.415 &                0.489 \\
drugs   &  0.093 &                0.068 &                0.104 &                0.099 &                0.073 &  0.057 &                 0.083 &  0.037 &               0.077 &               0.042 &           0.028 &                0.122 &                0.051 &  \textbf{0.019} &               0.071 &               0.036 &  0.044 &               0.126 &           0.022 &               0.088 &  0.094 &                0.429 \\
flare   &  0.296 &                0.180 &                0.200 &                0.294 &                0.188 &  0.244 &                 0.330 &  0.217 &               0.237 &               0.368 &           0.178 &                0.418 &                0.303 &           0.192 &               0.192 &               0.297 &  0.234 &               0.438 &  \textbf{0.081} &               0.304 &  0.711 &                0.737 \\
grid    &  0.034 &                0.027 &                0.043 &                0.018 &                0.006 &  0.015 &                 0.017 &  0.005 &               0.003 &      \textbf{0.001} &           0.005 &                0.007 &                0.004 &           0.002 &               0.003 &      \textbf{0.001} &  0.009 &               0.012 &           0.014 &               0.038 &  0.414 &                0.326 \\
ads     &  0.078 &                0.034 &                0.063 &                0.056 &                0.043 &  0.074 &                 0.062 &  0.033 &               0.024 &               0.027 &           0.024 &                0.046 &                0.036 &           0.018 &      \textbf{0.017} &               0.025 &  0.039 &               0.079 &           0.027 &               0.262 &  0.187 &                0.268 \\
mush    &  0.002 &                0.001 &                0.003 &                0.002 &                0.008 &  0.001 &                 0.001 &  0.001 &      \textbf{0.000} &               0.005 &           0.004 &                0.003 &                0.007 &  \textbf{0.000} &      \textbf{0.000} &               0.003 &  0.001 &               0.001 &           0.001 &               0.004 &  0.004 &                0.005 \\
music   &  0.258 &                0.187 &                0.192 &                0.276 &                0.179 &  0.248 &                 0.339 &  0.207 &               0.235 &               0.213 &           0.172 &                0.433 &                0.249 &           0.168 &               0.187 &               0.201 &  0.224 &               0.421 &  \textbf{0.082} &               0.216 &  0.829 &                0.780 \\
musk    &  0.045 &                0.030 &                0.051 &                0.036 &                0.024 &  0.028 &                 0.029 &  0.017 &               0.013 &               0.008 &           0.016 &                0.024 &                0.012 &  \textbf{0.007} &                0.01 &      \textbf{0.007} &  0.028 &               0.033 &           0.011 &               0.023 &  0.198 &                0.063 \\
craft   &  0.106 &                0.066 &                0.066 &                0.087 &                0.068 &  0.049 &                 0.055 &  0.021 &               0.137 &               0.029 &           0.018 &                0.049 &                0.041 &  \textbf{0.008} &                0.13 &               0.023 &  0.027 &               0.055 &           0.089 &               0.036 &  0.733 &                0.339 \\
spam    &  0.121 &                0.017 &                0.030 &                0.020 &                0.016 &  0.036 &                 0.018 &  0.025 &                0.01 &               0.007 &           0.011 &                0.016 &                0.013 &  \textbf{0.004} &               0.008 &               0.007 &  0.013 &               0.015 &           0.218 &               0.013 &  0.718 &                0.023 \\
alco    &  0.279 &                0.183 &                0.182 &                0.241 &                0.177 &  0.260 &                 0.332 &  0.207 &               0.212 &               0.211 &           0.192 &                0.428 &                0.249 &           0.176 &               0.184 &               0.199 &  0.262 &               0.429 &  \textbf{0.102} &               0.200 &  0.783 &                0.723 \\
study   &  0.145 &                0.077 &                0.078 &                0.085 &                0.073 &  0.115 &                 0.101 &  0.095 &                0.05 &               0.055 &           0.078 &                0.087 &                0.068 &           0.075 &      \textbf{0.048} &               0.055 &  0.103 &               0.101 &           0.084 &               0.053 &  0.689 &                0.121 \\
telco   &  0.120 &                0.074 &                0.075 &                0.070 &                0.087 &  0.040 &                 0.073 &  0.016 &               0.076 &               0.114 &           0.021 &                0.092 &                0.142 &           0.011 &               0.051 &               0.102 &  0.032 &               0.094 &  \textbf{0.007} &               0.044 &  0.532 &                0.561 \\
thrm    &  0.224 &                0.160 &                0.182 &                0.258 &                0.171 &  0.251 &                 0.292 &  0.222 &               0.188 &               0.231 &             0.2 &                0.343 &                0.241 &           0.183 &      \textbf{0.143} &                0.21 &  0.221 &               0.354 &           0.191 &               0.195 &  0.837 &                0.641 \\
turk    &  0.247 &                0.182 &                0.184 &                0.192 &                0.178 &  0.192 &                 0.267 &  0.133 &               0.306 &               0.321 &           0.138 &                0.389 &                0.352 &           0.109 &                0.28 &               0.291 &  0.207 &               0.416 &  \textbf{0.048} &               0.195 &  0.843 &                0.840 \\
vgame   &  0.131 &                0.056 &                0.099 &                0.075 &                0.063 &  0.045 &                 0.094 &  0.020 &               0.062 &               0.053 &            0.03 &                0.124 &                0.066 &           0.019 &               0.056 &                0.05 &  0.040 &               0.126 &  \textbf{0.013} &               0.062 &  0.763 &                0.296 \\
voice   &  0.067 &                0.008 &                0.009 &                0.008 &                0.007 &  0.024 &                 0.007 &  0.006 &      \textbf{0.002} &      \textbf{0.002} &           0.014 &                0.008 &                0.010 &  \textbf{0.002} &      \textbf{0.002} &      \textbf{0.002} &  0.014 &               0.007 &           0.121 &               0.023 &  0.467 &                0.032 \\
wine    &  0.164 &                0.080 &                0.072 &                0.102 &                0.069 &  0.049 &                 0.115 &  0.057 &               0.076 &               0.073 &           0.032 &                0.170 &                0.085 &  \textbf{0.020} &               0.069 &               0.064 &  0.048 &               0.168 &           0.211 &               0.093 &  0.831 &                0.786 \\
yeast   &  0.200 &                0.180 &                0.153 &                0.188 &                0.163 &  0.183 &                 0.285 &  0.179 &               0.201 &               0.215 &           0.133 &                0.402 &                0.231 &  \textbf{0.115} &               0.155 &               0.186 &  0.190 &               0.386 &           0.373 &               0.188 &  0.842 &                0.778 \\\midrule
Mean    &  0.140 &                0.086 &                0.097 &                0.109 &                0.085 &  0.106 &                 0.132 &  0.082 &               0.097 &               0.104 &           0.069 &                0.164 &                0.113 &  \textbf{0.062} &                0.08 &               0.092 &  0.092 &               0.171 &           0.103 &               0.122 &  0.556 &                0.407 \\
\bottomrule
\end{tabular}

    }
\subcaption{Average NKLD scores of adjusted count-based quantifiers}
	\label{tab:clf_binary_DM_NKLD}
\end{subtable}

\caption{Results of distribution matching-based quantifiers in the \textbf{binary} setting, where the base classifiers were tuned with respect to their accuracy. We show the averaged error scores for all scenarios per algorithm and dataset. Algorithms based on untuned logistic regression classifiers are denoted as before (no suffix), alternative tuned base classifiers are marked with respective suffixes: logistic regressors (LR), support vector machines (SV), random forests (RF) and AdaBoost (AB). Overall, it does not appear that tuning base classifiers is consistently beneficial for quantification with these methods.}
\label{tab:clf_binary_dm}
\end{table}

\begin{table}

\begin{subtable} {0.49 \linewidth}
	\resizebox{\linewidth}{!}{
        \begin{tabular}{l|ccccccccccc}
\toprule
{} &     CC & \makecell{CC\\-LR} &  \makecell{CC\\-RF} & \makecell{CC\\-AB} & \makecell{CC\\-SV} &    PCC & \makecell{PCC\\-LR} & \makecell{SVM\\-K} & \makecell{SVM\\-Q} & \makecell{RBF\\-K} & \makecell{RBF\\-Q} \\
\midrule
bc-cat  &  0.380 &     \textbf{0.127} &               0.207 &              0.174 &              0.166 &  0.390 &               0.202 &              0.304 &              0.753 &              0.146 &              0.202 \\
bc-cont &  0.172 &              0.084 &               0.116 &               0.14 &              0.107 &  0.245 &               0.251 &              0.167 &              0.838 &               0.08 &     \textbf{0.066} \\
cars    &  0.299 &              0.181 &               0.195 &              0.181 &     \textbf{0.140} &  0.306 &               0.195 &              0.228 &              0.227 &              0.499 &               0.54 \\
conc    &  0.699 &              0.434 &               0.454 &              0.421 &               0.37 &  0.608 &               0.446 &              0.304 &              0.601 &     \textbf{0.279} &              0.507 \\
contra  &  0.814 &              0.777 &               0.716 &              0.718 &              0.771 &  0.672 &               0.662 &     \textbf{0.565} &              0.802 &              0.579 &              0.719 \\
cappl   &  0.473 &              0.426 &               0.422 &              0.383 &              0.431 &  0.465 &               0.485 &               0.33 &     \textbf{0.322} &              0.454 &              0.496 \\
drugs   &  0.421 &              0.463 &               0.536 &              0.476 &              0.474 &  0.428 &               0.488 &     \textbf{0.318} &              0.337 &               0.52 &               0.62 \\
flare   &  0.694 &              0.712 &               0.735 &              0.727 &              0.731 &  0.629 &               0.653 &     \textbf{0.480} &              0.614 &              0.616 &              0.655 \\
grid    &  0.492 &              0.458 &               0.448 &              0.391 &     \textbf{0.158} &  0.468 &               0.468 &              0.749 &              0.668 &              0.194 &               0.52 \\
ads     &  0.352 &              0.234 &               0.322 &     \textbf{0.218} &              0.283 &  0.352 &               0.287 &              0.255 &              0.341 &              0.416 &              0.479 \\
mush    &  0.027 &              0.011 &               0.018 &     \textbf{0.010} &              0.012 &  0.054 &               0.017 &              0.098 &              0.054 &              0.022 &              0.364 \\
music   &  0.748 &               0.77 &               0.792 &              0.751 &              0.711 &  0.651 &               0.666 &     \textbf{0.465} &              0.572 &              0.614 &              0.684 \\
musk    &  0.367 &              0.277 &               0.359 &              0.277 &     \textbf{0.180} &  0.379 &               0.289 &              0.248 &              0.321 &              0.313 &              0.509 \\
craft   &  0.602 &              0.515 &               0.509 &              0.509 &              0.549 &  0.543 &               0.492 &              0.344 &              0.684 &     \textbf{0.324} &               0.52 \\
spam    &  0.595 &              0.246 &               0.263 &     \textbf{0.216} &              0.236 &  0.537 &               0.264 &              0.261 &              0.638 &              0.217 &              0.519 \\
alco    &  0.693 &              0.731 &               0.741 &              0.746 &              0.695 &  0.625 &               0.647 &     \textbf{0.495} &              0.608 &              0.692 &              0.658 \\
study   &  0.589 &     \textbf{0.382} &               0.428 &              0.385 &              0.386 &  0.538 &      \textbf{0.382} &               0.61 &              0.696 &              0.567 &              0.641 \\
telco   &  0.571 &              0.582 &               0.600 &              0.583 &              0.603 &  0.525 &               0.544 &     \textbf{0.373} &              0.476 &              0.541 &              0.648 \\
thrm    &  0.773 &              0.694 &               0.679 &              0.677 &              0.675 &  0.655 &               0.627 &     \textbf{0.491} &              0.629 &              0.494 &              0.604 \\
turk    &  0.847 &              0.851 &               0.845 &              0.848 &              0.836 &  0.684 &               0.692 &     \textbf{0.558} &               0.64 &              0.562 &              0.734 \\
vgame   &  0.631 &              0.571 &               0.659 &              0.608 &              0.601 &  0.570 &               0.533 &     \textbf{0.407} &              0.594 &              0.749 &              0.699 \\
voice   &  0.346 &              0.081 &               0.089 &     \textbf{0.077} &               0.08 &  0.378 &               0.126 &              0.166 &              0.417 &              0.103 &              0.323 \\
wine    &  0.750 &              0.655 &               0.604 &              0.656 &              0.622 &  0.637 &               0.596 &              0.662 &              0.905 &     \textbf{0.408} &              0.661 \\
yeast   &  0.839 &              0.759 &               0.672 &              0.717 &              0.697 &  0.680 &               0.653 &              0.569 &              0.881 &     \textbf{0.516} &               0.78 \\\midrule
Mean    &  0.549 &              0.459 &               0.475 &              0.454 &              0.438 &  0.501 &               0.444 &     \textbf{0.394} &              0.567 &              0.413 &              0.548 \\
\bottomrule
\end{tabular}

    }
	\subcaption{Average AE scores for each algorithm per dataset}
	\label{tab:clf_binary_CC_AE}
\end{subtable}%
\hspace{\fill}
\begin{subtable} {0.49 \linewidth}
	\resizebox{\linewidth}{!}{
        \begin{tabular}{l|ccccccccccc}
\toprule
{} &     CC & \makecell{CC\\-LR} &  \makecell{CC\\-RF} & \makecell{CC\\-AB} & \makecell{CC\\-SV} &    PCC & \makecell{PCC\\-LR} & \makecell{SVM\\-K} & \makecell{SVM\\-Q} & \makecell{RBF\\-K} & \makecell{RBF\\-Q} \\
\midrule
bc-cat  &  0.182 &     \textbf{0.027} &               0.060 &              0.038 &              0.055 &  0.123 &               0.046 &               0.08 &              0.316 &              0.038 &               0.05 \\
bc-cont &  0.067 &              0.013 &               0.025 &              0.026 &              0.026 &  0.060 &               0.063 &              0.035 &              0.447 &              0.033 &     \textbf{0.009} \\
cars    &  0.099 &              0.048 &               0.065 &              0.046 &     \textbf{0.038} &  0.083 &               0.045 &              0.051 &              0.045 &              0.241 &              0.288 \\
conc    &  0.495 &              0.206 &               0.196 &              0.168 &              0.146 &  0.245 &               0.151 &              0.074 &              0.306 &     \textbf{0.067} &              0.253 \\
contra  &  0.581 &              0.541 &               0.430 &              0.461 &              0.514 &  0.286 &                0.28 &     \textbf{0.197} &              0.382 &              0.213 &              0.352 \\
cappl   &  0.244 &              0.232 &               0.218 &              0.177 &              0.238 &  0.159 &               0.173 &              0.093 &     \textbf{0.086} &              0.188 &              0.227 \\
drugs   &  0.144 &              0.223 &               0.322 &              0.242 &              0.244 &  0.134 &               0.171 &     \textbf{0.078} &              0.088 &              0.239 &              0.269 \\
flare   &  0.420 &              0.494 &               0.503 &               0.48 &              0.498 &  0.256 &               0.275 &     \textbf{0.159} &              0.243 &              0.259 &              0.295 \\
grid    &  0.188 &              0.152 &               0.176 &              0.124 &     \textbf{0.030} &  0.151 &               0.145 &              0.596 &              0.425 &              0.037 &               0.23 \\
ads     &  0.134 &              0.075 &               0.120 &     \textbf{0.056} &              0.108 &  0.108 &               0.078 &              0.071 &              0.107 &              0.156 &              0.173 \\
mush    &  0.003 &     \textbf{0.001} &               0.002 &     \textbf{0.001} &     \textbf{0.001} &  0.006 &               0.002 &              0.016 &              0.007 &              0.002 &              0.202 \\
music   &  0.474 &              0.537 &               0.545 &              0.477 &              0.451 &  0.270 &               0.284 &     \textbf{0.136} &              0.204 &               0.29 &              0.369 \\
musk    &  0.116 &              0.074 &               0.127 &              0.078 &     \textbf{0.041} &  0.109 &               0.073 &              0.049 &              0.087 &              0.088 &              0.283 \\
craft   &  0.318 &              0.216 &               0.208 &              0.231 &               0.25 &  0.199 &               0.167 &               0.09 &              0.306 &     \textbf{0.079} &              0.211 \\
spam    &  0.351 &              0.063 &               0.071 &              0.047 &              0.057 &  0.200 &               0.062 &              0.061 &              0.298 &     \textbf{0.045} &              0.265 \\
alco    &  0.392 &              0.501 &               0.498 &              0.501 &              0.458 &  0.254 &               0.273 &     \textbf{0.167} &              0.238 &              0.363 &              0.308 \\
study   &  0.337 &              0.153 &               0.162 &              0.124 &              0.153 &  0.202 &      \textbf{0.118} &              0.213 &              0.283 &              0.233 &              0.306 \\
telco   &  0.284 &              0.316 &               0.311 &               0.31 &              0.352 &  0.186 &               0.205 &     \textbf{0.099} &              0.151 &              0.224 &              0.299 \\
thrm    &  0.534 &              0.433 &               0.387 &               0.38 &              0.386 &  0.275 &               0.259 &     \textbf{0.164} &              0.295 &              0.176 &              0.282 \\
turk    &  0.613 &              0.642 &               0.637 &              0.652 &              0.593 &  0.292 &               0.299 &              0.215 &              0.294 &     \textbf{0.195} &              0.391 \\
vgame   &  0.323 &              0.287 &               0.373 &              0.325 &              0.312 &  0.215 &               0.196 &     \textbf{0.114} &              0.267 &              0.443 &              0.397 \\
voice   &  0.153 &              0.011 &               0.013 &     \textbf{0.010} &              0.011 &  0.113 &               0.021 &              0.032 &              0.183 &              0.013 &               0.15 \\
wine    &  0.524 &              0.379 &               0.296 &              0.389 &              0.328 &  0.262 &               0.237 &              0.248 &              0.513 &     \textbf{0.115} &              0.392 \\
yeast   &  0.652 &              0.532 &               0.370 &              0.424 &              0.434 &  0.291 &               0.273 &              0.228 &              0.636 &     \textbf{0.174} &              0.501 \\\midrule
Mean    &  0.318 &              0.256 &               0.255 &               0.24 &              0.239 &  0.187 &               0.162 &     \textbf{0.136} &              0.259 &              0.163 &              0.271 \\
\bottomrule
\end{tabular}

    }
\subcaption{Average NKLD scores for each algorithm per dataset}
	\label{tab:clf_binary_CC_NKLD}
\end{subtable}

\caption{Results of classify and count-based quantifiers in the \textbf{binary} setting, where the base classifiers were tuned with respect to their accuracy. We show the averaged error scores for all scenarios per algorithm and dataset. Algorithms based on untuned logistic regression classifiers are denoted as before (no suffix), alternative tuned base classifiers are marked with respective suffixes: logistic regressors (LR), support vector machines (SV), random forests (RF) and AdaBoost (AB).
In addition, we present results for the \emph{SVM-K} and \emph{SVM-Q} methods and their adaptations that use an RBF-kernel (\emph{RBF-K} and \emph{RBF-Q}). 
For these methods, there appears to be a slight trend that the tuned variants work better.}
\label{tab:clf_binary_cc}
\end{table}


\begin{table}
\begin{subtable} {1.0 \linewidth}
	\resizebox{\linewidth}{!}{
        \begin{tabular}{l|ccccccccccccccccccccc}
\toprule
{} &    GAC &  \makecell{GAC\\-LR} &  \makecell{GAC\\-RF} &  \makecell{GAC\\-AB} & \makecell{GAC\\-SV} &            GPAC &  \makecell{GPAC\\-LR} &    FMM & \makecell{FMM\\-LR} &  \makecell{FMM\\-SV} &              FM &  \makecell{FM\\-LR} &              EM & \makecell{EM\\-LR} &     CC &  \makecell{CC\\-LR} &  \makecell{CC\\-RF} &  \makecell{CC\\-AB} &  \makecell{CC\\-SV} &    PCC &  \makecell{PCC\\-LR} \\
\midrule
conc   &  0.486 &                0.313 &                0.299 &                0.423 &      \textbf{0.259} &           0.473 &                 0.298 &  0.564 &               0.494 &                0.294 &            0.51 &               0.305 &           0.498 &              0.283 &  0.915 &               0.555 &               0.563 &               0.733 &               0.459 &  0.692 &                0.527 \\
contra &  0.600 &                0.495 &                0.490 &                0.517 &               0.534 &           0.515 &                 0.579 &  0.467 &               0.463 &                0.551 &           0.512 &               0.620 &  \textbf{0.396} &              0.531 &  0.833 &               0.825 &               0.808 &               0.829 &               0.835 &  0.699 &                0.705 \\
drugs  &  0.256 &                0.252 &                0.284 &                0.391 &               0.247 &           0.199 &                 0.206 &  0.160 &      \textbf{0.157} &                0.185 &           0.181 &               0.203 &           0.218 &              0.278 &  0.465 &               0.516 &               0.623 &               0.648 &               0.518 &  0.482 &                0.554 \\
craft  &  0.296 &                0.238 &                0.250 &                0.337 &                0.28 &  \textbf{0.190} &                 0.194 &  0.531 &                0.43 &                0.341 &  \textbf{0.190} &               0.199 &           0.191 &              0.235 &  0.752 &               0.666 &               0.673 &               0.716 &               0.707 &  0.654 &                0.622 \\
thrm   &  0.780 &                0.694 &                0.565 &                0.760 &               0.645 &           0.629 &                 0.658 &  0.619 &                0.56 &                0.582 &           0.663 &               0.751 &  \textbf{0.494} &              0.533 &  1.042 &               1.026 &               0.893 &               1.115 &               0.928 &  0.769 &                0.759 \\
turk   &  0.525 &                0.498 &                0.572 &                0.562 &               0.518 &           0.342 &                 0.385 &  0.324 &               0.324 &                0.472 &           0.392 &               0.451 &  \textbf{0.277} &              0.441 &  0.976 &               0.984 &               1.003 &               0.987 &               1.028 &  0.727 &                0.732 \\
vgame  &  0.520 &                0.517 &                0.529 &                0.567 &               0.536 &            0.46 &                 0.463 &  0.600 &               0.568 &                0.572 &           0.474 &               0.465 &  \textbf{0.322} &              0.339 &  0.590 &               0.574 &               0.658 &               0.694 &               0.614 &  0.520 &                0.493 \\
wine   &  0.656 &                0.553 &                0.572 &                0.699 &               0.567 &           0.575 &                 0.647 &  0.637 &               0.566 &                0.557 &           0.605 &               0.693 &           0.757 &     \textbf{0.460} &  0.965 &               0.777 &               0.708 &               0.843 &               0.647 &  0.636 &                0.586 \\
yeast  &  0.567 &                0.425 &                0.386 &                0.497 &               0.415 &           0.408 &                 0.399 &  0.505 &               0.491 &                0.466 &           0.413 &               0.387 &           0.613 &     \textbf{0.336} &  0.878 &               0.514 &               0.478 &               0.611 &               0.512 &  0.612 &                0.468 \\\midrule
Mean   &  0.521 &                0.443 &                0.438 &                0.528 &               0.445 &           0.421 &                 0.425 &  0.490 &                0.45 &                0.447 &           0.438 &               0.453 &           0.419 &     \textbf{0.382} &  0.824 &               0.715 &               0.712 &               0.797 &               0.694 &  0.643 &                0.605 \\
\bottomrule
\end{tabular}

    }
	\subcaption{Average AE scores for each \textbf{natural multiclass quantifier} per dataset}
	\label{tab:clf_multiclass_mc_AE}
\end{subtable}

\begin{subtable} {1.0 \linewidth}
	\resizebox{\linewidth}{!}{
        \begin{tabular}{l|ccccccccccccccccccccc}
\toprule
{} &    GAC &  \makecell{GAC\\-LR} &  \makecell{GAC\\-RF} &  \makecell{GAC\\-AB} &  \makecell{GAC\\-SV} &   GPAC &  \makecell{GPAC\\-LR} &    FMM &  \makecell{FMM\\-LR} &  \makecell{FMM\\-SV} &     FM &  \makecell{FM\\-LR} &              EM & \makecell{EM\\-LR} &     CC &  \makecell{CC\\-LR} &  \makecell{CC\\-RF} &  \makecell{CC\\-AB} &  \makecell{CC\\-SV} &    PCC & \makecell{PCC\\-LR} \\
\midrule
conc   &  0.310 &                0.212 &                0.192 &                0.312 &                0.162 &  0.467 &                 0.226 &  0.407 &                0.335 &                0.252 &  0.455 &               0.234 &            0.46 &     \textbf{0.142} &  0.640 &               0.248 &               0.263 &               0.361 &               0.172 &  0.276 &               0.173 \\
contra &  0.448 &                0.340 &                0.335 &                0.359 &                0.368 &  0.469 &                 0.470 &  0.395 &                0.373 &                0.485 &  0.445 &               0.480 &  \textbf{0.237} &              0.256 &  0.464 &               0.458 &               0.451 &               0.469 &               0.455 &  0.280 &               0.284 \\
drugs  &  0.180 &                0.141 &                0.177 &                0.246 &                0.132 &  0.150 &                 0.156 &  0.108 &                0.127 &                0.109 &  0.126 &               0.132 &  \textbf{0.049} &              0.189 &  0.151 &               0.225 &               0.311 &               0.353 &               0.236 &  0.147 &               0.184 \\
craft  &  0.172 &                0.133 &                0.125 &                0.185 &                0.146 &  0.150 &                 0.133 &  0.438 &                0.401 &                0.298 &  0.117 &               0.123 &           0.159 &     \textbf{0.099} &  0.398 &               0.304 &               0.309 &               0.361 &               0.360 &  0.242 &                0.22 \\
thrm   &  0.605 &                0.531 &                0.481 &                0.575 &                0.510 &  0.648 &                 0.619 &  0.641 &                0.528 &                0.610 &  0.723 &               0.726 &           0.442 &     \textbf{0.303} &  0.692 &               0.648 &               0.496 &               0.711 &               0.539 &  0.340 &               0.334 \\
turk   &  0.412 &                0.348 &                0.378 &                0.405 &                0.384 &  0.347 &                 0.335 &  0.295 &                0.270 &                0.392 &  0.372 &               0.420 &  \textbf{0.105} &               0.22 &  0.585 &               0.606 &               0.691 &               0.636 &               0.639 &  0.296 &               0.299 \\
vgame  &  0.584 &                0.524 &                0.561 &                0.569 &                0.529 &  0.522 &                 0.501 &  0.548 &                0.498 &                0.597 &  0.509 &               0.472 &  \textbf{0.133} &              0.136 &  0.238 &               0.247 &               0.363 &               0.412 &               0.312 &  0.170 &               0.157 \\
wine   &  0.434 &                0.466 &                0.520 &                0.580 &                0.534 &  0.620 &                 0.594 &  0.620 &                0.546 &                0.586 &  0.617 &               0.621 &           0.781 &              0.247 &  0.714 &               0.492 &               0.446 &               0.606 &               0.372 &  0.240 &      \textbf{0.209} \\
yeast  &  0.358 &                0.380 &                0.298 &                0.407 &                0.328 &  0.431 &                 0.362 &  0.593 &                0.595 &                0.534 &  0.401 &               0.340 &           0.702 &              0.213 &  0.585 &               0.234 &               0.296 &               0.501 &               0.325 &  0.224 &      \textbf{0.143} \\\midrule
Mean   &  0.389 &                0.342 &                0.341 &                0.404 &                0.343 &  0.423 &                 0.377 &  0.449 &                0.408 &                0.429 &  0.418 &               0.394 &           0.341 &     \textbf{0.201} &  0.497 &               0.385 &               0.403 &               0.490 &               0.379 &  0.246 &               0.222 \\
\bottomrule
\end{tabular}

    }
	\subcaption{Average NKLD scores for each \textbf{natural multiclass quantifier} per dataset}
	\label{tab:clf_multiclass_mc_NKLD}
\end{subtable}

\begin{subtable} {1.0 \linewidth}
	\resizebox{\linewidth}{!}{
        \begin{tabular}{l|cccccccccccccccccccccccccccc}
\toprule
{} &     AC &  \makecell{AC\\-LR} & \makecell{AC\\-RF} &  \makecell{AC\\-AB} &  \makecell{AC\\-SV} &    PAC &  \makecell{PAC\\-LR} &    TSX &  \makecell{TSX\\-LR} &  \makecell{TSX\\-SV} &   TS50 &  \makecell{TS50\\-LR} &  \makecell{TS50\\-SV} &  TSMax &  \makecell{TSMax\\-LR} & \makecell{TSMax\\-SV} &              MS & \makecell{MS\\-LR} &  \makecell{MS\\-SV} &    DyS &  \makecell{DyS\\-LR} &  \makecell{DyS\\-SV} &             FMM & \makecell{FMM\\-LR} &  \makecell{FMM\\-SV} &    HDy & \makecell{HDy\\-LR} & \makecell{HDy\\-SV} \\
\midrule
conc   &  0.864 &               0.490 &              0.328 &               0.405 &               0.292 &  0.574 &                0.521 &  0.615 &                0.511 &                0.281 &  0.591 &                 0.513 &                 0.279 &  0.502 &                  0.434 &        \textbf{0.274} &           0.508 &              0.452 &               0.299 &  0.562 &                0.518 &                0.372 &           0.564 &               0.494 &                0.294 &  0.536 &               0.459 &                0.34 \\
contra &  0.829 &               0.490 &               0.54 &               0.543 &               0.583 &  0.483 &                0.468 &  0.496 &                0.494 &                0.616 &  0.508 &                 0.496 &                 0.615 &  0.466 &                  0.459 &                 0.525 &           0.462 &     \textbf{0.453} &               0.519 &  0.538 &                0.575 &                0.675 &           0.467 &               0.463 &                0.551 &  0.481 &               0.487 &               0.569 \\
drugs  &  0.228 &               0.157 &              0.351 &               0.270 &               0.211 &  0.166 &                0.165 &  0.170 &                0.158 &                0.185 &  0.177 &                 0.165 &                 0.177 &  0.171 &                  0.168 &                 0.193 &  \textbf{0.147} &               0.16 &               0.184 &  0.213 &                0.171 &                0.209 &            0.16 &               0.157 &                0.185 &  0.180 &                0.17 &               0.205 \\
craft  &  0.560 &               0.399 &     \textbf{0.290} &               0.344 &               0.379 &  0.525 &                0.467 &  0.515 &                0.409 &                0.338 &  0.488 &                 0.377 &                 0.312 &  0.474 &                  0.395 &                 0.327 &           0.464 &              0.422 &               0.330 &  0.494 &                0.539 &                0.380 &           0.531 &                0.43 &                0.341 &  0.475 &                0.41 &               0.395 \\
thrm   &  1.297 &               0.578 &              0.575 &               0.642 &               0.566 &  0.633 &                0.579 &  0.726 &                0.643 &                0.692 &  0.684 &                 0.626 &                 0.662 &  0.593 &                  0.524 &                 0.536 &           0.587 &     \textbf{0.521} &               0.537 &  0.694 &                0.702 &                0.669 &           0.619 &                0.56 &                0.582 &  0.634 &               0.636 &               0.549 \\
turk   &  0.651 &               0.382 &              0.691 &               0.643 &               0.577 &  0.326 &                0.338 &  0.375 &                0.378 &                0.614 &  0.392 &                 0.401 &                 0.632 &  0.349 &                  0.361 &                  0.49 &           0.348 &              0.359 &               0.490 &  0.455 &                0.432 &                0.568 &  \textbf{0.324} &      \textbf{0.324} &                0.472 &  0.372 &               0.382 &               0.492 \\
vgame  &  0.741 &               0.591 &              0.699 &               0.707 &               0.640 &  0.640 &                0.586 &  0.630 &                0.604 &                0.613 &  0.626 &                 0.598 &                 0.611 &  0.574 &                  0.543 &                  0.54 &           0.575 &              0.548 &               0.547 &  0.557 &                0.557 &                0.567 &             0.6 &               0.568 &                0.572 &  0.521 &      \textbf{0.518} &               0.555 \\
wine   &  1.061 &               0.618 &              0.708 &               0.720 &               0.632 &  0.706 &                0.613 &  0.700 &                0.618 &                0.611 &  0.693 &                 0.641 &                 0.625 &  0.595 &                  0.522 &                 0.515 &           0.607 &              0.538 &               0.524 &  0.719 &                0.591 &                0.616 &           0.637 &               0.566 &                0.557 &  0.546 &               0.511 &      \textbf{0.509} \\
yeast  &  1.015 &               0.481 &              0.494 &               0.500 &               0.476 &  0.541 &                0.533 &  0.518 &                0.498 &                0.492 &  0.487 &                 0.478 &                 0.501 &  0.446 &                  0.436 &                 0.422 &           0.464 &              0.463 &               0.449 &  0.527 &                0.438 &                0.491 &           0.505 &               0.491 &                0.466 &  0.412 &      \textbf{0.398} &               0.442 \\\midrule
Mean   &  0.805 &               0.465 &              0.519 &               0.530 &               0.484 &  0.510 &                0.474 &  0.527 &                0.479 &                0.494 &  0.516 &                 0.477 &                 0.491 &  0.463 &                  0.427 &        \textbf{0.425} &           0.462 &              0.435 &               0.431 &  0.529 &                0.503 &                0.505 &            0.49 &                0.45 &                0.447 &  0.462 &               0.441 &               0.451 \\
\bottomrule
\end{tabular}

    }
\subcaption{Average AE scores for each \textbf{one-v.-rest-based quantifier} per dataset}
	\label{tab:clf_multiclass_ovr_AE}
\end{subtable}

\begin{subtable} {1.0 \linewidth}
	\resizebox{\linewidth}{!}{
        \begin{tabular}{l|cccccccccccccccccccccccccccc}
\toprule
{} &     AC &  \makecell{AC\\-LR} &  \makecell{AC\\-RF} &  \makecell{AC\\-AB} &  \makecell{AC\\-SV} &    PAC &  \makecell{PAC\\-LR} &    TSX &  \makecell{TSX\\-LR} &  \makecell{TSX\\-SV} &   TS50 &  \makecell{TS50\\-LR} &  \makecell{TS50\\-SV} &  TSMax &  \makecell{TSMax\\-LR} &  \makecell{TSMax\\-SV} &     MS &  \makecell{MS\\-LR} &  \makecell{MS\\-SV} &    DyS & \makecell{DyS\\-LR} & \makecell{DyS\\-SV} &    FMM &  \makecell{FMM\\-LR} &  \makecell{FMM\\-SV} &             HDy & \makecell{HDy\\-LR} & \makecell{HDy\\-SV} \\
\midrule
conc   &  0.841 &               0.360 &               0.280 &               0.359 &               0.236 &  0.443 &                0.406 &  0.439 &                0.371 &                0.250 &  0.410 &                 0.381 &                 0.229 &  0.362 &                  0.303 &                  0.175 &  0.393 &               0.341 &               0.243 &  0.304 &               0.246 &      \textbf{0.165} &  0.407 &                0.335 &                0.252 &           0.275 &               0.209 &               0.173 \\
contra &  0.662 &               0.383 &               0.483 &               0.463 &               0.489 &  0.425 &                0.392 &  0.412 &                0.386 &                0.537 &  0.433 &                 0.425 &                 0.543 &  0.333 &                  0.297 &                  0.377 &  0.350 &               0.324 &               0.399 &  0.312 &               0.351 &               0.402 &  0.395 &                0.373 &                0.485 &  \textbf{0.275} &                0.28 &               0.337 \\
drugs  &  0.164 &               0.104 &               0.288 &               0.191 &               0.134 &  0.100 &                0.145 &  0.125 &                0.123 &                0.110 &  0.091 &                 0.093 &                 0.095 &  0.074 &                  0.089 &                  0.080 &  0.087 &               0.120 &               0.099 &  0.069 &               0.054 &               0.079 &  0.108 &                0.127 &                0.109 &  \textbf{0.046} &               0.048 &                0.07 \\
craft  &  0.502 &               0.353 &               0.199 &               0.283 &               0.344 &  0.457 &                0.459 &  0.423 &                0.370 &                0.318 &  0.377 &                 0.327 &                 0.302 &  0.420 &                  0.331 &                  0.231 &  0.416 &               0.372 &               0.254 &  0.222 &               0.299 &      \textbf{0.181} &  0.438 &                0.401 &                0.298 &           0.218 &               0.203 &               0.192 \\
thrm   &  0.969 &               0.574 &               0.525 &               0.652 &               0.548 &  0.608 &                0.523 &  0.729 &                0.635 &                0.713 &  0.706 &                 0.643 &                 0.694 &  0.530 &                  0.444 &                  0.510 &  0.533 &               0.460 &               0.537 &  0.517 &                0.49 &               0.507 &  0.641 &                0.528 &                0.610 &           0.502 &                0.49 &      \textbf{0.418} \\
turk   &  0.580 &               0.353 &               0.636 &               0.592 &               0.494 &  0.320 &                0.317 &  0.377 &                0.361 &                0.548 &  0.396 &                 0.379 &                 0.560 &  0.260 &                  0.245 &                  0.389 &  0.259 &               0.243 &               0.391 &  0.274 &               0.277 &               0.331 &  0.295 &                0.270 &                0.392 &  \textbf{0.193} &               0.213 &               0.286 \\
vgame  &  0.717 &               0.519 &               0.758 &               0.711 &               0.630 &  0.620 &                0.536 &  0.555 &                0.515 &                0.598 &  0.515 &                 0.482 &                 0.549 &  0.485 &                  0.460 &                  0.532 &  0.492 &               0.480 &               0.559 &  0.364 &      \textbf{0.342} &               0.389 &  0.548 &                0.498 &                0.597 &           0.385 &               0.375 &               0.427 \\
wine   &  0.810 &               0.598 &               0.700 &               0.728 &               0.630 &  0.714 &                0.617 &  0.690 &                0.596 &                0.604 &  0.665 &                 0.610 &                 0.608 &  0.521 &                  0.444 &                  0.471 &  0.552 &               0.496 &               0.522 &  0.537 &               0.371 &               0.422 &  0.620 &                0.546 &                0.586 &            0.41 &      \textbf{0.334} &               0.353 \\
yeast  &  0.817 &               0.534 &               0.497 &               0.494 &               0.493 &  0.598 &                0.605 &  0.580 &                0.588 &                0.543 &  0.502 &                 0.532 &                 0.541 &  0.485 &                  0.484 &                  0.452 &  0.519 &               0.543 &               0.501 &  0.479 &               0.344 &      \textbf{0.324} &  0.593 &                0.595 &                0.534 &           0.342 &                0.34 &               0.358 \\\midrule
Mean   &  0.674 &               0.420 &               0.485 &               0.497 &               0.444 &  0.476 &                0.445 &  0.481 &                0.438 &                0.469 &  0.455 &                 0.430 &                 0.458 &  0.385 &                  0.344 &                  0.357 &  0.400 &               0.375 &               0.389 &  0.342 &               0.308 &               0.311 &  0.449 &                0.408 &                0.429 &           0.294 &      \textbf{0.277} &               0.291 \\
\bottomrule
\end{tabular}

    }
\subcaption{Average NKLD scores for each \textbf{one-v.-rest-based quantifier}  per dataset}
	\label{tab:clf_multiclass_ovr_NKLD}
\end{subtable}

\caption{
Results of quantifiers that use tuned base classifiers in the \textbf{multiclass} setting.
For natural multiclass quantifiers, base classifiers were tuned with respect to their accuracy.
For quantifiers that use the one-vs.-rest approach, the binary base classifiers were tuned with respect to balanced accuracy.
We show the averaged error scores for all scenarios per algorithm and dataset. 
Algorithms based on untuned logistic regression classifiers are denoted as before (no suffix), alternative tuned base classifiers are marked with respective suffixes: logistic regressors (LR), support vector machines (SV), random forests (RF) and AdaBoost (AB).
Again, tuning base classifiers appears to improve the aggregate error scores, but this trend is not consistent over all datasets.
}
\label{tab:clf_multiclass}
\end{table}

\FloatBarrier

\section{Parameter Settings in the LeQua Case Study}\label{ap:lequa_params}

As noted in the main text, in the experiments on the LeQua challenge dataset, we used the same parameters as described in Section \ref{sec:param_main} for the experiments using untuned quantifiers, and the same parameters as described in Section \ref{sec:setting_tuned} for the experiments with tuned base classifiers.

In the same case study, we further explored the effects of tuning quantifiers with respect to the absolute error (AE) score on the given validation data.
In this experiment, we chose the following parameter grids to optimize on:

\begin{itemize}
    \item For all quantification methods that require a base classifier, a logistic regression classifier was chosen as base classifier. The parameters of this classifier were tuned individually for each quantifier, and in the corresponding grid search we varied the regularization weight $C$ within the set $\{2^i: i\in \{-15,-13,-11,\dots,13,15\}\}$. Further, for all values of $C$, we varied the weighting strategy for the instances, by either setting the weights of all instances to 1, or weighting the instances inversely proportional to the prevalence of their corresponding class. Like in all previous experiments, we applied the \texttt{L-BFGS} solver to efficiently learn the corresponding models, and set the number of maximum iterations to 1000.
    \item For the \emph{DyS} method, we varied the number of bins that the confidence scores from base classifiers were placed in among the values $\{2, 4, 6, 8, 10, 15, 20\}$.
    \item For the \emph{readme} method, we varied the number of features that were sampled for each subset among the values $\{2, 4, 6, 8, 10, 15, 20\}$.
    \item For the \emph{PWK} method, we used the same parameter grid that has been used in the experiments by \cite{barranquero_study_2013} when they proposed this method. Thus, we varied the number of neighbors to consider among the set $\{1,3,5,7,11,15,25,35,45\}$, and the weight factor $\alpha$ was varied in the set $\{1,2,3,4,5\}$.
    \item For the \SVMperf-based quantifiers, we tested tuning the variants of the \emph{SVM-K} and \emph{SVM-Q} methods which applied an RBF-kernel function. Toward that end, we varied the kernel parameter $\gamma$ among the values $ \{2^i: i\in \{-17,-15,-13,\dots,3,5\}\}$.
\end{itemize}

\section{Additional Plots for the LeQua Case Study}\label{ap:lequa_plots}

\begin{figure}[b]
	\centering
	\begin{subfigure}[b]{0.495\textwidth}
		\centering
		\includegraphics[width=\textwidth]{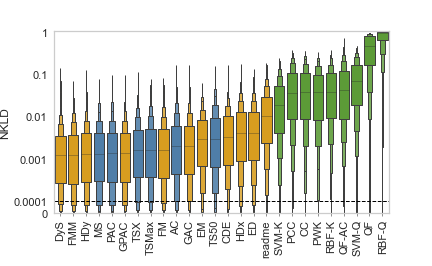}
		\caption{Distribution of NKLD scores on the \textbf{binary} LeQua data}
	\end{subfigure}
    \hfill
	\begin{subfigure}[b]{0.495\textwidth}
		\centering
		\includegraphics[width=\textwidth]{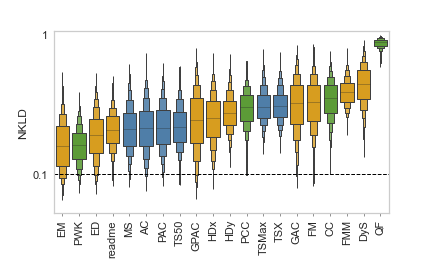}
		\caption{Distribution of NKLD scores on the \textbf{multiclass} LeQua data}
	\end{subfigure}
	\caption{Results of our experiments with untuned quantifiers on the LeQua test sets with respect to the NKLD. Plots are scaled logarithmically above the dotted vertical threshold, and linearly below.  Colors indicate the category of the algorithm. On the binary data, overall results are mostly in line with our findings from the main experiments and results with respect to the absolute error (AE) scores. On the multiclass data, results appear to differ from results with respect to the AE score, but the \emph{EM} method still appears to work best.}
	\label{fig:lequa_main_nkld}
\end{figure}

\begin{figure}
	\centering
		\includegraphics[width=\textwidth]{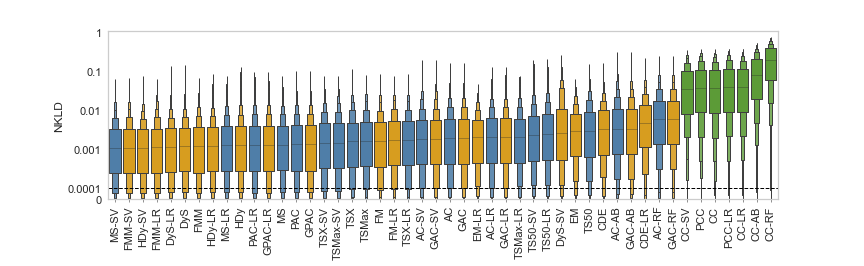}
	\caption{
    Results of our experiments with quantifiers that apply tuned classifiers on the \textbf{binary} LeQua data.
    Plots are scaled logarithmically above the dotted vertical threshold, and linearly below.  Colors indicate the category of the algorithm. Algorithms based on untuned logistic regression classifiers are denoted as before (no suffix), alternative tuned base classifiers are marked with respective suffixes: logistic regressors (LR), support vector machines (SV), random forests (RF) and AdaBoost (AB).  In general, we observe no clear trend that tuning base classifiers has a positive impact on the outcome.}
	\label{fig:lequa_clf_nkld}
\end{figure}

Finally, in Figures \ref{fig:lequa_main_nkld} and \ref{fig:lequa_clf_nkld}, we present additional plots regarding the case study on the LeQua dataset, in which we depict results with respect to the NKLD error score.
On the binary data, results generally align with the results with respect to the AE measure.
By contrast, in the multiclass case, results appear quite different from those for the AE scores, or related results from the main experiments, as can be seen in Figure \ref{fig:lequa_main_nkld}(b).
As discussed in the main text, we attribute this to the NKLD not being particularly suitable for this setting.
Thus, we omit further plots of results with respect to the NKLD in the multiclass setting.
Further, we omit plots of the NKLD scores from the experiments from Section \ref{sec:lequa_tuned}, as we argue that these are not really meaningful---in these experiments, methods were optimized for the \emph{AE} measure.

\end{document}